\documentclass[twoside,11pt]{article}

\usepackage{jmlr2earxiv}

\usepackage{algorithm}
\usepackage{algorithmic}
\usepackage{xcolor}
\usepackage{caption}
\usepackage{amssymb}
\usepackage{amsmath}
\usepackage{mathabx}
\usepackage{lineno}

\jmlrheading{}{}{}{}{}{Chicharro00}{Daniel Chicharro, Michel Besserve, and Stefano Panzeri}

\ShortHeadings{}{}
\firstpageno{1}

\begin{document}

\title{Causal learning with sufficient statistics: an information bottleneck approach}

\author{\name $^{a,b}$Daniel Chicharro, $^c$Michel Besserve, $^a$Stefano Panzeri\\
\\%\email daniel.chicharro@iit.
        %\\%\email michel.besserve@tuebingen.mpg.de
        %\\ %\email stefano.panzeri@iit.it
       \addr $^a$Neural Computation Laboratory, Center for Neuroscience and Cognitive Systems@UniTn,\\
       Istituto Italiano di Tecnologia, Corso Bettini 31, 38068 Rovereto, Italy.\\
       $^b$Dept.\,of Neurobiology, Harvard Medical School,\\
       220 Longwood Ave, Boston, MA 02115, USA. \\
       $^c$Max Planck Institute for Intelligent Systems and Max Planck ETH Center for Learning Systems,\\
        Max Planck Institute for Biological Cybernetics,\\
        Max-Planck-Ring, 72076 Tubingen, Germany.}

\editor{}

\maketitle

\begin{abstract}%
The inference of causal relationships using observational data from partially observed multivariate systems with hidden variables is a fundamental question in many scientific domains. Methods extracting causal information from conditional independencies between variables of a system are common tools for this purpose, but are limited in the lack of independencies. To surmount this limitation, we capitalize on the fact that the laws governing the generative mechanisms of a system often result in substructures embodied in the generative functional equation of a variable, which act as sufficient statistics for the influence that other variables have on it. These functional sufficient statistics constitute intermediate hidden variables providing new conditional independencies to be tested. We propose to use the Information Bottleneck method, a technique commonly applied for dimensionality reduction, to find underlying sufficient sets of statistics. Using these statistics we formulate new additional rules of causal orientation that provide causal information not obtainable from standard structure learning algorithms, which exploit only conditional independencies between observable variables. We validate the use of sufficient statistics for structure learning both with simulated systems built to contain specific sufficient statistics and with benchmark data from regulatory rules previously and independently proposed to model biological signal transduction networks.
\end{abstract}

\vspace*{10mm}


\vspace*{6mm}
{\small Corresponding author: Daniel Chicharro, Department of Neurobiology, Harvard Medical School, Warren Alpert Bldg 222, 220 Longwood Ave, Boston, MA 02115, USA. E-mail: chicharro31@yahoo.es}

\vspace*{4mm}

\section{Introduction}

Methods based on conditional independencies are a well-established framework for causal structure learning from observational data \citep{Spirtes00,Pearl09,Drton17,Heinze17,Petersbook,Malinsky18,Glymour19}. Orientation rules based on conditional independencies allow constructing a partially oriented graph \citep{Zhang08c} representing the equivalence class of all causal structures compatible with the set of conditional independencies present in the distribution of the observable variables (the so-called Markov equivalence class). However, the power of these methods is limited by a lack of independencies, e.g.\,in highly interdependent systems or in the presence of hidden variables. Beyond conditional independencies, the causal structure of a system also imposes further equality \citep{Tian2002b} and inequality \citep{Kang06} constraints, which can be used to test if a concrete causal model is compatible with some given observational data. To further discriminate within the Markov equivalence classes, other methods exploit specific properties associated with certain forms of the functional equations generating the variables, such as linear models with non-Gaussian noise \citep{Shimizu11}, or additive-noise models \citep{Hoyer2009, Zhang09, Chicharro19}, and hence can only provide additional causal knowledge when the system contains equations with the required form.

We here propose a method to augment the power of structure learning algorithms based on conditional independencies by identifying endogenous sufficient statistics present in the generative mechanisms of a system. The causal structure of a system corresponds to the set of variables (parents) constituting the arguments of the functional equations characterizing the generative mechanisms of each variable. Because the shape of these equations is determined by the physical laws ruling the system, often the dependence of a variable on several parents is structured in subfunctions embedded within its functional equation. For example, it is common that several parents contribute additively, such that it is their sum rather than their individual value that is informative about the variable generated by the functional equation. Such a subfunction hence deterministically determines an endogenous variable which acts as a sufficient statistic \citep{Casella02}, containing all the information its arguments have about the generated variable. Therefore, finding sufficient statistics within the functional equations uncovers additional conditional independencies useful to further discriminate which causal structures are compatible with the data. Importantly, inferring the form of a sufficient statistic does not require modeling the full functional equation of a variable, as it is done in score-based approaches to structure learning \citep{Chickering2002,Chickering03}. A subfunction creating a functional sufficient statistic may be much simpler than the full functional equation, which may involve a higher number of variables than the subequation, or may even be non-identifiable, in the presence of hidden variables.

Structure learning in the presence of deterministic relations has been studied before \citep{Geiger90,Spirtes00,Lemeire2012,Mabrouk2014}. However, this previous work considered deterministic relations between variables of the system, which hinder causal learning because they create independencies unfaithful to the causal structure. Conversely, the presence of endogenous functional sufficient statistics implies not deterministic relations between the variables of the system, but only with the hidden variables corresponding to the statistics. In this regard, whether a sufficient statistic is conceived as a meaningful hidden variable or simply as an auxiliary construct resulting from the form of the functional equation where it is embodied is not relevant. What matters is that the sufficient statistics create new conditional independencies between observable variables beneficiary to infer the causal structure. Accordingly, our proposal also differs substantially from other previous approaches to detect hidden variables \citep{Elidan2000,Silva06}. These approaches rely on detecting patterns of full connectivity between observable variables that could be explained by an unobserved common parent. A hidden common parent is conceived as an actual variable in the system, with its own functional equation, generally non-deterministic. Oppositely, sufficient statistics appear embedded within the functional equation of (most usually) a single observable variable, and hence do not operate as a hidden common parent.

We here propose to identify sufficient statistics from the new conditional independencies they create. The information bottleneck (IB) method \citep{Tishby99} is especially suited to infer sufficient statistics, since it determines low dimensional representations of a set of variables preserving the information about another target variable, in the way a functional sufficient statistic has to contain all the information some parents have about the variable in whose functional equation the statistic is embedded. This use of the IB method substantially differs from its application to model low dimensional latent common causes of multiple observable variables \citep{Elidan2005}. To our knowledge, sufficient statistics have not been previously exploited to extend the applicability of standard causal orientation rules based on conditional independencies.

This paper is organized as follows. Section \ref{ss1} reviews the basic causal orientation rules at the core of standard structure learning algorithms based on conditional independencies \citep{Spirtes00}. In Section \ref{ss2}, we provide a general presentation of functional sufficient sets of statistics and their application for structure learning from observational data in the presence of hidden variables; we introduce new orientation rules using sufficient statistics. In Section \ref{ss3} we describe how to use the IB method to find potential sufficient sets of statistics, and a general procedure to determine which identified statistics fulfill the necessary criteria to apply the causal orientation rules. We call this procedure the Information Bottleneck Sufficient Statistics Inference method, the IBSSI method. In Section \ref{ss4} we validate the applicability of the method to identify underlying sufficient statistics with high true positive rates and low false positive rates. We first study systems specifically designed to contain different types of sufficient sets of statistics. We then also apply the method to a concrete model of Boolean regulatory rules which has already previously been shown to accurately model a biological signal transduction network \citep{Li06}, and in whose structure we identify the presence of sufficient statistics. In Section \ref{ss5} we use sufficient statistics for structure learning in the presence of selection bias \citep{Spirtes96}, counteracting the selection bias by recovering conditional independencies between variables for which the bias introduced a dependence. In Section \ref{ss6} we further explore the combination of the IBSSI method with the identification of causal interventions, which can create new independencies \citep{Shpitser08}. Finally, in the Appendix, we provide further examples to illustrate the performance of the IBSSI method, we show in detail how the new rules are integrated within a standard algorithm \--such as the Causal Inference (CI) algorithm of \cite{Spirtes00}\--, and we describe how the standard faithfulness assumption \citep{Spirtes00,Pearl09} that ensures an isomorphic relation between conditional independencies and the causal structure can be equally formulated as an assumption of faithfulness between probability distributions from systems containing sufficient statistics and augmented causal graphs representing also those statistics.

\section{Preliminaries}
\label{ss1}

We first review the basic elements of structure learning from conditional independencies that underpin our application of sufficient statistics. We start with some basic notation for Directed Acyclic Graphs (DAGs). We use bold letters for sets and vectors. Consider a set of random variables $\mathbf{V} = \{ \mathrm{V}_1,...,\mathrm{V}_p\}$. A graph $G = (\mathbf{V}; \mathcal{E})$ consists of nodes $\mathbf{V}$ and edges $\mathcal{E}$  between the nodes. $(\mathrm{V}; \mathrm{V}) \notin \mathcal{E}$ for any $\mathrm{V} \in \mathbf{V}$. We write $\mathrm{V}_i \rightarrow \mathrm{V}_j$ for $(\mathrm{V}_i;\mathrm{V}_j)\in \mathcal{E}$. We denote by $\mathrm{V}$ both variable $\mathrm{V}$ and its corresponding node. A node $\mathrm{V}_i$ is called a parent of $\mathrm{V}_j$ if $(\mathrm{V}_i;\mathrm{V}_j)\in \mathcal{E}$. The set of parents of $\mathrm{V}_j$ is denoted by $\mathbf{Pa}_{\mathrm{V}_j}$. Two nodes $\mathrm{V}_i$ and $\mathrm{V}_j$ are adjacent if either $(\mathrm{V}_i; \mathrm{V}_j) \in \mathcal{E}$ or $(\mathrm{V}_j; \mathrm{V}_i) \in \mathcal{E}$. A path in $G$ is a sequence of (at least two) distinct nodes $\mathrm{V}_1, ... , \mathrm{V}_n,$ such that there is an edge between $\mathrm{V}_k$ and $\mathrm{V}_{k+1}$ for all $k = 1, ... , n-1 $. If all edges are $\mathrm{V}_k \rightarrow \mathrm{V}_{k+1}$ the path is a causal or directed path. The set of descendants $\mathbf{D}_{\mathrm{V}_i}$ of node $\mathrm{V}_i$ comprises those variables that can be reached going forward through causal pathways from $\mathrm{V}_i$. The set of non-descendants $\mathbf{ND}_{\mathrm{V}_i}$ is complementary to it. Since the graph is acyclic no node is its own descendant. A node $\mathrm{V}_i$ is a collider in a path if it has incoming arrows $\mathrm{V}_{i-1} \rightarrow \mathrm{V}_i \leftarrow \mathrm{V}_{i+1}$ and is a noncollider otherwise.

Assume that for a system the generative mechanisms of each variable $\mathrm{V}_i \in \mathbf{V}$ can be captured by a functional equation $\mathrm{V}_i := f_{\mathrm{V}_i}(\mathbf{Pa}_{\mathrm{V}_i}, \varepsilon_{\mathrm{V}_i})$, where $\varepsilon_{\mathrm{V}_i}$ represents exogenous noises and $\mathbf{Pa}_{\mathrm{V}_i}$ indicates that in the associated DAG representing the causal structure of the system each variable is connected by an incoming arrow to all and only the arguments of its functional equation. Structure learning from conditional independencies relies on the possibility to relate the causal structure to conditional independencies in the joint probability distribution of the variables. The keystone for this connection is the concept of \emph{d-separation} \citep{Pearl86}, which defines a graphical criterion of separability between nodes analogous to the statistical criterion of independence between variables. Two nodes $\mathrm{X}$ and $\mathrm{Y}$ are \emph{d-separated} given a set of nodes $\mathbf{S}$ if and only if no $\mathbf{S}$-active paths exist between $\mathrm{X}$ and $\mathrm{Y}$ \citep{Pearl86}. A path is active given the set of conditioning variables $\mathbf{S}$ ($\mathbf{S}$-active) if no noncollider in the path belongs to $\mathbf{S}$ and every collider either is in $\mathbf{S}$ or has a descendant in $\mathbf{S}$. Assuming that a causal structure $G$ and a generated probability distribution $p(\mathbf{V})$ are \emph{faithful} to one another, a conditional independence between $\mathrm{X}$ and $\mathrm{Y}$ given $\mathbf{S}$ \--denoted by $(\mathrm{X} \perp \mathrm{Y}|\mathbf{S})_{P}$\-- holds if and only if there is no $\mathbf{S}$-active path between them, that is, if $\mathrm{X}$ and $\mathrm{Y}$ are d-separated given $\mathbf{S}$ \--denoted by $(\mathrm{X} \perp \mathrm{Y}|\mathbf{S})_{G}$. Under this faithfulness assumption \citep{Spirtes00}, an isomorphic relation holds between $(\mathrm{X} \perp \mathrm{Y}|\mathbf{S})_{P}$ and $(\mathrm{X} \perp \mathrm{Y}|\mathbf{S})_{G}$, which can both simply be denoted by $\mathrm{X} \perp \mathrm{Y}|\mathbf{S}$. When only a subset of the variables are observable, two types of graphs have been used to represent only the causal relations between the observable variables, without explicitly including the hidden variables, namely the so-called Inducing Path Graphs (IPGs) \citep{Spirtes00} and Maximal Ancestral Graphs (MAGs) \citep{Richardson2002}. Despite their difference, it here suffices to say that they both represent with bidirected arrows ($\leftrightarrow$) the existence of paths between observable variables only containing hidden nodes, and that, incorporating $\leftrightarrow$ to the type of edges that create colliders and noncolliders, for these graphs the same graphical criterion of separation serves to connect conditional dependencies to the existence of active paths between variables. See Appendix A for a more formal review of causal models and the faithfulness assumption, which we also extend to systems containing sufficient statistics.

Structure learning algorithms based on conditional independencies use the connection between the independencies and the causal structure to construct a partially oriented graph which represents the class of causal structures that result in the same observed conditional independencies, the so-called Markov equivalence class \citep{Spirtes00, Pearl09}. The graph represents the causal properties common to any causal structure in the same class. An edge $\mathrm{V}_i \-- \mathrm{V}_j$ indicates that no conditioning set $\mathbf{S}$ can create an independence between $\mathrm{V}_i$ and $\mathrm{V}_j$, that is, that they are nonseparable. In these partially oriented graphs, like for IPGs and MAGs, in the presence of hidden variables nonseparability may also be due to active paths between the variables conformed by hidden variables, e.g.\,due to a hidden common cause ($\mathrm{V}_i \leftrightarrow \mathrm{V}_j$). The algorithms use orientation rules to infer the presence or lack of arrows from combinations of conditional dependencies and independencies. We use the notation $\bullet \--$ to refer to an edge in the partially oriented graph whose end is undetermined and we use $* \--$ as a placeholder for either $\bullet \--$, $\--$, or $\leftarrow$. Accordingly, $\mathrm{V}_i * \rightarrow \mathrm{V}_j$ indicates that an arrow pointing to $\mathrm{V}_j$ has been inferred, while the other end of the edge can have an arrow ($ \leftrightarrow $), no arrow ($\rightarrow $), or be undetermined ($\bullet\rightarrow$). Furthermore, $*-\underline{*\mathrm{V}_i*}-*$ indicates that it has been inferred that $\mathrm{V}_i$ is a noncollider.

\cite{Zhang08c} introduced additional orientation rules to the set used in the original FCI algorithm \citep{Spirtes00} and proved the completeness of that extended set to exploit all the causal information from conditional independencies. That set of rules \--or a subset of it\-- is at the core of all the structure learning algorithms based on conditional independencies. Many proposals in the literature have contributed refining the algorithms implementing these rules \citep[see][for a review]{Drton17,Heinze17,Petersbook,Malinsky18,Glymour19}. In this work our aim is to introduce additional rules that can be added to the complete set of standard rules, and which can then be included in any of the specific implementations. For this reason, we now focus on reviewing the two basic rules at the core of the complete set of rules of \cite{Zhang08c}. These two basic rules follow directly from the definition of d-separation \citep{Pearl86} and exploit the different effect of colliders and noncolliders in the propagation of dependencies, as stated in the following propositions:

\vspace*{2mm}
\noindent \textbf{Proposition 1}: \emph{Consider variables} $\mathrm{X}$, $\mathrm{Y}$, \emph{and} $\mathrm{Z}$, \emph{with} $\mathrm{X} \-- \mathrm{Y}$ \emph{and} $\mathrm{Y} \-- \mathrm{Z}$ \emph{nonseparable}.
\emph{If} $\exists \mathbf{S}$ \emph{nonoverlapping with} $\{ \mathrm{X}, \mathrm{Y}, \mathrm{Z}\}$ \emph{such that} $ \mathrm{X} \perp \mathrm{Z} |\mathbf{S}$, \emph{then in the underlying causal structure} $\mathrm{Y}$ \emph{is a collider in the junction of any $\mathbf{S}$-active path} $\mathrm{X} \cdot \cdot \cdot \mathrm{Y}$ \emph{and any $\mathbf{S}$-active path} $\mathrm{Y} \cdot \cdot \cdot \mathrm{Z}$.

\vspace*{1mm}
\noindent \textbf{Proposition 2}: \emph{Consider variables $\mathrm{X}$, $\mathrm{Y}$, and $\mathrm{Z}$ with $\mathrm{X} \-- \mathrm{Y}$ and $\mathrm{Y} \-- \mathrm{Z}$ nonseparable. If $\exists \mathbf{S}$ \emph{nonoverlapping with} $\{ \mathrm{X}, \mathrm{Y}, \mathrm{Z}\}$ \emph{such that} $\mathrm{X} \notperp \mathrm{Z} |\mathbf{S}$  and $ \mathrm{X} \perp \mathrm{Z} |\mathbf{S},\mathrm{Y}$, then in the underlying causal structure $\mathrm{Y}$ is a noncollider in the junction of any $\mathbf{S}$-active path $\mathrm{X} \cdot \cdot \cdot \mathrm{Y}$ and $\mathbf{S}$-active path $\mathrm{Y} \cdot \cdot \cdot \mathrm{Z}$}.

These propositions reflect that conditioning has the effect of activating colliders and inactivating noncolliders. The logic of Proposition 1 is that, since $\mathrm{Y}$ is not separable from $\mathrm{X}$ and $\mathrm{Z}$, for any conditioning set $\mathbf{S}$ there must be some active path between $\mathrm{X}$ and $\mathrm{Y}$ and between $\mathrm{Y}$ and $\mathrm{Z}$. If $\mathbf{S}$ does not include $\mathrm{Y}$, the concatenation of any $\mathbf{S}$-active path between $\mathrm{X}$ and $\mathrm{Y}$ and any $\mathbf{S}$-active path between $\mathrm{Y}$ and $\mathrm{Z}$ would result in an $\mathbf{S}$-active path between $\mathrm{X}$ and $\mathrm{Z}$, unless $\mathrm{Y}$ is a collider between those paths. In the case of Proposition 2, since it is the addition of $\mathrm{Y}$ to the conditioning set what creates the independence $ \mathrm{X} \perp \mathrm{Z} |\mathbf{S},\mathrm{Y}$, conditioning on $\mathrm{Y}$ must deactivate the $\mathbf{S}$-active paths responsible for $ \mathrm{X} \notperp \mathrm{Z} |\mathbf{S}$, which means that $\mathrm{Y}$ has to be a noncollider in those paths. These propositions indicate whether $\mathrm{Y}$ is a collider or a noncollider in specific $\mathbf{S}$-active paths, but in general determining which paths are $\mathbf{S}$-active in itself requires additional knowledge of the causal structure along those paths. However, the nonseparability of $\mathrm{X}$ and $\mathrm{Y}$ indicates that there is some path between $\mathrm{X}$ and $\mathrm{Y}$ active for $\mathbf{S} = \emptyset$ , which cannot be deactivated with any set $\mathbf{S}$ without simultaneously activating a $\mathbf{S}$-active path, and analogously from the nonseparability of $\mathrm{Z}$ and $\mathrm{Y}$. This means that, irrespectively of which $\mathbf{S}$ is selected, propositions 1 and 2 apply to those paths corresponding to the direct links $\mathrm{X} \-- \mathrm{Y}$ and $ \mathrm{Y} \-- \mathrm{Z}$ in the partially oriented graph. Therefore, under the assumption of faithfulness between conditional independencies and the causal structure, the propositions allow inferring whether $\mathrm{Y}$ is a collider or noncollider in the paths corresponding to the concatenation of $\mathrm{X} \-- \mathrm{Y}$ and $ \mathrm{Y} \-- \mathrm{Z}$:

\vspace*{2mm}
\noindent \textbf{Rule $\mathbf{R.c}$}\ \ \textbf{Inference of a collider}: \emph{Consider variables $\mathrm{X}$, $\mathrm{Y}$, and $\mathrm{Z}$, with $\mathrm{X} \-- \mathrm{Y}$ and $\mathrm{Y} \-- \mathrm{Z}$ nonseparable.
If $\exists \mathbf{S}$ nonoverlapping with $\{ \mathrm{X}, \mathrm{Y}, \mathrm{Z}\}$ such that $ \mathrm{X} \perp \mathrm{Z} |\mathbf{S}$, then orient $\mathrm{X} *\--*\mathrm{Y} *\--* \mathrm{Z}$ as $\mathrm{X} *\rightarrow \mathrm{Y} \leftarrow* \mathrm{Z}$}.

%The rule can equally be formulated substituting the condition on the existence of $S$ such that $ \mathrm{X} \perp \mathrm{Z} |\mathbf{S}$ by the condition that $\mathrm{X}$ and $\mathrm{Z}$ are separable, which is implied by the former. However, we formulate the rule explicitly in terms of the existence of a conditioning set $\mathbf{S}$ to facilitate the posterior comparison with the new analogous rule based on sufficient statistics.

\vspace*{1mm}
\noindent \textbf{Rule $\mathbf{R.nc}$}\ \ \textbf{Inference of a noncollider}: \emph{Consider variables $\mathrm{X}$, $\mathrm{Y}$, and $\mathrm{Z}$ with $\mathrm{X} \-- \mathrm{Y}$ and $\mathrm{Y} \-- \mathrm{Z}$ nonseparable. If $\exists \mathbf{S}$ nonoverlapping with $\{ \mathrm{X}, \mathrm{Y}, \mathrm{Z}\}$ such that $\mathrm{X} \notperp \mathrm{Z} |\mathbf{S}$  and $ \mathrm{X} \perp \mathrm{Z} |\mathbf{S},\mathrm{Y}$, then mark $\mathrm{X} *\--*\mathrm{Y} *\--* \mathrm{Z}$ as $\mathrm{X} *-\underline{*\mathrm{Y}*}-* \mathrm{Z}$}.
%\vspace*{1mm}

\vspace*{1mm}
These two rules are at the core of the structure learning algorithms based on conditional independencies. For example, they correspond to step C) of the Causal Inference (CI) algorithm of \cite{Spirtes00}. Importantly, these rules can only be applied because $\mathrm{X}$ and $\mathrm{Z}$ are separable. The sufficient statistics will allow us to formulate analogous rules for cases in which they are not separable.

\section{Structure Learning with Sufficient Statistics}
\label{ss2}

We here present the general formulation of the use of sufficient statistics to augment the inferential power of structure learning algorithms based on conditional independencies. We will introduce counterparts of rules R.c and R.nc that do not require that $\mathrm{X}$ and $\mathrm{Z}$ are separable when conditioning on a set of observable variables, but alternatively use sufficient statistics to separate them. These additional rules can then be added to the set used in any standard structure learning algorithm that uses conditional independencies to infer causal relations in systems possibly containing hidden variables.

\begin{figure*}%[h]
  \begin{center}
    \scalebox{0.5}{\includegraphics*{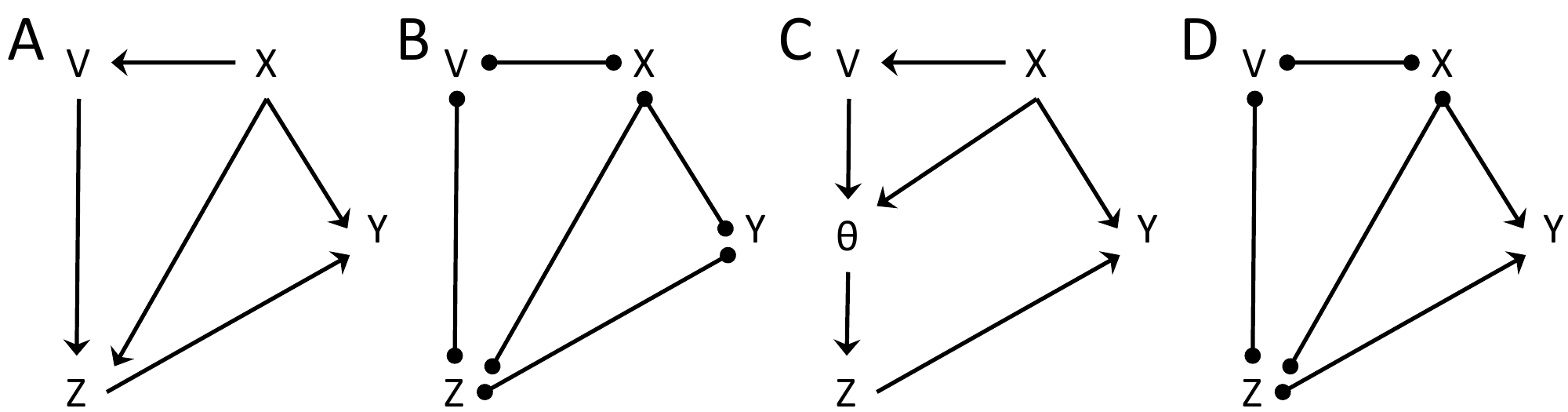}}
  \end{center}
  %\vspace{0.25in}
  \caption{Sufficient statistics provide new causal information by creating new conditional independencies. \textbf{A}) Example of a DAG representing the causal structure of a system. \textbf{B}) Partially oriented graph obtained applying the standard rules of causal orientation. No edge can be oriented, all ends of the edges are undetermined, as denoted by $\bullet\--$. The only conditional independence is $\mathrm{V} \perp \mathrm{Y}| \mathrm{X},\mathrm{Z}$ which using rule R.nc provides the only causal information that can be extracted in this case, namely the presence of the noncolliders $\mathrm{V} \bullet-\underline{\bullet\mathrm{X}\bullet}-\bullet \mathrm{Y}$ and $\mathrm{V} \bullet-\underline{\bullet\mathrm{Z}\bullet}-\bullet \mathrm{Y}$. \textbf{C}) Graph with the same causal structure of the DAG in A) but explicitly representing an underlying sufficient statistic $\theta$ embodied in the functional equation of $\mathrm{Z}$. \textbf{D}) Partially oriented graph obtained when inferring the causal structure in A) complementing the standard rules of causal orientation with causal knowledge that can be learned from the new conditional independence $\mathrm{Z} \perp \mathrm{X} \mathrm{V}| \theta$ created by the sufficient statistic. In comparison to B), the collider $\mathrm{Z} \bullet\rightarrow \mathrm{Y} \leftarrow\bullet \mathrm{X}$ is identified.}
  \label{f0}
\end{figure*}

We start examining a concrete example of the use of sufficient statistics for structure learning. Figure \ref{f0}A shows a causal structure for which the standard orientation rules do not allow orienting any edge (Figure \ref{f0}B). The only independence is $\mathrm{V} \perp \mathrm{Y}| \mathrm{X},\mathrm{Z}$ which using rule R.nc determines the noncolliders $\mathrm{V} \bullet-\underline{\bullet\mathrm{X}\bullet}-\bullet \mathrm{Y}$ and $\mathrm{V} \bullet-\underline{\bullet\mathrm{Z}\bullet}-\bullet \mathrm{Y}$. Consider now that in the functional equation $f_\mathrm{z}(\mathrm{X},\mathrm{V}, \varepsilon_\mathrm{z})$, the role of $\mathrm{X}$ and $\mathrm{V}$ can be jointly captured by a single function $\theta = g(\mathrm{X}, \mathrm{V})$, that is, $f_\mathrm{z}(\mathrm{X},\mathrm{V},\varepsilon_\mathrm{z})$ can be expressed as $f_\mathrm{z}(\theta, \varepsilon_\mathrm{z})$. For example, $g(\mathrm{X}, \mathrm{V})$ could be $a\mathrm{X} +b\mathrm{V}$, $a\mathrm{X} -b\mathrm{V}$ \--with arbitrary coefficients $a,b$\--, could be $\mathrm{X}\cdot\mathrm{V}$, $\mathrm{X}/\mathrm{V}$, or more complicate functions, e.g. $\cos(\mathrm{X})+ \exp(\mathrm{V}^2)$. The key point is that \--in this case both $\mathrm{X}$ and $\mathrm{V}$\--, only determine $\mathrm{Z}$ through $\theta$. The function $f_\mathrm{z}$ can have an arbitrarily complex form as a function of $\theta$, but once a value $\theta = \theta_0$ is fixed, $\mathrm{Z}$ becomes independent of $\mathrm{X}$ and $\mathrm{V}$, that is, $\theta$ is a sufficient statistic for $\mathrm{X}$ and $\mathrm{V}$. In Figure \ref{f0}C we graphically represent the sufficient statistic. Below we will explain which is in general the relation between a graph $G$ representing the causal structure of the variables of a system and an augmented graph $G^+_{\theta}$ representing also the sufficient statistics present in the system. We will use the graphical representations of the statistics to visualize the conditional independencies they create. In particular, in Figure \ref{f0}C, $\mathrm{Z}$ is d-separated from both $\mathrm{X}$ and $\mathrm{V}$ given $\theta$. As we will see, a rule analogous to rule R.c can exploit the independence $\mathrm{Z} \perp \mathrm{X}| \theta$ to orient the collider $\mathrm{Z} \bullet\rightarrow \mathrm{Y} \leftarrow\bullet \mathrm{X}$ (Figure \ref{f0}D). We now define when, within the generative functional equation of a variable, a sufficient statistic exists for one of its parents.

\vspace*{1mm}
\noindent \textbf{Definition 1}\ \textbf{Sufficient statistic in a functional equation}: \emph{A functional sufficient statistic $\theta_z(\mathrm{X}; \tilde{\mathbf{V}})$ for $\mathrm{X} \in \mathbf{Pa}_z$ exists if there is a set $\tilde{\mathbf{V} }= \{\tilde{\mathrm{V}}_1,...,\tilde{\mathrm{V}}_m\} \subset \mathbf{Pa}_z$ such that a function $\theta_z = g(\mathrm{X}, \tilde{\mathbf{V}})$ exists that allows reparameterizing the functional equation $\mathrm{Z}:= f_z(\mathbf{Pa}_z, \varepsilon_z)$ to $\mathrm{Z}:= f_z(\mathbf{Pa}_z \backslash \mathrm{X}, \theta_z(\mathrm{X}; \tilde{\mathbf{V}}), \varepsilon_z)$.}

\vspace*{1mm}
The sufficient statistic is a function $g(\mathrm{X}, \tilde{\mathbf{V}})$ of some parents of $\mathrm{Z}$, including $\mathrm{X}$, embedded within the functional equation of $\mathrm{Z}$. Trivially, $\mathrm{X}$ is a sufficient statistic for itself if it is a parent of $\mathrm{Z}$. However, in this case and generally if the relation between the statistic and $\mathrm{X}$ is invertible, the statistic will not be useful to create a conditional independence between $\mathrm{X}$ and $\mathrm{Z}$ when conditioning on it, since the entropy $H(\mathrm{X}|\theta_z)$ is zero. In Section \ref{ss3} we will discuss criteria to select sufficient statistics useful for structure learning.

\begin{figure*}[t]
  \begin{center}
    \scalebox{0.34}{\includegraphics*{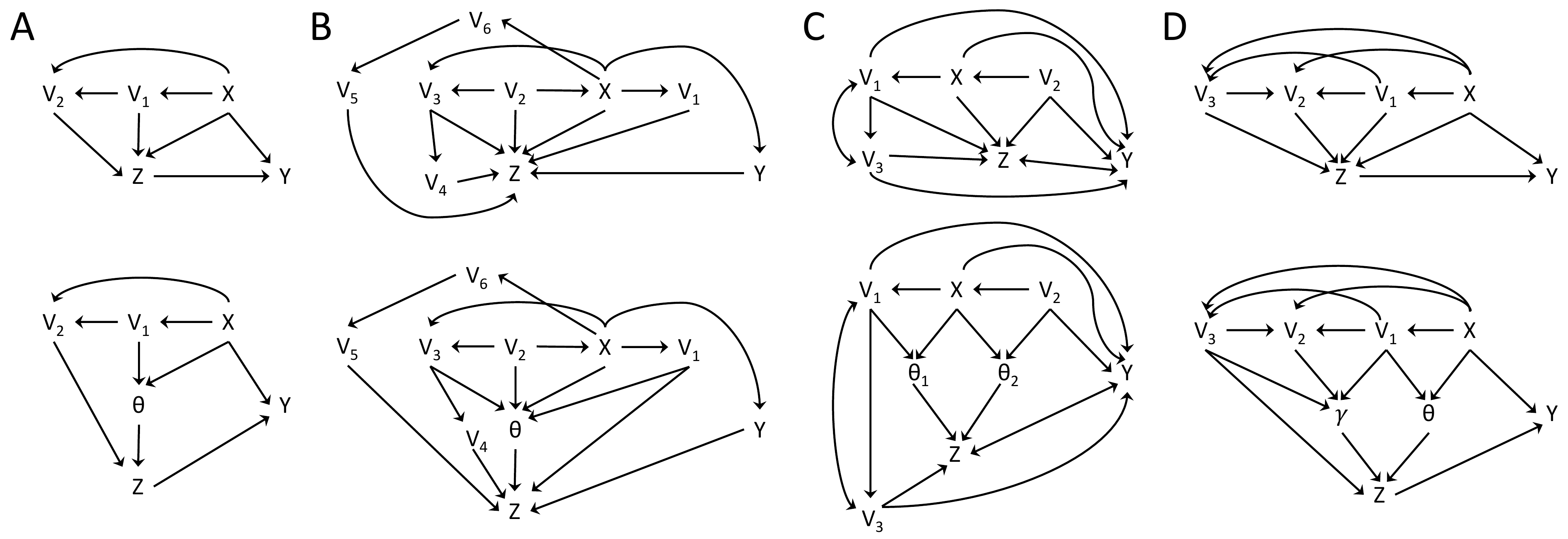}}
  \end{center}
  %\vspace{0.25in}
  \caption{Structure learning with sufficient statistics. Examples of causal structures in which $\mathrm{X}$ and $\mathrm{Z}$ are nonseparable and the standard rules of causal orientation cannot determine whether $\mathrm{Y}$ is a collider or a noncollider in $\mathrm{X} \-- \mathrm{Y} \-- \mathrm{Z}$, while the existence of sufficient statistics provides this additional causal information. In each column the upper graph $G$ is a standard graph representing the causal structure of the system. The lower graph $G^+_{\theta}$ also represents the functional sufficient statistics present in the systems (see text for details on the relation between $G$ and $G^+_{\theta}$). \textbf{A-B}) Systems with a functional sufficient statistic $\theta$ for $\mathrm{X}$ in the functional equation of $\mathrm{Z}$. \textbf{C}) System with a functional statistics' sufficient set for $\mathrm{X}$ in the equation of $\mathrm{Z}$, formed by two functional statistics. \textbf{D}) System in which, apart from a functional statistic $\theta$ for $\mathrm{X}$ in the functional equation of $\mathrm{Z}$, there is also an auxiliary functional statistic $\gamma$ that reduces the conditioning set $\mathbf{S}$ needed to create an independence. See main text for definitions and further explanations.}
  \label{f1}
\end{figure*}

Figure \ref{f1}A-B show two more examples of systems with a sufficient statistic for $\mathrm{X}$ in the functional equation of $\mathrm{Z}$. In both columns, the graph $G$ on the top corresponds to a standard DAG representing the causal structure of the system. The bottom graph provides a graphical representation of the functional sufficient statistics using an augmented graph $G^+_{\theta}$ which incorporates the structure of the statistics to the graph. We will use this type of graphs only for visualization of the statistics and of the independencies they create. The purpose is still to infer a partially oriented graph associated with the causal structure of the observable variables. A graph $G^+_{\theta}$ associated with a graph $G$ has the same causal structure as $G$, except that it explicitly represents a sufficient statistic $\theta$ as a node with incoming arrows from all the variables corresponding to the arguments in the subfunction that defines it, and an outgoing arrow to the variable (or variables) in which functional equation the statistic is embedded. Therefore, the parenthood structure of $G^+_{\theta}$ is the same as the one of $G$, except that each node has also as parents the sufficient statistics embodied in its functional equation, while those variables that only appear in the functional equation of another through sufficient statistics lose their parenthood status. This means that by construction d-separation graphically represents also the conditional independencies created conditioning on the sufficient statistics. In Appendix A we formalize the concept of causal models with sufficient statistics and we formalize the connection between a graph $G$ and the augmented graph $G^+_{\theta}$ that also represents sufficient statistics (Definition S4). If not stated otherwise, we will refer to adjacency and parenthood relations as determined in $G$, and not in the associated $G^+_{\theta}$. The same procedure to construct $G^+_{\theta}$ is applicable independently of whether $\mathrm{G}$ is a DAG containing only directed arrows or an IPG, in which hidden variables are not represented and their presence is represented with bidirected arrows. Similarly, given that for any DAG there is a unique MAG that represents the conditional independencies and causal relations embodied in the DAG \citep{Richardson2002}, the augmented graphs can be constructed for MAGs with the same procedure.

In Figure \ref{f1}A, $\mathrm{X}$ only determines $\mathrm{Z}$ through $\theta = g(\mathrm{X}, \mathrm{V}_1)$. Similarly, in Figure \ref{f1}B, $\theta = g(\mathrm{X}, \mathrm{V}_1, \mathrm{V}_2, \mathrm{V}_3)$. In more general cases, it may be needed more than one functional statistic to capture the effect of $\mathrm{X}$ on $\mathrm{Z}$. We therefore extend the definition to a sufficient set of statistics.

\vspace*{1mm}
\noindent \textbf{Definition 2}\ \textbf{Sufficient set of statistics in a functional equation}: \emph{In the functional equation of $\mathrm{Z}$, there is a sufficient set of K functional statistics (fss-set) $\Theta_z(\mathrm{X}; \alpha) = \{\theta_{z1}(\mathrm{X}; \tilde{\mathbf{V}}_1),...,\theta_{zK}(\mathrm{X}; \tilde{\mathbf{V}}_K) \}$ for $\mathrm{X} \in \mathbf{Pa}_z$ if there is a set $\alpha = \{\tilde{\mathbf{V}}_1,...,\tilde{\mathbf{V}}_K\}$ such that a set of functions $\theta_{zi} = g_i(\mathrm{X}, \tilde{\mathbf{V}}_i)$ $\forall \theta_{zi} \in \Theta_z(\mathrm{X}; \alpha)$ exists, with $\tilde{\mathbf{V}_i }= \{\tilde{\mathrm{V}}_1,...\tilde{\mathrm{V}}_{m_i}\} \subset \mathbf{Pa}_z$, which allow reparameterizing  $f_z(\mathbf{Pa}_z, \varepsilon_z)$ to $f_z(\mathbf{Pa}_z \backslash \mathrm{X}, \Theta_z(\mathrm{X}; \alpha) , \varepsilon_z)$.}

Figure \ref{f1}C shows a fss-set $\Theta_z(\mathrm{X}; \alpha) = \{ \theta_1, \theta_2\}$, with
$\tilde{\mathbf{V}}_1 = \mathrm{V}_1$ and $\tilde{\mathbf{V}}_2 = \mathrm{V}_2$. Importantly, a fss-set $\Theta_z(\mathrm{X}; \alpha)$ creates a new conditional independence, separating $\mathrm{Z}$ from $\mathrm{X}$.

\vspace*{1mm}
\noindent \textbf{Proposition 3}\ \textbf{Conditional independence with functional sufficient statistics}: \emph{If the functional equation of $\mathrm{Z}$ has an fss-set $\Theta_z(\mathrm{X}; \alpha)$ for $\mathrm{X} \in \mathbf{Pa}_z$ there is at least one set $\mathbf{S}$ disjoint to $\{\mathrm{X}, \mathrm{Z}\}$ such that for $\mathbf{S}_{\Theta_z} = \{ \Theta_z(\mathrm{X};\alpha), \mathbf{S}\}$, $\mathrm{Z} \perp \mathrm{X} | \mathbf{S}_{\Theta_z}$.}

\vspace*{1mm}
\noindent Proof: By definition of a sufficient set of functional statistics, $\mathrm{Z}$ and $\mathrm{X}$ are conditionally independent given $\mathbf{S}_{\Theta_z} = \{ \Theta_z(\mathrm{X};\alpha), \mathbf{Pa}_z \backslash \mathrm{X}\}$.\ \ $\Box$

Given the existence of the fss-set $\Theta_z(\mathrm{X}; \alpha)$ for $\mathrm{X}$ in the functional equation of $\mathrm{Z}$, in the augmented graph $G^+_{\theta}$ node $\mathrm{X}$ is not anymore a parent of node $\mathrm{Z}$, and the new set of parents comprises $\Theta_z(\mathrm{X};\alpha)$. As described above, the parenthood structure of $G^+_{\theta}$ is constructed so that, despite the deterministic relations defining the sufficient statistics, d-separation can also be used to read the new conditional independencies created when conditioning on sets comprising sufficient statistics. This is in contrast to the case in which deterministic relations exist between the variables of the system, which generally requires an extended criterion of graphical separability \citep{Geiger90}. Like with standard algorithms such as the FCI, we assume that no deterministic relations exist between the variables in the system, that is, that the only deterministic relations are the ones defining the sufficient statistics. See Appendix A for a comparison of how deterministic relations between variables of the system or in the definition of the statistics affect the connection between conditional independencies and criteria of graphical separability.
%This change in the parenthood structure when introducing the statistics is just a consequence of the general fact that parenthood is always relative to the set of nodes included in the graph, and any link $\mathrm{V}_i \rightarrow \mathrm{V}_j$ may correspond to an underlying structure $\mathrm{V}_i \rightarrow \mathrm{U} \rightarrow \mathrm{V}_j$, where $\mathrm{U}$ is hidden.
The existence of a sufficient set of statistics in general does not only create the conditional independence corresponding to using $\{ \Theta_z(\mathrm{X};\alpha), \mathbf{Pa}_z \backslash \mathrm{X}\}$ to separate $\mathrm{Z}$ and $\mathrm{X}$. As we will discuss below, a small conditioning set $\mathbf{S}$ is desirable in order to not constrain $\mathrm{X}$. In Figure \ref{f1}A, $\mathbf{S} = \mathrm{V}_2$ leads to $\mathrm{Z} \perp \mathrm{X} | \theta, \mathrm{V}_2$. In Figure \ref{f1}B, both with $\mathbf{S} = \{\mathrm{Y}, \mathrm{V}_1, \mathrm{V}_4, \mathrm{V}_6\}$ and $\mathbf{S} = \{\mathrm{Y}, \mathrm{V}_1, \mathrm{V}_3, \mathrm{V}_5\}$, the set $\{\theta, \mathbf{S}\}$ separates $\mathrm{Z}$ and $\mathrm{X}$. Similarly $\mathrm{Z} \perp \mathrm{X} | \theta_1, \theta_2, \mathrm{V}_3$ in Figure \ref{f1}C. As seen from these examples, the set $\mathbf{S}$ may include %a subset of arguments of the statistics,
other parents of $\mathrm{Z}$, and also other variables that inactivate paths between $\mathrm{X}$ and $\mathrm{Z}$ (such as $\mathrm{V}_6$ in Figure \ref{f1}B).

The definitions above use knowledge about the functional equation of $\mathrm{Z}$ and its parents which is not directly available from the data. We therefore define a sufficient condition for the existence of a sufficient set of statistics for a pair of variables without requiring any knowledge about the functional structure.

\vspace*{1mm}
\noindent \textbf{Definition 3}\ \textbf{Sufficient set of statistics for a pair of variables}: \emph{Two nonseparable variables $\mathrm{X},\mathrm{Z}$ have a sufficient set of K statistics (ss-set) $\Psi_{x,z}(\alpha) = \{\psi_1(\mathrm{X},\mathrm{Z}; \tilde{\mathbf{V}}_1) ,...,$ $\psi_K(\mathrm{X},\mathrm{Z}; \tilde{\mathbf{V}}_K)\}$ if there is a set $\alpha = \{ \tilde{\mathbf{V}}_1,...,\tilde{\mathbf{V}}_K\}$ such that $\mathrm{X}, \mathrm{Z} \notin \bigcup \alpha$, a set of functions $\psi_i = g_i(\mathrm{W}_i, \tilde{\mathbf{V}}_i)$  $\forall \psi_i \in \Psi_{x,z}(\alpha)$ exists, with $\mathrm{W}_i= \mathrm{Z}\ \forall i$ or $\mathrm{W}_i= \mathrm{X}\ \forall i$, and there is a set $\mathbf{S}$ nonoverlapping with $\{\mathrm{X},\mathrm{Z}\}$ such that $\mathrm{Z} \perp \mathrm{X} | \mathbf{S}_{\Psi_{x,z}}$, where $\mathbf{S}_{\Psi_{x,z}} = \{ \Psi_{x,z}(\alpha), \mathbf{S}\}$.}

\vspace*{1mm}
The union $\bigcup \alpha$ indicates the union of all elements in the sets composing set $\alpha$, that is, $\mathrm{X}, \mathrm{Z} \notin \bigcup \alpha$, indicates that $\forall \tilde{\mathbf{V}}_i \in \alpha \ \mathrm{X}, \mathrm{Z} \notin \tilde{\mathbf{V}}_i $. Definition 3 does not rely on information about the causal structure, and defines the statistics' sufficient set $\Psi_{x,z}(\alpha)$ based on the conditional independence $\mathrm{Z} \perp \mathrm{X} | \mathbf{S}_{\Psi_{x,z}}$ it creates. This means that, to be able to use sufficient sets of statistics for structure learning, we need to extend the standard faithfulness assumption to enforce also an isomorphic relation between conditional independencies that can be created inferring sufficient sets of statistics from the distribution $p(\mathbf{V})$ of observable variables and the form of underlying functional sufficient statistics existing in the system. Accordingly, if a graph $G$ is used to represent the causal structure of a system and an augmented graph $G^+_{\theta}$ is built as mentioned above to represent the sufficient statistics existing in the system, the extended faithfulness assumption serves to guarantee that $\mathrm{X} \perp \mathrm{Z}| \mathbf{S}_{\Psi_{x,z}}$ is embodied in $p(\mathbf{V})$ if and only if the causal structure and the form of the sufficient statistics embodied in the functional equations is consistent with the form of $\Psi_{x,z}(\alpha)$. See Appendix A for a formalization of this extended faithfulness assumption (Definition S6).

Under the extended faithfulness assumption, $\Psi_{x,z}(\alpha)$ corresponds to an underlying sufficient set of functional statistics, which may be a fss-set $\Theta_z(\mathrm{X}; \alpha)$ within the functional equation of $\mathrm{Z}$, a fss-set $\Theta_x(\mathrm{Z}; \alpha)$ within the functional equation of $\mathrm{X}$, or also may reflect the presence of statistics in the functional equation of an intermediate variable that is both a collider and a noncollider in paths between $\mathrm{X}$ and $\mathrm{Z}$, rendering them nonseparable. For example, in a system with $\mathrm{X} \rightarrow \mathrm{V} \rightarrow \mathrm{Z}$ and $\mathrm{V} \leftrightarrow \mathrm{Z}$,  conditioning on $\mathrm{V}$ to inactivate the directed path in which it is a noncollider activates the path through the bidirected arc, in which it is a collider, rendering $\mathrm{X}$ and $\mathrm{Z}$ nonseparable. Sufficient statistics embodied in the functional equation of $\mathrm{V}$ would allow inactivating the causal path without activating the collider in $\mathrm{V}$. Furthermore, as we will see in Section \ref{ss5}, in the presence of selection bias a sufficient set of statistics may exist also in the functional equation of a hidden variable which is conditioned.

Regardless of within which functional equations the statistics are located, the utility of $\Psi_{x,z}(\alpha)$ lies in separating $\mathrm{Z}$ and $\mathrm{X}$ with $\mathbf{S}_{\Psi_{x,z}}$. Under the extended faithfulness assumption that guarantees an isomorphic relation between conditional independencies and both the causal structure of the observable variables and the structure of existing sufficient statistics, the new conditional independencies created by the statistics allow extending the rules of inference to the case in which the variables are nonseparable without sufficient statistics.

\vspace*{1mm}
\noindent \textbf{Rule $\mathbf{R.c-ss}$}\ \ \textbf{Inference of a collider with sufficient statistics}: \emph{Consider variables $\mathrm{X}$, $\mathrm{Y}$, $\mathrm{Z}$, all nonseparable. Find a sufficient set of statistics $\Psi_{x,z}(\alpha)$ and a set $\mathbf{S}$ nonoverlapping with $\{ \mathrm{X}, \mathrm{Y}, \mathrm{Z}\}$ such that $\mathrm{Y} \notin \bigcup \alpha$ and, for $\mathbf{S}_{\Psi_{x,z}} = \{\mathbf{S}, \Psi_{x,z}(\alpha)\}$, it holds that $\mathrm{X} \notperp \mathrm{Y} |\mathbf{S}_{\Psi_{x,z}}$, $\mathrm{Y} \notperp \mathrm{Z} |\mathbf{S}_{\Psi_{x,z}}$, and $\mathrm{X} \perp \mathrm{Z} |\mathbf{S}_{\Psi_{x,z}}$, then orient $\mathrm{X} *\--*\mathrm{Y} *\--* \mathrm{Z}$ as $\mathrm{X} *\rightarrow \mathrm{Y} \leftarrow* \mathrm{Z}$.}

\vspace*{1mm}
\noindent Proof: Conditions $\mathrm{X} \notperp \mathrm{Y} |\mathbf{S}_{\Psi_{x,z}}$ and $\mathrm{Y} \notperp \mathrm{Z} |\mathbf{S}_{\Psi_{x,z}}$ ensure that $\mathrm{X}$ and $\mathrm{Z}$ are still nonseparable from $\mathrm{Y}$ with a conditioning set $\mathbf{S}_{\Psi_{x,z}}$ that apart from the observable variables in $\mathbf{S}$ also includes $\Psi_{x,z}(\alpha)$. This discards that the set $\Psi_{x,z}(\alpha)$ does include not only a sufficient set of statistics to separate $\mathrm{X}$ and $\mathrm{Z}$, but also statistics that separate $\mathrm{X}$ or $\mathrm{Z}$ from $\mathrm{Y}$. With the nonseparability of $\mathrm{X}$ and $\mathrm{Z}$ from $\mathrm{Y}$ preserved, the conditional independence $\mathrm{X} \perp \mathrm{Z} |\mathbf{S}_{\Psi_{x,z}}$ is only compatible with $\mathrm{Y}$ being a collider between $\mathrm{X}$ and $\mathrm{Z}$, since $\mathrm{Y} \notin \bigcup \alpha$ and $\mathbf{S}$ does not include $\mathrm{Y}$, and hence $\mathrm{Y}$ is not necessary to avoid that the dependencies between $\mathrm{X}$ and $\mathrm{Y}$ and between $\mathrm{Y}$ and $\mathrm{Z}$ result in a dependence between $\mathrm{X}$ and $\mathrm{Z}$. $\ \ \Box$
%si fuera sufficiente condicionar en un descendant statistic tambien lo seria condicionar en Y
%or on a statistic involving $\mathrm{Y}$ is not necessary to avoid that the dependencies between $\mathrm{X}$ and $\mathrm{Y}$ and between $\mathrm{Y}$ and $\mathrm{Z}$ result in a dependence between $\mathrm{X}$ and $\mathrm{Z}$, as reflected in the fact the set $\mathbf{S}_{\Psi_{x,z}}$ does not include $\mathrm{Y}$ and $\mathrm{Y}$ is not an argument of any of the statistics in $\Psi_{x,z}(\alpha)$. $\ \ \Box$

\vspace*{1mm}
\noindent \textbf{Rule $\mathbf{R.nc-ss}$}\ \textbf{Inference of a noncollider with sufficient statistics}: \emph{Consider variables $\mathrm{X}$, $\mathrm{Y}$, $\mathrm{Z}$, all nonseparable. Find a sufficient set of statistics $\Psi_{x,z}(\alpha)$ and a set $\mathbf{S}_0$ nonoverlapping with $\{ \mathrm{X}, \mathrm{Y}, \mathrm{Z}\}$ such that, for $\mathbf{S} = \{ \mathbf{S}_0, \mathrm{Y}\}$, $\mathbf{S}_{\Psi_{x,z}} = \{\mathbf{S}, \Psi_{x,z}(\alpha) \}$, and $\mathbf{\Psi}^{(y)}_{x,z}(\alpha)$ the subset of statistics in $\Psi_{x,z}(\alpha)$ which have $\mathrm{Y}$ as an argument, it holds that $\mathrm{X} \notperp \mathrm{Y} |\mathbf{S}'$,  $\mathrm{Y} \notperp \mathrm{Z} |\mathbf{S}'$, $\forall \mathbf{S}' \subseteq \mathbf{S}_{\Psi_{x,z}}  \backslash \{\mathrm{Y}, \mathbf{\Psi}^{(y)}_{x,z}(\alpha) \}$, $\mathrm{X} \notperp \mathrm{Z} |\mathbf{S}_{\Psi_{x,z}}  \backslash \{\mathrm{Y}, \mathbf{\Psi}^{(y)}_{x,z}(\alpha) \}$, and at least $\mathrm{X} \perp \mathrm{Z} |\mathbf{S}_{\Psi_{x,z}}$ or $\mathrm{X} \perp \mathrm{Z} |\mathbf{S}_{\Psi_{x,z}} \backslash \mathrm{Y}$, then mark $\mathrm{X} *\--*\mathrm{Y} *\--* \mathrm{Z}$ as $\mathrm{X} *-\underline{*\mathrm{Y}*}-* \mathrm{Z}$.}

\vspace*{1mm}
\noindent Proof: The combination of $\mathrm{X} \notperp \mathrm{Z} |\mathbf{S}_{\Psi_{x,z}}  \backslash \{\mathrm{Y},  \mathbf{\Psi}^{(y)}_{x,z}(\alpha)\}$ and at least $\mathrm{X} \perp \mathrm{Z} |\mathbf{S}_{\Psi_{x,z}}$ or $\mathrm{X} \perp \mathrm{Z} |\mathbf{S}_{\Psi_{x,z}} \backslash \mathrm{Y}$ indicates that it is the fact of further adding $\{\mathrm{Y}, \mathbf{\Psi}^{(y)}_{x,z}(\alpha)\}$ or $\mathbf{\Psi}^{(y)}_{x,z}(\alpha)$ to the conditioning set what creates the independence. This means that at least $\mathrm{Y}$ or a sufficient statistic having it as an argument is required to inactivate paths that were leading to $\mathrm{X} \notperp \mathrm{Z} |\mathbf{S}_{\Psi_{x,z}}  \backslash \{\mathrm{Y}, \mathbf{\Psi}^{(y)}_{x,z}(\alpha) \}$, and hence it has to be a noncollider in those paths, since only conditioning on noncolliders deactivates dependencies. The fact that $\mathrm{X} \notperp \mathrm{Y} |\mathbf{S}'$,  $\mathrm{Y} \notperp \mathrm{Z} |\mathbf{S}'$, $\forall \mathbf{S}' \subseteq \mathbf{S}_{\Psi_{x,z}}  \backslash \{\mathrm{Y}, \mathbf{\Psi}^{(y)}_{x,z}(\alpha) \}$ discards that the use of sufficient statistics included in $\mathbf{S}_{\Psi_{x,z}}  \backslash \{\mathrm{Y}, \mathbf{\Psi}^{(y)}_{x,z}(\alpha) \}$ allows inactivating all the paths that were creating the adjacency $\mathrm{X}\--\mathrm{Y}$ or $\mathrm{Y}\--\mathrm{Z}$ when no statistics were used. This discards that there is a set $\mathbf{S}^* \subset \mathbf{S}_{\Psi_{x,z}}  \backslash \{\mathrm{Y}, \mathbf{\Psi}^{(y)}_{x,z}(\alpha) \}$ for which at least $\mathrm{X} \perp \mathrm{Y} |\mathbf{S}^*$ or $\mathrm{Y} \perp \mathrm{Z} |\mathbf{S}^* $, and that it is only because of further adding to $\mathbf{S}^*$ some additional variables $\mathrm{V}^* \in \mathbf{S}_{\Psi_{x,z}}  \backslash \{\mathrm{Y}, \mathbf{\Psi}^{(y)}_{x,z}(\alpha), \mathbf{S}^* \}$ that a dependence $\mathrm{X} \notperp \mathrm{Y} |\mathbf{S}_{\Psi_{x,z}}  \backslash \{\mathrm{Y}, \mathbf{\Psi}^{(y)}_{x,z}(\alpha) \}$ or $\mathrm{Y} \notperp \mathrm{Z} |\mathbf{S}_{\Psi_{x,z}}  \backslash \{\mathrm{Y}, \mathbf{\Psi}^{(y)}_{x,z}(\alpha) \}$ exists, because of variables in $\mathrm{V}^*$ activating some paths in which they are colliders. This discards that conditioning on $\{\mathrm{Y}, \mathbf{\Psi}^{(y)}_{x,z}(\alpha) \}$ or $\mathbf{\Psi}^{(y)}_{x,z}(\alpha)$ creates the independence $\mathrm{X} \perp \mathrm{Z} |\mathbf{S}_{\Psi_{x,z}}$ or $\mathrm{X} \perp \mathrm{Z} |\mathbf{S}_{\Psi_{x,z}} \backslash \mathrm{Y}$ from $\mathrm{X} \notperp \mathrm{Z} |\mathbf{S}_{\Psi_{x,z}}  \backslash \{\mathrm{Y},  \mathbf{\Psi}^{(y)}_{x,z}(\alpha)\}$ only because inactivating the paths activated by conditioning on $\mathrm{V}^*$. The fact that the use of sufficient statistics included in $\mathbf{S}_{\Psi_{x,z}}  \backslash \{\mathrm{Y}, \mathbf{\Psi}^{(y)}_{x,z}(\alpha) \}$ does not inactivate all the paths that were creating the adjacency $\mathrm{X}\--\mathrm{Y}$ or $\mathrm{Y}\--\mathrm{Z}$ when no statistics were used, also ensures that it cannot exist a collider in $\mathrm{Y}$ \--activated when conditioning on $\mathrm{Y}$ or $\mathbf{\Psi}^{(y)}_{x,z}(\alpha)$\--, which does not create a dependence between $\mathrm{X}$ and $\mathrm{Z}$ only because conditioning on the sufficient statistics in $\mathbf{S}_{\Psi_{x,z}}  \backslash \{\mathrm{Y}, \mathbf{\Psi}^{(y)}_{x,z}(\alpha) \}$ inactivated all paths between $\mathrm{Y}$ and $\mathrm{X}$ or between $\mathrm{Y}$ and $\mathrm{Z}$ incoming to $\mathrm{Y}$. This means that not only $\mathrm{Y}$ has to be a noncollider in some path corresponding to $\mathrm{X} *\--*\mathrm{Y} *\--* \mathrm{Z}$, but also that it cannot be a collider in any of those paths, meaning that it has to be a noncollider in all paths corresponding to $\mathrm{X} *\--*\mathrm{Y} *\--* \mathrm{Z}$. $\ \ \Box$

These new rules are analogous to R.c and R.nc, but use the sufficient statistics to obtain a conditional independence between $\mathrm{X}$ and $\mathrm{Z}$ even when $\mathrm{X}$ and $\mathrm{Z}$ are nonseparable, that is, when a conditional independence cannot be obtained conditioning only on observable variables. As mentioned above, even if the new orientation rules exploit sufficient statistics, the purpose is, like with the standard rules, to determine the partially oriented graph representing inferred causal relationships between the observable variables. In more detail, in R.c-ss the condition $\mathrm{X} \perp \mathrm{Z} |\mathbf{S}_{\Psi_{x,z}}$ is analogous to $\mathrm{X} \perp \mathrm{Z} |\mathbf{S}$ in R.c, and the conditions $\mathrm{X} \notperp \mathrm{Y} |\mathbf{S}_{\Psi_{x,z}}$ and $\mathrm{Y} \notperp \mathrm{Z} |\mathbf{S}_{\Psi_{x,z}}$ play the role that in R.c plays the nonseparability of $\mathrm{Y}$ with $\mathrm{X}$ and $\mathrm{Z}$, which already guarantees that $\mathrm{X} \notperp \mathrm{Y} |\mathbf{S}$ and $\mathrm{Y} \notperp \mathrm{Z} |\mathbf{S}$. In R.nc-ss, $\mathrm{X} \notperp \mathrm{Z} |\mathbf{S}_{\Psi_{x,z}} \backslash \{\mathrm{Y}, \mathbf{\Psi}^{(y)}_{x,z}(\alpha) \}$, and at least $\mathrm{X} \perp \mathrm{Z} |\mathbf{S}_{\Psi_{x,z}}$ or $\mathrm{X} \perp \mathrm{Z} |\mathbf{S}_{\Psi_{x,z}} \backslash \mathrm{Y} $ are analogous to $\mathrm{X} \notperp \mathrm{Z} |\mathbf{S}$, and $\mathrm{X} \perp \mathrm{Z} |\mathbf{S}, \mathrm{Y}$ in R.nc, and $\mathrm{X} \notperp \mathrm{Y} |\mathbf{S}'$,  $\mathrm{Y} \notperp \mathrm{Z} |\mathbf{S}'$, $\forall \mathbf{S}' \subseteq \mathbf{S}_{\Psi_{x,z}}  \backslash \{\mathrm{Y}, \mathbf{\Psi}^{(y)}_{x,z}(\alpha) \}$ is already guaranteed in the standard rule by the nonseparability of $\mathrm{Y}$ with $\mathrm{X}$ and $\mathrm{Z}$. To be able to apply these rules it is required that the dependencies and independencies involved in the rules can be evaluated, and hence that the conditional entropies of the variables when conditioning on the sets including sufficient statistics are nonzero, that is, that some uncertainty remains after conditioning.

In all examples of Figure \ref{f1} $\mathrm{X}$ and $\mathrm{Z}$ are nonseparable, and hence rules R.c and R.nc are not applicable. In fact, for all these examples a standard algorithm such as the FCI \citep{Spirtes00} cannot determine whether $\mathrm{Y}$ is a collider or a noncollider in $\mathrm{X} \-- \mathrm{Y} \-- \mathrm{Z}$. Conversely, in Figure \ref{f1}A, rule R.c-ss determines that $\mathrm{Y}$ is a collider using that $\mathrm{X} \perp \mathrm{Z}| \theta, \mathrm{V}_2$, that is, $\mathbf{S}_{\Psi_{x,z}} = \{ \theta, \mathrm{V}_2 \}$. Similarly, in Figure \ref{f1}C rule R.c-ss can be applied with $\mathbf{S}_{\Psi_{x,z}} = \{ \theta_1, \theta_2, \mathrm{V}_3 \}$. In Figure \ref{f1}B, rule R.nc-ss can be applied with $\mathbf{S}_{\Psi_{x,z}} = \{ \mathrm{Y}, \theta, \mathrm{V}_1, \mathrm{V}_4, \mathrm{V}_6 \}$.

Note that, despite providing a sufficient condition for the existence of a sufficient set of statistics, Definition 3 is limited by the requirement that either $\mathrm{X}$ or $\mathrm{Z}$ appears as an argument in the functions $\psi_i = g_i(\mathrm{W}_i, \tilde{\mathbf{V}}_i)$ defining the statistics. More generally, sufficient statistics that do not have any of the two variables as an argument can also contribute to create a new conditional independence between $\mathrm{X}$ and $\mathrm{Z}$. We call this type of additional statistics \emph{auxiliary} statistics. Figure \ref{f1}D shows an example of an auxiliary statistic. Consider that in the system only the functional sufficient statistic $\theta = g_{\theta}(\mathrm{X}, \mathrm{V}_1)$ existed. Conditioning on $\theta$ inactivates the direct path from $\mathrm{X}$ to $\mathrm{Z}$, but not the indirect path from $\mathrm{X}$ to $\mathrm{Z}$ through $\mathrm{V}_1$, so $\mathrm{V}_1$ would need to be conditioned. However, conditioning on $\mathrm{V}_1$, the deterministic constraint $\theta_0 = g_\theta(\mathrm{X}, \mathrm{v}_1)$, for fixed values $\theta= \theta_0$ and $\mathrm{V}_1 = \mathrm{v}_1$, may result in a small space of solutions for $\mathrm{X}$, or even a unique value of $\mathrm{X}$ if $g_{\theta}(\mathrm{X}, \mathrm{V}_1 = \mathrm{v}_1)$ is invertible. A small or null entropy $H(\mathrm{X}| \theta, \mathrm{V}_1)$ would complicate or impede the evaluation of the dependencies and independencies as required in the new rules. The auxiliary statistic $\gamma = g_\gamma(\mathrm{V}_1,\mathrm{V}_2,\mathrm{V}_3)$ allows weakening the constraint imposed by conditioning on $\theta, \mathrm{V}_1$. This is because, using $\gamma$, the set $\mathbf{S}_{\Psi_{x,z}} = \{ \theta, \mathrm{V}_1, \mathrm{V}_2, \mathrm{V}_3 \}$ can be replaced by $\mathbf{S}_{\Psi_{x,z}} = \{  \theta, \gamma, \mathrm{V}_3\}$. The constraint $\gamma_0 = g_\gamma(\mathrm{V}_1,\mathrm{V}_2, \mathrm{v}_3)$ may be compatible with a higher number of solutions for $\mathrm{V}_1$, which in turn may increase the number of solutions for $\mathrm{X}$ in $\theta_0 = g_\theta(\mathrm{X}, \mathrm{V}_1)$.

In general, an auxiliary set of statistics can allow using a set $\mathbf{S}_{\Psi_{x,z}}$ that increases the entropy $H(\mathrm{X} | \mathbf{S}_{\Psi_{x,z}})$, hence facilitating, or even enabling, the application of the rules.  As we will explain and exemplify below, the IB method is powerful enough to identify sufficient statistics without any \emph{a priori} assumption of whether they contain $\mathrm{X}$ or $\mathrm{Z}$ as an argument or they are auxiliary statistics. For simplicity, we leave the formal definition of a sufficient set of statistics that includes auxiliary statistics for Appendix B.

%However, in the limit of $\mathbf{S}$ containing all the variables $\tilde{\mathbf{V}}_k$ of a certain subfunction $\psi_k = g_k(\mathrm{X}, \tilde{\mathbf{V}}_k)$, the constraint $\psi_{k0} = g_k(\mathrm{X}, \tilde{\mathbf{v}}_{k0})$ for fixed values $\psi_{k} = \psi_{k0}$ and $\tilde{\mathbf{V}}_{k} = \tilde{\mathbf{v}}_{k0}$, may have a single solution for $\mathrm{X}$. In that case, the fact that the entropy $H(\mathrm{X} | \mathbf{S}_{\Psi_{x,z}})=0$ is incompatible with evaluating the dependencies and independencies contained in the rules.

We have here introduced new orientation rules to extract causal information from conditional independencies created by sufficient sets of statistics. Importantly, these additional rules are not to be used in isolation, as an alternative to the standard set of orientation rules. On the contrary, the new rules are to be inserted within the algorithms already implementing the standard rules, such that in the presence of sufficient statistics additional information about the causal structure is inferred, while in the lack of sufficient statistics the standard Markov equivalence class is identified. To illustrate this, in Appendix C we show in detail how to insert R.c-ss and R.nc-ss in a particular standard algorithm, such as the Causal Inference (CI) algorithm of \cite{Spirtes00}. The new rules could be alternatively incorporated to any of the refined algorithms proposed to improve the implementation of the standard rules \citep[see][for a review]{Drton17,Heinze17,Petersbook,Malinsky18,Glymour19}. In Appendix C we also further introduce a counterpart based on sufficient statistics for another standard rule, and we discuss the synergies created by the combination of the standard and new rules. In the rest of the main article we will focus on how the IB method can be used to infer sufficient sets of statistics to implement rules R.c-ss and R.nc-ss.

\section{Identification and selection of sufficient statistics}
\label{ss3}

We now address the question of how to identify and select sufficient sets of statistics with the IB method. Despite the fact that the concept of sufficient set of statistics and the rules introduced above are valid also for continuous variables, we will from now on focus on the original implementation of the IB method by \cite{Tishby99}, which works for discrete variables. In Appendix F we briefly discuss model-based approaches alternative to the IB method.

\subsection{Identification of sufficient statistics with the information bottleneck method}
\label{ss32}

%We here describe the identification of sufficient statistics with the information bottleneck method \citep{Tishby99}.
Traditionally, in the context of estimation theory, a sufficient statistic $\mathrm{T}(\mathbf{X})$ is conceived as a function of the sampled data $\mathbf{X}$ which contains all the information of those data to estimate an underlying parameter $\mu$ of the generative model \citep{Casella02}. That is, the sufficient statistic creates a conditional independence $\mu \perp \mathbf{X} | \mathrm{T}(\mathbf{X})$, or equivalently $p(\mu|\mathrm{T}(\mathbf{X}), \mathbf{X})= p(\mu|\mathrm{T}(\mathbf{X}))$. The IB method generalizes the concept of sufficient statistic from parametric to arbitrary distributions. It formulates the finding of sufficient statistics as a problem of data compression, implemented via a cost function minimization. Consider a possibly multivariate variable $\mathbf{X}$ to be compressed into a lower dimensional variable $\tilde{\mathbf{X}}$, while preserving the information about another target variable $\mathrm{Z}$. \citet{Tishby99} introduced an algorithm to define $\tilde{\mathbf{X}}$ finding the mapping $p(\tilde{\mathbf{X}}|\mathbf{X})$ optimized as

\begin{subequations}
\begin{align}
p(\tilde{\mathbf{X}}|\mathbf{X}) & \equiv \mathrm{argmin}\  I(\tilde{\mathbf{X}};\mathbf{X})-\beta I(\tilde{\mathbf{X}};\mathrm{Z}) \label{e20}\\
 & = \mathrm{argmin}\  I(\tilde{\mathbf{X}};\mathbf{X}) + \beta I(\mathbf{X};\mathrm{Z}|\tilde{\mathbf{X}}), \label{e21}
\end{align}
\end{subequations}
where the parameter $\beta$ determines the tradeoff between compression (low mutual information $I(\tilde{\mathbf{X}};\mathbf{X})$) and information preservation (high $I(\tilde{\mathbf{X}};\mathrm{Z})$, or equivalently low $I(\mathbf{X};\mathrm{Z}|\tilde{\mathbf{X}})$). The IB algorithm proposed by \citet{Tishby99} iteratively updates the projection $p(\tilde{\mathbf{X}}|\mathbf{X})$ according to a set of self-consistent equations for $p(\tilde{\mathbf{X}}|\mathbf{X})$, $p(\mathrm{Z}|\tilde{\mathbf{X}})$, and $p(\tilde{\mathbf{X}})$, analogously to the Blahut-Arimoto algorithm used to estimate compression distortion rates \citep{Cover06}. In each iteration $p(\tilde{\mathbf{X}}|\mathbf{X})$ is iteratively updated according to
\begin{equation}
\label{IBeq}
%\tag{S1}
  p(\tilde{\mathbf{X}}|\mathbf{X}) = \frac{p(\tilde{\mathbf{X}})}{Z(\mathbf{X}, \beta)} \exp \left [ -\beta \mathrm{KL}(p(\mathrm{Z}|\mathbf{X}); p(\mathrm{Z}|\tilde{\mathbf{X}})) \right ],
\end{equation}
where $Z(\mathbf{X}, \beta)$ is a normalization factor and $\mathrm{KL}(p(\mathrm{Z}|\mathbf{X}); p(\mathrm{Z}|\tilde{\mathbf{X}}))$ is the Kullback-Leibler divergence \citep{Kullback1959} between the conditional distribution $p(\mathrm{Z}|\mathbf{X})$ and the distribution $p(\mathrm{Z}|\tilde{\mathbf{X}})$ resulting from $p(\tilde{\mathbf{X}}|\mathbf{X})$. The final $p(\tilde{\mathbf{X}}|\mathbf{X})$ defines the output $\tilde{\mathbf{X}}$ of the algorithm. See \citet{Tishby99} for more details.

In common applications of the IB method, $\mathbf{X}$ is a set of variables predetermined a priori. Conversely, when the IB method is used to identify sufficient statistics for causal inference, the selection of $\mathbf{X}$ is part of the process of determining between which variables and with which conditioning set a new conditional independence can be created identifying a sufficient set of statistics. To see this, we now examine the relation between the term $I(\mathbf{X};\mathrm{Z}|\tilde{\mathbf{X}})$ to be minimized in the IB method (Eq.\,\ref{e21}) and a mutual information of the form $I(\mathrm{X}; \mathrm{Z}| \mathbf{S}_{\Psi_{x,z}})$, which would quantify $\mathrm{X} \perp \mathrm{Z} |\mathbf{S}_{\Psi_{x,z}}$ as tested in R.c-ss or R.nc-ss. For simplicity, we consider first the case of a single sufficient statistic $\theta_z = g(\mathrm{X}, \tilde{\mathbf{V}})$, which given a set $\mathbf{S}$ creates the independence $\mathrm{X} \perp \mathrm{Z} |\mathbf{S}, \theta_z$. % As indicated in Section \ref{ss2}, $\mathbf{S}$ and $\tilde{\mathbf{V}}$ can overlap.
Since the function $\theta_z = g(\mathrm{X}, \tilde{\mathbf{V}})$ creates a dependence between $\mathrm{X}$ and any variable in $\tilde{\mathbf{V}}$ when conditioning on $\theta_z$, any variable $\mathrm{V}_i \in \tilde{\mathbf{V}}$ for which $\mathrm{Z} \notperp \mathrm{V}_i | \mathbf{S} \backslash \mathrm{V}_i, \theta_z$ must be included in $\mathbf{S}$ to obtain $\mathrm{X} \perp \mathrm{Z} |\mathbf{S}, \theta_z$. Accordingly, the sufficient statistic does not only create the independence $\mathrm{X} \perp \mathrm{Z} | \mathbf{S}, \theta_z$, but more generally $\{\mathrm{X}, \tilde{\mathbf{V}} \backslash \mathbf{S} \}\perp \mathrm{Z} | \mathbf{S}, \theta_z$. %In fact, in the limit that $\tilde{\mathbf{V}} \subseteq \mathbf{S}$ is required, the function $\theta_z = g(\mathrm{X}, \tilde{\mathbf{V}})$ cannot be invertible in $\mathrm{X}$, or $H(\mathrm{X}|\mathrm{S}, \theta_z)=0$.
This conditional independence leads to $I(\mathrm{X}, \tilde{\mathbf{V}}, \mathbf{S}; \mathrm{Z}|\mathbf{S}, \theta_z)=0$, since this mutual information can be decomposed into $I(\mathbf{S}; \mathrm{Z}|\mathbf{S}, \theta_z)$, which is zero by construction, and $I(\mathrm{X}, \tilde{\mathbf{V}} \backslash \mathbf{S} ; \mathrm{Z}|\mathbf{S}, \theta_z)$, which is zero due to the independence $\{\mathrm{X}, \tilde{\mathbf{V}} \backslash \mathbf{S} \}\perp \mathrm{Z} | \mathbf{S}, \theta_z$. We can now map $I(\mathrm{X}, \tilde{\mathbf{V}}, \mathbf{S}; \mathrm{Z}|\mathbf{S}, \theta_z)$ to $I(\mathbf{X};\mathrm{Z}|\tilde{\mathbf{X}})$ from Eq.\,\ref{e21}. The input to the IB algorithm must be at least $\mathbf{X}= \{ \mathrm{X}, \tilde{\mathbf{V}}, \mathbf{S} \}$, and the output for this selection of $\mathbf{X}$, if the algorithm performs correctly, would be $\tilde{\mathbf{X}} = \{\mathbf{S}, \theta_z\}$ .

The same logic to select $\mathbf{X}$ holds in general. The input must at least be $\mathbf{X} = \{ \mathrm{X}, \mathbf{S}, \alpha \}$. That is, the input must include the variable $\mathrm{X}$ for which the sufficient set of statistics has to be identified, the collection $\alpha$ of all the other arguments of the statistics, and the required conditioning set $\mathbf{S}$. If the algorithm performs correctly, the output for this $\mathbf{X}$ will be $\tilde{\mathbf{X}} = \mathbf{S}_{\Psi_{x,z}}$ or $\tilde{\mathbf{X}} = \mathbf{S}_{\Psi_{x,z}} \backslash \mathrm{Y}$, for $\mathrm{X} \perp \mathrm{Z}| \mathbf{S}_{\Psi_{x,z}}$ and $\mathrm{X} \perp \mathrm{Z}| \mathbf{S}_{\Psi_{x,z}} \backslash \mathrm{Y}$, respectively. Accordingly, what we will call the IB sufficient statistic $\theta_{IB} \equiv \tilde{\mathbf{X}}$ does not correspond only to the sufficient set of statistics $\Psi_{x,z}(\alpha)$, but to the whole conditioning set required to create the new conditional independence. In the case that auxiliary statistics are part of the sufficient set, the input $\mathbf{X}$ must also include their arguments (see Appendix B for details). Note that the fact that the input $\mathbf{X} = \{ \mathrm{X}, \mathbf{S}, \alpha \}$ jointly includes without distinction the arguments of all statistics \--and potentially of auxiliary statistics\-- as well as the conditioning set indicates that the IB algorithm estimates $\theta_{IB}$ in the space of all underlying sufficient statistics, without requiring an assumption about how many statistics exist, a distinction of which are auxiliary, or of which arguments of the statistics also belong to the conditioning set $\mathbf{S}$. From now on we will use $\theta_{IB}$ to refer to the correct output that the IB algorithm should return given a certain system and input $\mathbf{X}$, and we will use $\hat{\theta}_{IB}$ to refer to the actual output from the algorithm, which is an estimate of the underlying $\theta_{IB}$.

Given the relation between $I(\mathrm{X}; \mathrm{Z}| \mathbf{S}_{\Psi_{x,z}})$ and $I(\mathrm{X}; \mathrm{Z}| \theta_{IB})$ described above, the rules R.c-ss and R.nc-ss can be implemented as follows:

\vspace*{1mm}
\noindent \textbf{Implementation of rule $\mathbf{R.c-ss}$ with the IB method}: \emph{Consider variables $\mathrm{X}$, $\mathrm{Y}$, $\mathrm{Z}$, all nonseparable. For $\mathrm{Z}$ the target variable, find an input $\mathbf{X}$ for the IB method with $\mathrm{X} \in \mathbf{X}$ nonoverlapping with $\{\mathrm{Z}, \mathrm{Y} \}$ such that, given the output $\hat{\theta}_{IB}$, $\mathrm{X} \notperp \mathrm{Y}| \hat{\theta}_{IB}$, $\mathrm{Y} \notperp \mathrm{Z}| \hat{\theta}_{IB}$, and $\mathrm{X} \perp \mathrm{Z}| \hat{\theta}_{IB}$, then orient $\mathrm{X} *\--*\mathrm{Y} *\--* \mathrm{Z}$ as $\mathrm{X} *\rightarrow \mathrm{Y} \leftarrow* \mathrm{Z}$.}

\vspace*{1mm}
\noindent \textbf{Implementation of rule $\mathbf{R.nc-ss}$ with the IB method}: \emph{Consider variables $\mathrm{X}$, $\mathrm{Y}$, $\mathrm{Z}$, all nonseparable. For $\mathrm{Z}$ the target variable, find an input $\mathbf{X}$ for the IB method with $\mathrm{X} \in \mathbf{X}$ and $\mathrm{Y} \in \mathbf{X}$ nonoverlapping with $\mathrm{Z}$ such that, given the output $\hat{\theta}_{IB}$, $\mathrm{X} \perp \mathrm{Z}| \hat{\theta}_{IB}$. Check that, for $\mathrm{Z}$ the target variable, using as input $\mathbf{X} \backslash \mathrm{Y}$ the output $\hat{\theta}'_{IB}$ leads to $\mathrm{X} \notperp \mathrm{Z}| \hat{\theta}'_{IB}$ or to $H(\mathrm{X}|\hat{\theta}'_{IB})=0$ and that, for $\mathrm{Y}$ the target variable, for all inputs $\mathbf{X}' \subseteq \mathbf{X} \backslash \mathrm{Y}$ the output $\hat{\theta}''_{IB}$ leads to $\mathrm{X} \notperp \mathrm{Y}| \hat{\theta}''_{IB}$ or $H(\mathrm{X}|\hat{\theta}''_{IB})=0$. Then mark $\mathrm{X} *\--*\mathrm{Y} *\--* \mathrm{Z}$ as $\mathrm{X} *-\underline{*\mathrm{Y}*}-* \mathrm{Z}$.}

\vspace*{1mm}
For R.c-ss, the conditions $\mathrm{X} \notperp \mathrm{Y}| \hat{\theta}_{IB}$, $\mathrm{Y} \notperp \mathrm{Z}| \hat{\theta}_{IB}$, and $\mathrm{X} \perp \mathrm{Z}| \hat{\theta}_{IB}$ correspond to $\mathrm{X} \notperp \mathrm{Y} |\mathbf{S}_{\Psi_{x,z}}$, $\mathrm{Y} \notperp \mathrm{Z} |\mathbf{S}_{\Psi_{x,z}}$, and $\mathrm{X} \perp \mathrm{Z} |\mathbf{S}_{\Psi_{x,z}}$. The requirement in R.c-ss that $\mathbf{S}$ is nonoverlapping with $\{ \mathrm{Y}, \mathrm{Z}\}$ and $\mathrm{Y} \notin \bigcup \alpha$, is implemented by excluding $\{\mathrm{Y}, \mathrm{Z}\}$ from the input $\mathbf{X}$. Since $\mathrm{X} \in \mathbf{X}$, it is not excluded a priori that $\mathrm{X} \in \mathbf{S}$ as required in R.c-ss, but in that case the entropy $H(\mathrm{X} | \hat{\theta}_{IB})=0$ would invalidate $\hat{\theta}_{IB}$ as a useful statistic. For R.nc-ss, the conditions $\mathrm{X} \notperp \mathrm{Z}| \hat{\theta}'_{IB}$ or $H(\mathrm{X}|\hat{\theta}'_{IB})=0$ and $\mathrm{X} \perp \mathrm{Z}| \hat{\theta}_{IB}$ implement $\mathrm{X} \notperp \mathrm{Z} |\mathbf{S}_{\Psi_{x,z}}  \backslash \{\mathrm{Y}, \mathbf{\Psi}^{(y)}_{x,z}(\alpha)\}$ and at least $\mathrm{X} \perp \mathrm{Z} |\mathbf{S}_{\Psi_{x,z}}$ or $\mathrm{X} \perp \mathrm{Z} |\mathbf{S}_{\Psi_{x,z}} \backslash \mathrm{Y}$. The condition $H(\mathrm{X}|\hat{\theta}'_{IB})=0$ covers the case in which $\hat{\theta}'_{IB}$ includes $\mathrm{X}$ itself and $\mathrm{X} \notperp \mathrm{Z}| \hat{\theta}'_{IB}$ cannot be evaluated. The IB algorithm will include $\mathrm{Y}$ in $\hat{\theta}_{IB}$ only if $\mathrm{Z} \notperp \mathrm{Y}| \mathbf{S}_{\Psi_{x,z}} \backslash \mathrm{Y}$. In more detail, rule R.nc-ss checks that it is necessary the addition of $\{\mathrm{Y}, \mathbf{\Psi}^{(y)}_{x,z}(\alpha)\}$ or $\mathbf{\Psi}^{(y)}_{x,z}(\alpha)$ to the conditioning set to separate $\mathrm{X}$ and $\mathrm{Z}$. A direct implementation of the rule would need some procedure to identify the structure of $\hat{\theta}_{IB}$ corresponding to $\mathbf{\Psi}^{(y)}_{x,z}(\alpha)$. However, this is avoided by combining $\hat{\theta}_{IB}$ from input $\mathbf{X}$ and $\hat{\theta}_{IB}'$ from input $\mathbf{X} \backslash \mathrm{Y}$, since not including $\mathrm{Y}$ in the input excludes both that $\mathrm{Y}$ can be part of $\hat{\theta}'_{IB}$ and that the IB method can identify any statistic that has $\mathrm{Y}$ as an argument. Similarly, for the target variable being $\mathrm{Y}$, using all inputs $\mathbf{X}' \subseteq \mathbf{X} \backslash \mathrm{Y}$ allows guaranteeing $\mathrm{X} \notperp \mathrm{Y}|\mathbf{S}'$ $\forall \mathbf{S}' \subseteq \mathbf{S}_{\Psi_{x,z}}  \backslash \{\mathrm{Y}, \mathbf{\Psi}^{(y)}_{x,z}(\alpha)\}$ from $\mathrm{X} \notperp \mathrm{Y}| \hat{\theta}''_{IB}$ or $H(\mathrm{X}|\hat{\theta}''_{IB})=0$ $\forall \hat{\theta}''_{IB}$. Furthermore, since the IB method is applied with $\mathrm{Z}$ and not $\mathrm{X}$ as the target variable, $\hat{\theta}_{IB}$ cannot comprise any sufficient statistic with $\mathrm{Z}$ as an argument that could lead to $\mathrm{Y} \perp \mathrm{Z} | \mathbf{S}'$ for some $\mathbf{S}' \subseteq \mathbf{S}_{\Psi_{x,z}}  \backslash \{\mathrm{Y}, \mathbf{\Psi}^{(y)}_{x,z}(\alpha)\}$.

Besides the implementation of the rules, we continue discussing the implementation of the IB algorithm. Apart from $\mathbf{X}$, the other input critical to the IB algorithm is parameter $\beta$. In common applications of the IB method, $\beta$ may be selected such that some information about $\mathrm{Z}$ is lost, if this allows a desirable higher compression. This is not the case here; $\tilde{\mathbf{X}}$ has to preserve all the information to create an additional independence $I(\mathbf{X};\mathrm{Z}|\tilde{\mathbf{X}})=0$. Consider a single sufficient statistic, such that $\mathrm{Z} \perp \mathrm{X} | \mathbf{S}, \theta_z(\mathrm{X}; \tilde{\mathbf{V}})$. If the distribution $p(\mathrm{Z}, \mathrm{X}, \tilde{\mathbf{V}}, \mathbf{S})$ was estimated perfectly, a value $\beta \rightarrow \infty$ could be selected. This is because if it exists a $p(\tilde{\mathbf{X}} | \mathbf{X})$ such that $I(\mathbf{X};\mathrm{Z}|\tilde{\mathbf{X}})=0$, the minimization in Eq.\,\ref{e21} reduces to finding $\tilde{\mathbf{X}}$ such that apart from producing $I(\mathbf{X};\mathrm{Z}|\tilde{\mathbf{X}})=0$ also minimizes $I(\tilde{\mathbf{X}};\mathbf{X})$. However, with an imperfect estimation from finite data, $\beta \rightarrow \infty$ may enforce the preservation of information resulting only from differences between probability values due to the imperfect estimation, leading to a useless output $\tilde{\mathbf{X}} = \mathbf{X}$. In practice, high values of $\beta$ should be selected to prioritize information preservation over compression, but not so high that sampling fluctuations dominate the determination of $\tilde{\mathbf{X}}$. The scale to assess the magnitude of $\beta$ values is determined by the ratio $I(\mathbf{X};\mathrm{Z})/H(\mathbf{X})$, since the entropy $H(\mathbf{X})$ is the maximum possible value of $I(\tilde{\mathbf{X}};\mathbf{X})$ and $I(\mathbf{X};\mathrm{Z})$ the maximum possible value of $I(\tilde{\mathbf{X}};\mathrm{Z})$. Defining  $\beta' \equiv \beta I(\mathbf{X};\mathrm{Z})/H(\mathbf{X})$, the function to be minimized to select $p(\tilde{\mathbf{X}}|\mathbf{X})$ can be reexpressed as:
\small
\begin{equation}
\label{e22}
\begin{split}
p(\tilde{\mathbf{X}}|\mathbf{X}) & \equiv \mathrm{argmin}\  I(\tilde{\mathbf{X}};\mathbf{X})-\beta I(\tilde{\mathbf{X}};\mathrm{Z}) \\
& = \mathrm{argmin}\  I(\tilde{\mathbf{X}};\mathbf{X})-\beta' \frac{H(\mathbf{X})}{I(\mathbf{X};\mathrm{Z})} I(\tilde{\mathbf{X}};\mathrm{Z}) = \mathrm{argmin}\ H(\mathbf{X}) \left [ \frac{I(\tilde{\mathbf{X}};\mathbf{X})}{H(\mathbf{X})} -  \beta' \frac{I(\tilde{\mathbf{X}};\mathrm{Z})}{I(\mathbf{X};\mathrm{Z})}  \right ].
\end{split}
\end{equation}
\normalsize
Because $I(\tilde{\mathbf{X}};\mathbf{X})/H(\mathbf{X}) \leq 1$ and $I(\tilde{\mathbf{X}};\mathrm{Z})/ I(\mathbf{X};\mathrm{Z}) \leq 1$, values $\beta' \gg 1$ enforce information preservation. As discussed in Section \ref{ss33} below, we will combine results from a whole set of high $\beta'$ values in order to select the sufficient statistics. Furthermore, the IB algorithm also requires as input a maximal cardinality $\max |\hat{\theta}_{IB}|$ for the output variable $\hat{\theta}_{IB} \equiv \tilde{\mathbf{X}}$, needed to initialize the projection $p(\tilde{\mathbf{X}}|\mathbf{X})$. In our procedure we also do not preselect a fixed maximal cardinality, but instead combine the results of the IB algorithm across different maximal cardinalities to identify candidate sufficient statistics $\hat{\theta}_{IB}$. Selection criteria to select these candidate statistics are detailed in Section \ref{ss33} below. More technical details of the implementation of the IB algorithm are provided in Appendix G.

The IB algorithm returns a compressed representation $\tilde{\mathbf{X}}$, determined by the optimized conditional distribution $p(\tilde{\mathbf{X}}|\mathbf{X})$. In the presence of a single underlying sufficient statistic, the mapping $p(\tilde{\mathbf{X}}|\mathbf{X})$ correctly infers the form of the statistic if it approximates the deterministic relation $\theta_z = g(\mathrm{X}, \tilde{\mathbf{V}})$ so that for each value $\mathbf{X}= \mathbf{x}$ there is a value $\tilde{\mathbf{X}} = \tilde{\mathbf{x}}_0$ such that $p(\tilde{\mathbf{X}}= \tilde{\mathbf{x}}_0 |\mathbf{X}= \mathbf{x}) \simeq 1$, while $p(\tilde{\mathbf{X}}\neq \tilde{\mathbf{x}}_0 |\mathbf{X}= \mathbf{x}) \simeq 0$. Accordingly, a candidate sufficient statistic is constructed setting $p(\tilde{\mathbf{X}}= \tilde{\mathbf{x}}_0|\mathbf{X})=1 $ for $\tilde{\mathbf{x}}_0 = \mathrm{argmax}_j\ p(\tilde{\mathbf{X}}= \tilde{\mathbf{x}}_j|\mathbf{X})$ and $p(\tilde{\mathbf{X}}\neq \tilde{\mathbf{x}}_0|\mathbf{X})=0$ otherwise. This modified distribution defines the potential sufficient statistic $\hat{\theta}_{IB}$. The same procedure works with a sufficient set comprising several statistics to define $\hat{\theta}_{IB}$ in the joint space of those statistics.

\subsection{Selection of sufficient statistics identified with the information bottleneck method}
\label{ss33}

As described above, a candidate sufficient statistic $\hat{\theta}_{IB}$ is obtained from the IB algorithm for each selection of the inputs $\mathbf{X}$, $\beta$, and $\max |\hat{\theta}_{IB}|$. We now describe a procedure to combine $\hat{\theta}_{IB}$ outputs in order to select the sufficient statistic used with rules R.c-ss and R.nc-ss. We call the proposed procedure the Information Bottleneck Sufficient Statistics Inference method, the IBSSI method. The core of the IBSSI method is a procedure to select sufficient statistics for fixed $\mathbf{X}$ and $\beta$ (see algorithm \ref{alg1}). For a fixed $\mathbf{X}$, the maximal cardinality $\max |\hat{\theta}_{IB}|$ has to be selected from the range $\{2,..., |\mathbf{X}|\}$, where $|\mathbf{X}|$ is the cardinality of the input. % If $|\mathbf{X}|$ is high and the cardinality of $\theta_{IB}$ is expected to be close to $|\mathbf{X}|$, a higher lower bound can be used instead of $2$.
Algorithm \ref{alg1} starts applying the IB algorithm for $\max |\hat{\theta}_{IB}|= |\mathbf{X}|$, that is, the initial maximal dimensionality of $\tilde{\mathbf{X}}$ for the mapping $p(\tilde{\mathbf{X}}| \mathbf{X})$ is equal to $|\mathbf{X}|$. If the cardinality of the output $\hat{\theta}_{IB}$ is lower than $|\mathbf{X}|$, the algorithm iterates the IB algorithm using $\max |\hat{\theta}_{IB}|= |\mathbf{X}|-k$ for the $k$-th iteration, until one of the following things happens. First, if at iteration $k$ the output is such that $|\hat{\theta}_{IB}^{(k)}| > |\hat{\theta}_{IB}^{(k-1)}|$, the algorithm does not return any sufficient statistic. If the iterations stop because the bound $\max |\hat{\theta}_{IB}|=2$ is reached, or because $|\hat{\theta}_{IB}^{(k)}| < |\hat{\theta}_{IB}^{(k-1)}|$, it checks that the statistic retrieved in the last two iterations of equal cardinality was the same. Lines $12-14$ serve to ensure that the comparison in line $15$ is applied to the right iterations. The equality of the inferred $\hat{\theta}_{IB}$ in adjacent iterations ensures some degree of robustness of the sufficient statistics retrieved, and could be made more demanding asking not only that $\hat{\theta}_{IB}^{(k)} = \hat{\theta}_{IB}^{(k-1)}$ (line $15$), but an equality for a wider range of $\max |\hat{\theta}_{IB}|$ values. On the other hand, the requirement that already when using $\max |\hat{\theta}_{IB}|= |\mathbf{X}|$ a sufficient statistic of lower cardinality is found (line $4$) could be removed, as long as it is found for a lower value of $\max |\hat{\theta}_{IB}|$.

\begin{algorithm}[H]
\caption{Selection of a sufficient statistic $\hat{\theta}_{IB}$ for $\mathrm{X}$ with respect to $\mathrm{Z}$ for fixed values of $\mathbf{X}$ and $\beta$.}
\label{alg1}
\begin{algorithmic}[1]
\renewcommand{\algorithmicrequire}{\textbf{Input:}}
\renewcommand{\algorithmicensure}{\textbf{Output:}}
 \REQUIRE $\mathrm{X}$, $\mathrm{Z}$, $\mathbf{X}$, $\beta$, $a_{I}$, $a_{H_\mathrm{X}}$, $a_{H_\mathrm{Z}}$
 \ENSURE  $\hat{\theta}_{IB}$
 \STATE $\hat{\theta}_{IB}$ empty
 \STATE $\max |\hat{\theta}_{IB}|\leftarrow |\mathbf{X}|$
 \STATE $\hat{\theta}_{IB}^{(0)} \leftarrow IB(\mathrm{Z},\mathbf{X}, \beta, \max |\hat{\theta}_{IB}|)$
 \IF{$|\hat{\theta}_{IB}^{(0)}|< |\mathbf{X}|$}
    \STATE $k \leftarrow 0$
    \REPEAT
      \STATE $k \leftarrow k +1$
      \STATE $\max |\hat{\theta}_{IB}|\leftarrow |\mathbf{X}| -k$
      \STATE $\hat{\theta}_{IB}^{(k)} \leftarrow IB(\mathrm{Z},\mathbf{X}, \beta, \max |\hat{\theta}_{IB}|)$
    \UNTIL{$\max |\hat{\theta}_{IB}| = 2\ \  \mathrm{OR} \ \ |\hat{\theta}_{IB}^{(k)}| \neq |\hat{\theta}_{IB}^{(k-1)}| $}
    \IF{$|\hat{\theta}_{IB}^{(k)}| \leq |\hat{\theta}_{IB}^{(k-1)}|$}
       \IF{$|\hat{\theta}_{IB}^{(k)}| < |\hat{\theta}_{IB}^{(k-1)}|$}
       \STATE $k \leftarrow k-1$
       \ENDIF

       \IF{$\hat{\theta}_{IB}^{(k)} = \hat{\theta}_{IB}^{(k-1)}$}

         \STATE $c_\mathrm{X} \leftarrow H(\mathrm{X}|\hat{\theta}_{IB}^{(k)}) > a_{H_\mathrm{X}} H(\mathrm{X})$
         \STATE $c_\mathrm{Z} \leftarrow H(\mathrm{Z}|\hat{\theta}_{IB}^{(k)}) > a_{H_\mathrm{Z}} H(\mathrm{Z})$
         \STATE $c_I \leftarrow I(\mathrm{X}; \mathrm{Z}| \hat{\theta}_{IB}^{(k)}) < a_{I} I(\mathrm{X}; \mathrm{Z})$

         \IF{$ c_I = c_\mathrm{X} = c_\mathrm{Z} = \mathrm{true} $}
            \STATE $\hat{\theta}_{IB} \leftarrow \hat{\theta}_{IB}^{(k)}$
         \ENDIF
       \ENDIF
    \ENDIF
 \ENDIF
\end{algorithmic}
\end{algorithm}

Finally, lines $16-18$ evaluate the Selection Criteria that make $\hat{\theta}_{IB}$ useful to apply the rules of causal learning. First, $\hat{\theta}_{IB}$ cannot be such that $H(\mathrm{X}|\hat{\theta}_{IB})=0$ or $H(\mathrm{Z}|\hat{\theta}_{IB})=0$, since this would prevent from evaluating any conditional dependence or independence as required in the rules. Because $H(\mathrm{X})$ is the maximum value that $H(\mathrm{X}|\hat{\theta}_{IB})$ can have, we require $H(\mathrm{X}|\hat{\theta}_{IB})> a_{H_\mathrm{X}} H(\mathrm{X})$, and analogously $H(\mathrm{Z}|\hat{\theta}_{IB})> a_{H_\mathrm{Z}} H(\mathrm{Z})$, where $a_{H_\mathrm{X}}$ and $a_{H_\mathrm{Z}}$ are factors determining the percentage of entropy left after conditioning. We present our results for $a_{H_\mathrm{X}} = a_{H_\mathrm{Z}} = 0.1$, and these results are robust as long as the bounds are not too close to zero. Requiring $H(\mathrm{X}|\hat{\theta}_{IB})>0$ discards that $\tilde{\mathbf{X}} = \mathbf{X}$, as may happen if $\beta$ is too high, as discussed above.  Second, the condition that defines a sufficient statistic is that $I(\mathrm{X}; \mathrm{Z}| \theta_{IB})=0$, which as discussed above corresponds to $I(\mathrm{X}; \mathrm{Z}| \mathbf{S}_{\Psi_{x,z}})=0$. To assess the creation of a conditional independence we check that conditioning on $\hat{\theta}_{IB}$ reduces the mutual information $I(\mathrm{X}; \mathrm{Z})$ below a small percentage of its value. In particular, we chose $a_{I} = 0.025$. This value should be small enough to discard candidates of $\hat{\theta}_{IB}$ that do not create an independence, but cannot be too small, given that information theoretic quantities have an intrinsic positive estimation bias, which increases for conditional mutual informations due to the higher dimensionality \citep{Panzeri2007}. The selection criteria of lines $16-18$ are to be evaluated in a testing set different from the training data set used to infer $\hat{\theta}_{IB}$ with the IB algorithm, in order to avoid selecting sufficient statistics due to overfitting. Since these selection criteria constitute a step previous to the evaluation of all conditional dependencies and independencies involved in the rules, further tests of significance can be applied subsequently. The testing of conditional independencies is not specific to our new rules, but an issue common to standard rules, and it is not our objective here to optimize the tests. Accordingly, the criterion of line $18$ based on mutual information stands for any analogous test of a conditional independence created by $\hat{\theta}_{IB}$.

In a subsequent step, we need to combine the sufficient statistics $\hat{\theta}_{IB}$ obtained for different inputs $\mathbf{X}$ and $\beta$. The IBSSI method proceeds selecting a set of high $\beta$ values and accepting any statistic found for any $\beta$ value within the set. As discussed above, given $\beta = \beta' H(\mathbf{X})/ I(\mathbf{X};\mathrm{Z})$, the high $\beta$ values are determined by $\beta' \gg 1$. The selection criteria imposed in algorithm \ref{alg1} make the procedure robust to the $\beta'$ range. If $\beta'$ is too low, allowing for an excess of compression, the condition $c_I = \mathrm{true}$ (line 18 in the algorithm) will not be fulfilled. Similarly, if $\beta'$ is too high and $\tilde{{\mathbf{X}}} = \mathbf{X}$ is retrieved, the condition $c_\mathrm{X} = \mathrm{true}$ (line 16) will not be fulfilled. Nonetheless, because the results depend on the estimation of the information theoretic quantities and the particular selection of the thresholds $a_{I}$ and $a_{H_\mathrm{X}}$, the selection of lower or higher $\beta'$ values determines a tradeoff between sensitivity and specificity identifying sufficient statistics, as we will illustrate comparing performance results for two sets $\beta' \in \{25, 50 , 75, 100\}$ and $\beta' \in \{50 , 75, 100\}$. Instead of accepting any statistic found for any $\beta'$ value, a more conservative approach would be to check that the identity of $\hat{\theta}_{IB}$ is the same across a certain range of $\beta'$ values, analogously to how algorithm \ref{alg1} checks that the same $\hat{\theta}_{IB}$ is inferred with different $\max |\hat{\theta}_{IB}|$ values (line 15). Again, the preferred implementation depends on the tradeoff between sensitivity and specificity, and the same principles can be applied adjusting for various robustness requirements.

As a last step, a criterion is needed to select the inputs $\mathbf{X}$. As explained in Section 4.1, apart from variable $\mathrm{X}$, for which the existence of a sufficient set of statistics with respect to $\mathrm{Z}$ is hypothesized, $\mathbf{X}$ should comprise the variables hypothesized to be the other arguments of the sufficient statistics, and the variables hypothesized to belong to $\mathbf{S}$, a required conditioning set. In this work we will analyze examples involving only few variables, and adopt the simplest strategy of starting with the lowest possible set $\mathbf{X}$ and consider larger sets if a valid sufficient set of statistics is not found. In general, the selection of a conditioning set $\mathbf{S}$ is an issue common to the standard algorithms, and strategies to select it have been studied before. The FCI algorithm \citep{Spirtes00} has a procedure to establish the order in which different conditioning sets are tested and subsequent refinements to make the application of the rules order-independent have been proposed \citep{Colombo14}. The selection of candidate variables $\mathrm{X}$, $\mathrm{Z}$, and candidate arguments for the statistics can be informed by the partially oriented graphs obtained as the output of the standard structure learning algorithms. Note that the identification of a sufficient statistic may lead to the orientation of more edges not only because of the immediate application of rules R.c-ss and R.nc-ss. If some new edges have been oriented with R.c-ss and R.nc-ss, this new information may enable the standard rules to be applied for some instances for which they could not before. This naturally leads to conceive an iterative procedure in which at each step where to seek for candidate sufficient statistics is determined by the available causal information from previous iterations. The implementation of this iterative identification of sufficient statistics combined with a standard structure learning algorithm is out of the scope of this work. In Appendix C we provide a detailed description of the CI algorithm of \cite{Spirtes00} augmented with rules based on sufficient statistics, but under the assumption that the statistics have been already previously inferred, instead of considering an iterative identification of the statistics and application of the orientation rules.

\section{Examples of identification of sufficient statistics}
\label{ss4}

We now study examples for which sufficient statistics are inferred with the IBSSI method. First, we analyze simulated data to study performance across different configurations of systems with common sufficient sets of statistics. For this purpose, we simulated systems in which the conditional distribution of the variables in whose functional equation the sufficient statistics are embodied has the form of Generalised Linear Models (GLMs) \citep{Nelder72}, which are widely applied in social and biological sciences. Second, we also study performance across data sets sampled from fixed configurations. For this purpose, we generated data from a concrete set of functional equations previously shown to accurately model a biological signal transduction network \citep{Li06}. The equations have the form of Boolean regulatory rules, which are widely applied to model regulatory and signaling networks in systems biology \citep{Wang12, Abou16, Chen18} and systems pharmacology \citep{Bloomingdale18}. The system we study has also previously been included \citep{Jenkins08} in the workbench designed for the Causation and Prediction Challenge of \cite{Guyon08} (http://www.causality.inf.ethz.ch). Further examples are studied in Appendix E.

\subsection{Characterization of the IBSSI method with simulated data}
\label{ss42}

\begin{figure*}[t]
  \begin{center}
    \scalebox{0.53}{\includegraphics*{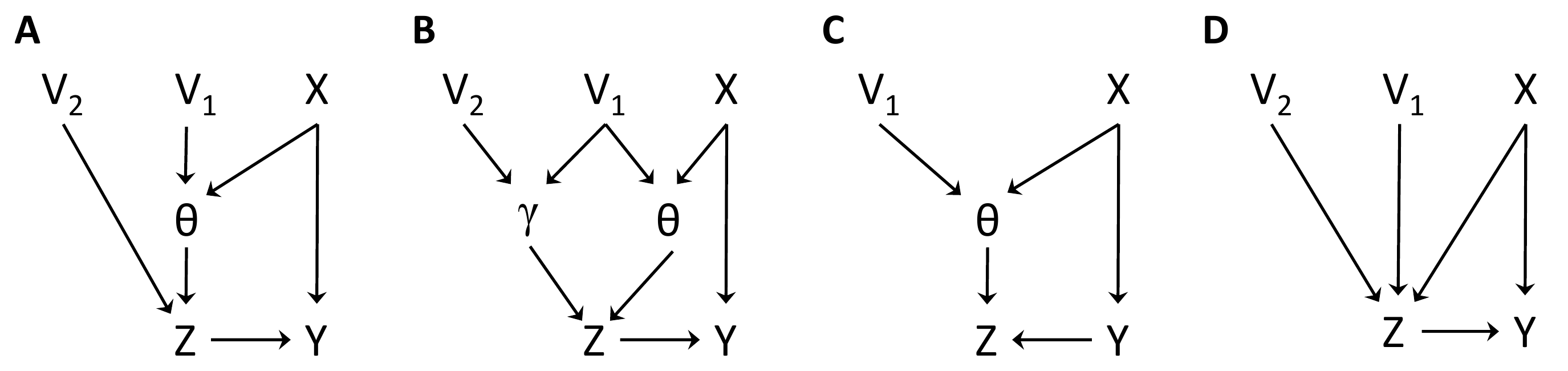}}
  \end{center}
  %\vspace{0.25in}
  \caption{Concrete causal structures studied to assess the performance of the IBSSI method finding sufficient statistics useful for structure learning. \textbf{A}) Structure with a single sufficient statistic $\theta$ that allows applying rule R.c-ss to infer that $\mathrm{Y}$ is a collider. \textbf{B}) Structure with a statistic $\theta$ and an auxiliary statistic $\gamma$ that allow applying rule R.c-ss to infer that $\mathrm{Y}$ is a collider. \textbf{C}) Structure with a sufficient statistic $\theta$ that allows applying rule R.nc-ss to infer that $\mathrm{Y}$ is a noncollider. \textbf{D}) Structure with no sufficient statistics.}
  \label{f02}
\end{figure*}

We here study the performance identifying sufficient statistics across configurations of systems with the causal structures of Figure \ref{f02}. For the sake of space, we only represent the graphs $G^+_{\theta}$ already incorporating sufficient statistics. In Figure \ref{f02}A there is a sufficient statistic $\theta$ for $\mathrm{X}$ with respect to $\mathrm{Z}$, and in Figure \ref{f02}B there is also an auxiliary statistic $\gamma$, such that in both cases the new conditional independence obtained from the statistics would allow applying rule R.c-ss to infer a collider $\mathrm{X} * \rightarrow \mathrm{Y} \leftarrow * \mathrm{Z}$. In Figure \ref{f02}C the statistic allows applying rule R.nc-ss to infer that $\mathrm{Y}$ is a noncollider. In all these three cases the nonseparability of $\mathrm{X}$ and $\mathrm{Z}$ prevents the application of rules R.c and R.nc. Finally, no statistic exists in Figure \ref{f02}D, and hence the IB algorithm should not return a valid $\hat{\theta}_{IB}$.

We will focus on evaluating the performance in the identification of sufficient sets of statistics, and not on the full application of rules R.c-ss and R.nc-ss. We chose this strategy because the identification of a sufficient set of statistics already involves verifying the creation of a new conditional independence, which is the key distinctive component of the new rules. Furthermore, except for the system of Figure \ref{f02}C for which rule R.nc-ss applies, the application of the IBSSI method to these examples does not require specifying a functional equation for $\mathrm{Y}$, so that we can segregate the statistics' identification analysis from the \--contingent\-- selection of a functional equation for $\mathrm{Y}$. In more detail, for the systems of Figure \ref{f02}A-B, the identification of the correct $\theta_{IB}$ involves the verification of $\mathrm{X} \perp \mathrm{Z}| \hat{\theta}_{IB}$ (line 18 of algorithm 1), as required in the implementation of R.c-ss with the IB method (Section \ref{ss32}). For these two causal structures, the presence of a sufficient set of statistics only depends on the form of the functional equation of $\mathrm{Z}$, and the performance identifying the statistics will depend only on the properties of $p(\mathrm{Z},\mathrm{X},\mathrm{V}_1,\mathrm{V}_2)$, not on the properties of the functional equation of $\mathrm{Y}$ and of $p(\mathrm{Y}| \mathrm{X},\mathrm{Z})$. On the other hand, a full implementation of rule R.c-ss further requires using the inferred $\hat{\theta}_{IB}$ to check that $\mathrm{X} \notperp \mathrm{Y}| \hat{\theta}_{IB}$ and $\mathrm{Y} \notperp \mathrm{Z}| \hat{\theta}_{IB}$. The actual fulfillment of these dependencies, as well as the performance verifying them, would depend on the properties of the specific functional equation of $\mathrm{Y}$ and $p(\mathrm{Y}| \mathrm{X},\mathrm{Z})$ studied. For example, if we chose a system in which the link $\mathrm{X} \rightarrow \mathrm{Y}$ is too weak, the dependence $\mathrm{X} \notperp \mathrm{Y}| \hat{\theta}_{IB}$ may not be detected and the application of the rule would fail. However, we want to isolate the evaluation of the performance of the IBSSI method from these additional factors that affect the applicability of the new rules but that are not distinctive of these new rules. Similarly, in the case of Figure \ref{f02}D, the selection of a particular form of the functional equation of $\mathrm{Y}$ is not required to examine whether the IBSSI method avoids false positives when an underlying sufficient set of statistics does not exist. Only in the case of Figure \ref{f02}C we will explicitly model variable $\mathrm{Y}$, since it is involved in the detection of the statistic with the independence $\mathrm{Z} \perp \mathrm{X}| \theta, \mathrm{Y}$.

In more detail, in this section for each of the causal structures in Figure \ref{f02} we study configurations generated from the same type of systems, with fixed generative mechanisms, and a concrete form of the corresponding sufficient statistics. Additional examples in Appendix E comprise systems generated with alternative mechanisms or with other forms of the sufficient statistics. In particular, here we study systems in which $\mathrm{Z}$ depends on its parents through a binomial GLM. $\mathrm{Z}$ is generated from a binomial distribution $B(n,p_\mathrm{z})$, where $n$ is the number of trials and $p_\mathrm{z}$ the probability of obtaining $\mathrm{Z} = 1$ in a trial, which is determined with a logit link function $p_\mathrm{z} = 1/(1+\mathrm{exp}(-h(\mathbf{Pa}_z)))$. For the systems corresponding to Figure \ref{f02}A-B, we constructed functions $h(\mathbf{Pa}_z)$ with the following form:

\begin{subequations}
\label{e1}
\begin{align}
h &=  a_0+ a_1 \mathrm{V}_2 + a_2(\mathrm{X}+\mathrm{V}_1) + a_3 (\mathrm{X}+\mathrm{V}_1)^2 + a_4(\mathrm{X}+\mathrm{V}_1)\mathrm{V}_2 + a_5(\mathrm{X}+\mathrm{V}_1)^2 \mathrm{V}_2 \label{e1a} \\
h &=  a_0+ a_1(\mathrm{V}_1+\mathrm{V}_2) + a_2(\mathrm{V}_1+\mathrm{V}_2)^2 + a_3(\mathrm{X}+\mathrm{V}_1) + a_4(\mathrm{X}+\mathrm{V}_1)(\mathrm{V}_1+\mathrm{V}_2) \label{e1b}\\
                 &\ \ \ + a_5(\mathrm{X}+\mathrm{V}_1)^2 + a_6(\mathrm{X}+\mathrm{V}_1)^2(\mathrm{V}_1+\mathrm{V}_2)^2 \notag,
\end{align}
\end{subequations}
where $\mathrm{X}, \mathrm{V}_1$, and $\mathrm{V}_2$ are independent binary variables with values $0,1$. In Eq.\,\ref{e1a}, $\theta = \mathrm{X}+\mathrm{V}_1$ is a sufficient statistic for $\mathrm{X}$ (Fig.\,\ref{f02}A). In Eq.\,\ref{e1b}, $\theta = \mathrm{X}+\mathrm{V}_1$ is again a statistic for $\mathrm{X}$ and furthermore $\gamma = \mathrm{V}_1+\mathrm{V}_2$ is an auxiliary statistic (Fig.\,\ref{f02}B). For the systems corresponding to Figure \ref{f02}C-D, we constructed functions $h(\mathbf{Pa}_z)$ with the following form:

\begin{subequations}
\label{e2}
\begin{align}
h &=  a_0+ a_1 \mathrm{Y} + a_2(\mathrm{X}+\mathrm{V}_1) + a_3 (\mathrm{X}+\mathrm{V}_1)^2
                 + a_4(\mathrm{X}+\mathrm{V}_1)\mathrm{Y} + a_5(\mathrm{X}+\mathrm{V}_1)^2 \mathrm{Y} \label{e2a}\\
h &=  a_0+ a_1 \mathrm{V}_1 + a_2 \mathrm{V}_2 + a_3 \mathrm{X} + a_4 \mathrm{X} \mathrm{V}_1
                 + a_5 \mathrm{X} \mathrm{V}_2 + a_6 \mathrm{V}_1 \mathrm{V}_2, \label{e2b}
\end{align}
\end{subequations}
where all variables are binary with values $0, 1$. In Fig.\,\ref{f02}C, the link $\mathrm{X} \rightarrow \mathrm{Y}$ was modeled by $p(\mathrm{Y} = 1| \mathrm{X} = \mathrm{x}) = 0.3 + 0.4 \mathrm{x}$ and $\theta = \mathrm{X}+\mathrm{V}_1$ is a sufficient statistic for $\mathrm{X}$. Oppositely, Eq.\,\ref{e2b} does not contain any sufficient statistic (Fig.\,\ref{f02}D).

\begin{figure*}[t]
  \begin{center}
    \scalebox{0.42}{\includegraphics*{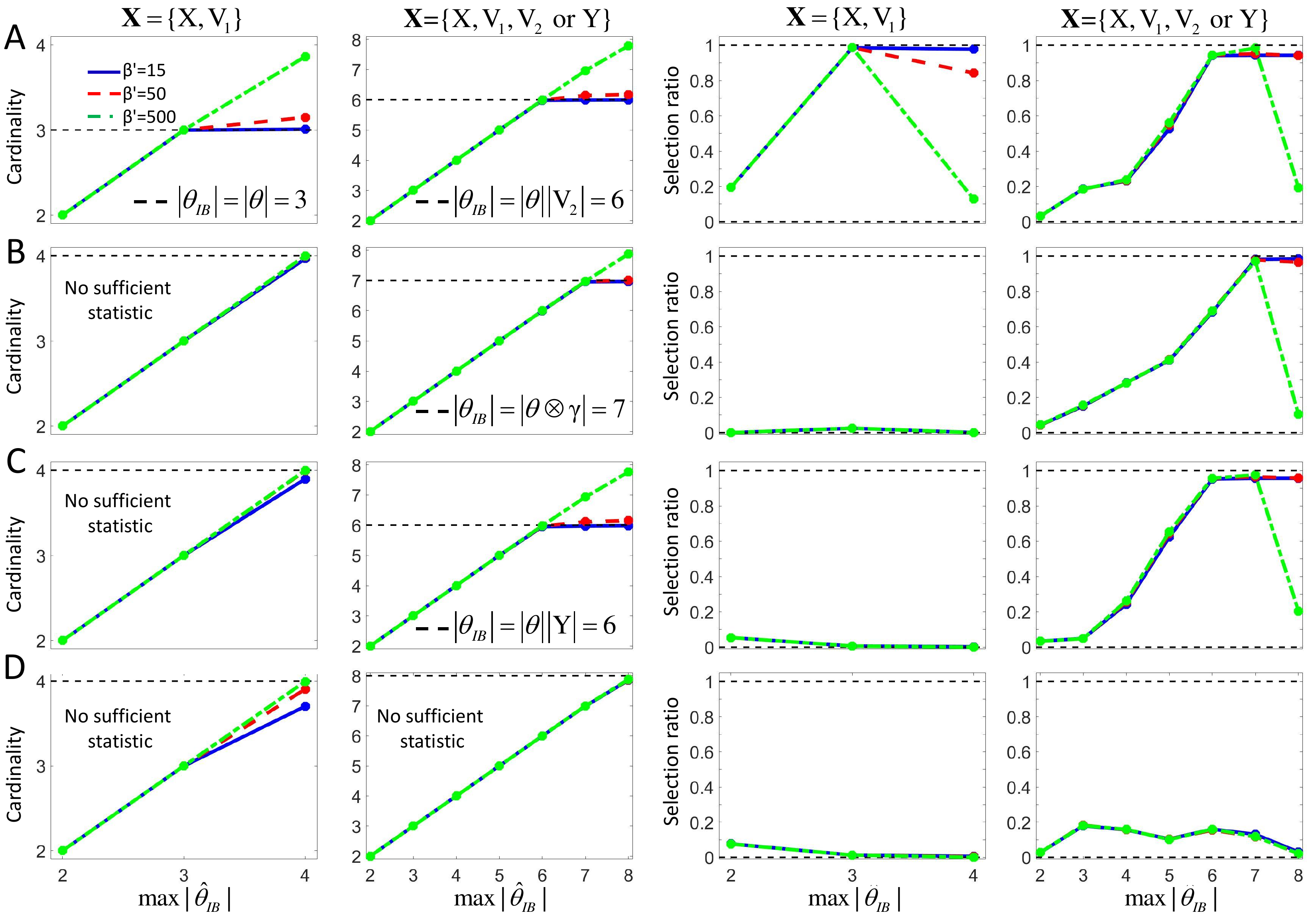}}
  \end{center}
  %\vspace{0.25in}
  \caption{Identification of sufficient statistics with the Information Bottleneck method for the systems of Figure \ref{f02}. \textbf{A}) Results for configurations generated following Eq.\,\ref{e1a}. \textbf{B}-\textbf{D}) Same as A) for systems generated from Eqs.\,\ref{e1b}, \ref{e2a}, and \ref{e2b}. First column shows the average cardinality of the estimated $\hat{\theta}_{IB}$ as a function of the input maximal cardinality $\max |\hat{\theta}_{IB}|$ when the input to the IB algorithm is $\mathbf{X} = \{ \mathrm{X}, \mathrm{V}_1\}$. Second column shows the average cardinality when the input is $\mathbf{X} = \mathbf{Pa}_z$, which corresponds to $\mathbf{X} = \{ \mathrm{X}, \mathrm{V}_1, \mathrm{V}_2\}$ for all systems except $\mathbf{X} = \{ \mathrm{X}, \mathrm{V}_1, \mathrm{Y}\}$ for the system in C). Results are shown for three values of $\beta'$ (Eq.\,\ref{e22}), increasingly enforcing information preservation for higher $\beta'$. The dashed horizontal line indicates the cardinality of the subjacent sufficient statistic $\theta_{IB}$ when it exists, or the cardinality $|\mathbf{X}|$ if it does not exist. The legend specifies the cardinality $|\theta_{IB}|$ in terms of the specific $\mathbf{S}_{\Psi_{x,z}}$ that results from the input $\mathbf{X}$. The third and fourth columns show the ratio of configurations for which the candidate sufficient statistics retrieved from the IB algorithm fulfill the selection criteria (lines 16-18 in algorithm \ref{alg1}). Here and throughout the results we selected thresholds $H(\mathrm{X}|\hat{\theta}_{IB})/H(\mathrm{X})> 0.1$, $H(\mathrm{Z}|\hat{\theta}_{IB})/H(\mathrm{Z})> 0.1$, and $I(\mathrm{X};\mathrm{Z}|\hat{\theta}_{IB})/I(\mathrm{X};\mathrm{Z})< 0.025$ to select a sufficient statistic.}
  \label{f2}
\end{figure*}

For each type of system we generated a set of $1440$ configurations consistent with its causal structure and functional equation. Because we aimed to assess the performance inferring the sufficient statistics characteristic of the systems, we used faithfulness constraints to select the sets of parameters. % resulting in distributions faithful to the causal structures \citep{Spirtes00}.
As discussed in Section \ref{ss2}, beyond the standard faithfulness assumption \citep{Pearl86,Spirtes00,Pearl09} %guarantees that a conditional independence exists if and only if associated with the corresponding d-separation in the causal structure \citep{Pearl86,Spirtes00,Pearl09}.
our extended faithfulness assumption further requires that a new conditional independence between observable variables appears by introducing a new variable $\hat{\theta}$ deterministically determined by some observable variables if and only if $\hat{\theta}$ corresponds to an underlying functional sufficient statistic. %, as represented in the augmented graphs $G^+_{\theta}$ (i.e.\, $\hat{\theta} = \theta$).
%That is, we require not only that the joint distribution of the configuration is isomorphic to the corresponding graph $G$ of the system \citep{Pearl86,Pearl09}, but more strongly that it is also isomorphic to the corresponding augmented graph $G^+_{\theta}$.
In Appendix A we discuss in more detail this extended faithfulness assumption. %To ensure faithfulness of the generated distributions to $G^+_{\theta}$,
To ensure faithfulness we discarded random instantiations of the coefficients $\mathbf{a}$ if for two events of $\mathbf{Pa}_z$ that following the generative equations of Eqs.\,\ref{e1}-\ref{e2} should correspond to a different value of $p_z$, the difference in $p_z$ was smaller than $0.05$. This ensures that the generated distributions contain all and only the structural sufficient statistics present in the functional equations, and allowed us to evaluate the performance inferring the statistics by comparing the output of the IBSSI algorithm to the known underlying sufficient statistics present in each functional equation. First, we randomly generated $\mathrm{K} = 40$ sets of parameters $\mathbf{a}$ for each type of system. Second, we used $n \in \{4, 8, 16, 64\}$ to simulate configurations with a different signal-to-noise ratio for $\mathrm{Z}$, given that for the binomial distribution the ratio of the mean and the standard deviation is proportional to $\sqrt{n}$. In Eqs.\,\ref{e1a}, \ref{e1b}, and \ref{e2b} we fixed $p(\mathrm{V}_2=1) = 0.5$. We then simulated data for $3 \times 3$ combinations of $p(\mathrm{X}=1)$ and $p(\mathrm{V}_1=1)$ with values $\{ 0.3, 0.5, 0.7\}$, so that overall we generated $40 \times 4 \times 3 \times3 = 1440$ configurations.

\subsubsection{Estimation of sufficient statistics with the Information Bottleneck algorithm}
\label{ss421}

We first study how the output $\hat{\theta}_{IB}$ of the IB method depends on its inputs $\mathbf{X}$, $\beta$, and $\max |\hat{\theta}_{IB}|$ (Figure \ref{f2}). For this purpose we examine how the cardinality $|\hat{\theta}_{IB}|$ depends on these factors, as well as how depends on them the ratio of configurations for which the criteria to assess the validity of an estimated $\hat{\theta}_{IB}$ (lines 16-18 in algorithm \ref{alg1}) are fulfilled. The overall performance of the IBSSI method, which combines outputs of the IB method across inputs parameters, will be studied subsequently.

Each row in Figure \ref{f2} studies one type of system from Figure \ref{f02}. Each panel presents the results as a function of $\max |\hat{\theta}_{IB}|$ for three values of $\beta'$. Different columns show the results for $\mathbf{X} = \{ \mathrm{X}, \mathrm{V}_1 \}$ and $\mathbf{X} = \mathbf{Pa}_z$. A sample size $N = 20000$ was used for each the fitting and testing sets. In Figure \ref{f2}A, for the configurations from Eq.\,\ref{e1a}, the statistic $\theta = \mathrm{X}+\mathrm{V}_1$ can be identified with both $\mathbf{X} = \{ \mathrm{X}, \mathrm{V}_1 \}$ and $\mathbf{X} = \{ \mathrm{X}, \mathrm{V}_1 , \mathrm{V}_2\}$. Since $\mathrm{X} \in \{0, 1\}$ and $\mathrm{V}_1 \in \{0, 1\}$, then $\theta \in \{0, 1, 2 \}$ with cardinality $|\theta|=3$. Both $\mathrm{X} =0,\mathrm{V}_1 =1$ and $\mathrm{X} =1,\mathrm{V}_1 =0$ result in $\theta =1$, so that $\mathrm{X}$ is still uncertain after conditioning on $\theta$. With $\mathbf{X} = \{ \mathrm{X}, \mathrm{V}_1\}$, $|\hat{\theta}_{IB}|=|\theta|= 3$ is correctly identified for $\max |\hat{\theta}_{IB}|$ equal $3$ or $4$, when $\beta' = 15$. When using a too high $\beta'= 500$, the algorithm becomes too sensitive to differences in probability values due to the finite sample size, and in most cases $\tilde{\mathbf{X}} = \mathbf{X}$. With $\mathbf{X} = \{ \mathrm{X}, \mathrm{V}_1 , \mathrm{V}_2\}$, since $\mathrm{V}_2$ is not involved in any statistic, the IB algorithm should return $\theta_{IB} = \{ \theta, \mathrm{V}_2\}$, whose cardinality is $|\theta_{IB}| = |\theta| |\mathrm{V}_2|=3 * 2$. Again, the valid $\hat{\theta}_{IB}$ is retrieved if $\beta'$ is not too high. %The dependence of the cardinality $|\hat{\theta}_{IB}|$ on $\max |\hat{\theta}_{IB}|$ supports the strategy used in algorithm \ref{alg1} to select sufficient statistics, namely identifying a plateau of $|\hat{\theta}_{IB}|$ when decreasing $\max |\hat{\theta}_{IB}|$.
The selection ratio indicates the ratio of configurations for which the selection criteria of lines 16-18 of algorithm 1 are fulfilled. This ratio increases until $\max |\hat{\theta}_{IB}|$ is equal to $|\theta_{IB}|$ and drops for $\beta' = 500$ when $\max |\hat{\theta}_{IB}|> |\theta_{IB}|$, reflecting that the condition on $H(\mathrm{X}|\hat{\theta}_{IB})$ (line 16) is not fulfilled when $\tilde{\mathbf{X}} = \mathbf{X}$. For $\mathbf{X} = \{ \mathrm{X}, \mathrm{V}_1 , \mathrm{V}_2\}$ this drop only happens with $\max |\hat{\theta}_{IB}|= 8$ and not $\max |\hat{\theta}_{IB}|= 7$. In the latter case, even if the full statistic with cardinality $|\theta_{IB}| = |\theta| |\mathrm{V}_2|=6$ is not identified, still a valid statistic is selected, which identifies $\theta$ only for one of the two values of $\mathrm{V}_2$ and has $|\hat{\theta}_{IB}| = 7 < |\mathbf{X}|$.

Figure \ref{f2}B, shows the results for the system of Eq.\,\ref{e1b}, with a statistic $\theta = \mathrm{X}+\mathrm{V}_1$ and an auxiliary statistic $\gamma = \mathrm{V}_1 + \mathrm{V}_2$. As mentioned in Section \ref{ss32}, the IB algorithm infers a single $\hat{\theta}_{IB}$ in the space of $\theta \otimes \gamma$. As before, $\theta \in \{0, 1, 2\}$, with $\theta =1$ for the events $\mathrm{X} =0,\mathrm{V}_1 =1$ or $\mathrm{X} =1,\mathrm{V}_1 =0$. However, now $I(\mathrm{X};\mathrm{Z}|\theta =1)>0$, because the two events differ in the value of $\mathrm{V}_1$, which also affects $\mathrm{Z}$ through $\gamma$. Only $I(\mathrm{X};\mathrm{Z}|\theta, \gamma)=0$. The underlying cardinality is $|\theta_{IB}|= |\theta \otimes \gamma|= 7$, with only $\mathrm{X}=0, \mathrm{V}_1 =1, \mathrm{V}_2 =0$ and $\mathrm{X}=1, \mathrm{V}_1 =0, \mathrm{V}_2 =1$ leading to the same value of $p_\mathrm{z}$. With an input $\mathbf{X} = \{ \mathrm{X}, \mathrm{V}_1\}$, the IB algorithm correctly indicates that there is no sufficient statistic, returning always a $\hat{\theta}_{IB}$ with cardinality $|\hat{\theta}_{IB}| = \max |\hat{\theta}_{IB}|$. The selection ratio is close to zero for all $\max |\hat{\theta}_{IB}|$. For $\max |\hat{\theta}_{IB}|<4$ the condition of line 18 is not fulfilled, that is, no statistic can be found containing all the information. For $\max |\hat{\theta}_{IB}|=4$ the algorithm returns $\tilde{\mathbf{X}} = \mathbf{X}$ and the condition of line 16 is not fulfilled. With $\mathbf{X} = \{ \mathrm{X}, \mathrm{V}_1, \mathrm{V}_2\}$, the cardinality correctly saturates at $|\hat{\theta}_{IB}|= 7$, except for a too high $\beta'$. The same behavior is observed for the configurations from Eq.\,\ref{e2a} (Figure \ref{f2}C). In this case, there is a single statistic $\theta = \mathrm{X}+\mathrm{V}_1$, but the path $\mathrm{X} \rightarrow \mathrm{Y} \rightarrow \mathrm{Z}$ requires to condition on $\mathrm{Y}$ to obtain the independence $\mathrm{Z} \perp \mathrm{X} | \theta, \mathrm{Y}$. For $\mathbf{X} = \{ \mathrm{X}, \mathrm{V}_1 \}$, with $\max |\hat{\theta}_{IB}|=4$ the average $|\hat{\theta}_{IB}|$ is below $4$ when using $\beta' = 15$, which indicates that this low $\beta'$ promotes in some cases some degree of compression despite the lack of a statistic. However, these false statistics are not accepted and the selection ratio remains close to zero, since the condition of line 18 in the algorithm is not fulfilled. For $\mathbf{X} = \{ \mathrm{X}, \mathrm{V}_1, \mathrm{Y} \}$, the statistic is correctly identified, with a saturation at $|\hat{\theta}_{IB}| = |\theta||\mathrm{Y}|= 3*2$, again if $\beta'$ is not too high. The fact that for this system a sufficient statistic is identified with $\mathbf{X} = \{ \mathrm{X}, \mathrm{V}_1, \mathrm{Y} \}$ and not with $\mathbf{X} = \{ \mathrm{X}, \mathrm{V}_1 \}$ is consistent with the requirements of the implementation of rule R.nc-ss (Section \ref{ss32}). In particular, given that $\mathrm{X} \perp \mathrm{Z}| \hat{\theta}_{IB}$ for $\mathbf{X} = \{ \mathrm{X}, \mathrm{V}_1, \mathrm{Y} \}$, the lack of a statistic for $\mathbf{X} = \{ \mathrm{X}, \mathrm{V}_1 \}$ corresponds to the requirement that using as input $\mathbf{X} \backslash \mathrm{Y}$ the output $\hat{\theta}'_{IB}$ leads to $\mathrm{X} \notperp \mathrm{Z}| \hat{\theta}'_{IB}$ or to $H(\mathrm{X}|\hat{\theta}'_{IB})=0$.

Finally, Figure \ref{f2}D shows the results for the system with no sufficient statistics (Eq.\,\ref{e2b}). Correctly, the cardinality of $\hat{\theta}_{IB}$ does not saturate, independently of the input $\mathbf{X}$ used. For a certain percentage of configurations the selection criteria are fulfilled even in the lack of a sufficient set of statistics but, as we will now see, this does not directly lead to false positives. Indeed, so far we have examined the dependence of the IB output on $\mathbf{X}$, $\beta$, and $\max |\hat{\theta}_{IB}|$, but algorithm \ref{alg1} imposes additional constraints, in particular when combining results across $\max |\hat{\theta}_{IB}|$ values (line 4 and 11), which control the acceptance of invalid statistics. We will now evaluate the overall performance of the IBSSI method.%The overall performance of the IBSSI method will be examined below.

\subsubsection{Performance of the IBSSI method identifying sufficient sets of statistics}
\label{ss422}

%The analysis in Figure \ref{f2} characterizes the output of the IB algorithm for particular selections of the inputs $\mathbf{X}$, $\beta$, and $\max |\hat{\theta}_{IB}|$.
The selection ratio examined above is a measure calculated from sampled data, which quantifies how many configurations fulfill the selection criteria, but cannot discriminate true from false selected sufficient statistics. To further evaluate performance, we used our knowledge of the form of the true underlying sufficient statistics in Eqs.\,\ref{e1}-\ref{e2}. For each type of system we pre-specified which forms of $\hat{\theta}_{IB}$ are consistent with the underlying statistics. Here consistency means that $\hat{\theta}_{IB}$ either contains the underlying sufficient statistics, or at least partially identifies them, in a way that the selection criteria are fulfilled and $\hat{\theta}_{IB}$ is valid to apply rules R.c-ss or R.nc-ss. For example, as mentioned above for the system of Eq.\,\ref{e1a} (Figure \ref{f2}A), when $\mathbf{X} = \{ \mathrm{X}, \mathrm{V}_1, \mathrm{V}_2 \}$, the full sufficient statistic has a cardinality $|\theta_{IB}| = |\theta| |\mathrm{V}_2|= 6$, but the IB algorithm may only identify $\theta = g(\mathrm{X}, \mathrm{V}_1)$ for one of the two values of $\mathrm{V}_2$, while for the other it does not compress $\{\mathrm{X}, \mathrm{V}_1\}$. In that case, although the sufficient statistic is not fully identified and $|\hat{\theta}_{IB}|=7$, the estimated statistic is still valid, since it would allow applying rule R.c-ss.

The IBSSI method combines results across $\max |\hat{\theta}_{IB}|$ values following algorithm \ref{alg1} and further combines results across values of $\beta$ and selections of $\mathbf{X}$ as described in Section \ref{ss33}. We calculated the overall true positive (TP) rates and false positive (FP) rates across all configurations as follows: For each sample size, algorithm \ref{alg1} was applied with $\mathbf{X} = \{ \mathrm{X}, \mathrm{V}_1\}$ for a whole range of $\beta'$ values. For the cases in which a true underlying sufficient set of statistics exists, a configuration produced a false positive if for any $\beta'$ value a sufficient set of statistics was accepted as valid following the selection criteria, but was inconsistent with the true underlying one. When no underlying statistics exist, any accepted statistics were considered a false positive. A configuration produced a true positive if there was some $\beta'$ value for which a sufficient set of statistics was accepted as valid following the selection criteria and, for those accepted as valid, they all were consistent with true underlying statistics. As mentioned above, here by consistent we mean that $\hat{\theta}_{IB}$ either corresponds to the sufficient statistics, or identifies a compressed $\tilde{\mathbf{X}}$ that, although only partially identifying the structural sufficient statistics, still fulfills the selection criteria. For those configurations that using $\mathbf{X} = \{ \mathrm{X}, \mathrm{V}_1\}$ were not already accounted as false positives or true positives, we repeated the same procedure with the corresponding enlarged $\mathbf{X}$ \--with $\mathbf{X} = \{ \mathrm{X}, \mathrm{V}_1, \mathrm{V}_2\}$ for Eqs.\,\ref{e1} and \ref{e2b}, and $\mathbf{X} = \{ \mathrm{X}, \mathrm{V}_1, \mathrm{Y}\}$ for Eq.\,\ref{e2a}. Overall FP and TP rates were calculated from the total number of false and true positives over the two choices of $\mathbf{X}$.

\begin{figure*}
  \begin{center}
    \scalebox{0.38}{\includegraphics*{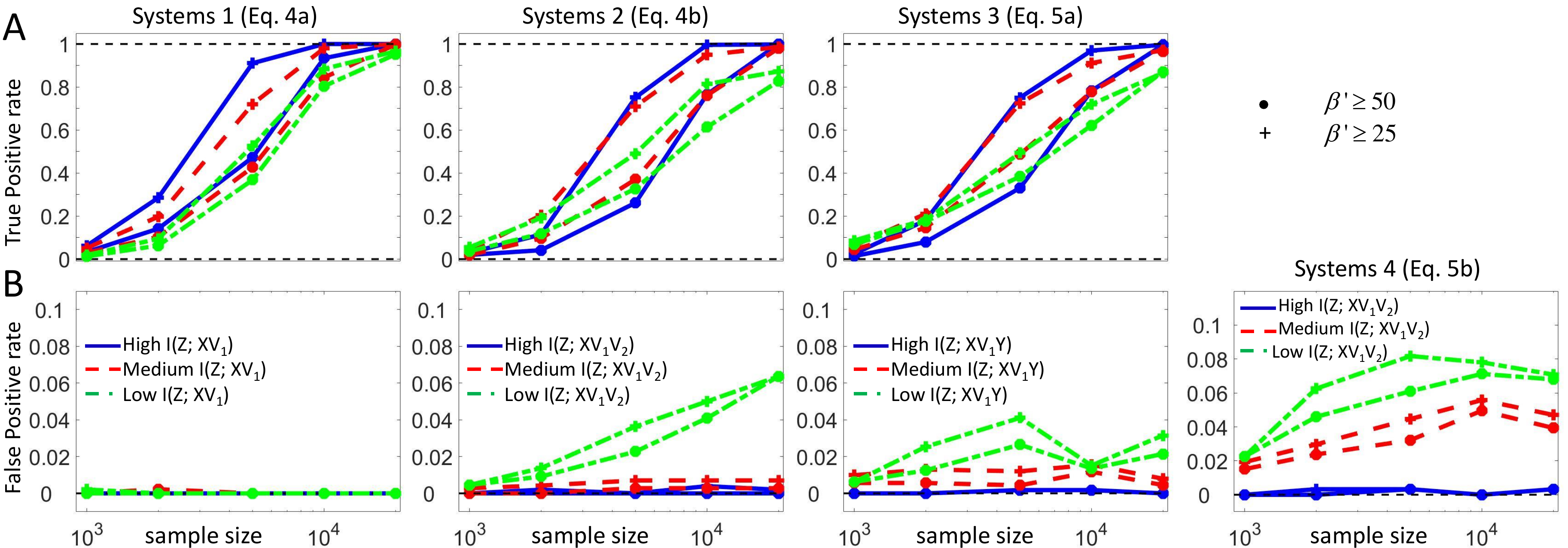}}
  \end{center}
  %\vspace{0.25in}
  \caption{Performance of the IBSSI method for the systems of Figure \ref{f02}. \textbf{A}) True positive (TP) rate identifying valid sufficient statistics in the systems from Eqs.\,\ref{e1a}, \ref{e1b}, and \ref{e2a}. \textbf{B}) False positive (FP) rate for those systems and the one of Eq.\,\ref{e2b}. Results are shown as a function of the sample size and configurations are grouped in levels of information about $\mathrm{Z}$ (see main text and Appendix D.2 for details on the stratification procedure). For systems 1 (Eq.\,\ref{e1a}), given $\theta = g(\mathrm{X}, \mathrm{V}_1)$, $I'(\mathrm{X} \mathrm{V}_1 ; \mathrm{Z}) \equiv I(\mathrm{X} \mathrm{V}_1; \mathrm{Z})/ H(\mathrm{Z})$ was used for binning. The average information at each level is: Low: $\langle I'(\mathrm{X} \mathrm{V}_1; \mathrm{Z}) \rangle = 0.03$. Medium: $\langle I'(\mathrm{X} \mathrm{V}_1; \mathrm{Z}) \rangle = 0.07$. High: $\langle I'(\mathrm{X} \mathrm{V}_1; \mathrm{Z}) \rangle = 0.14$. For systems 2 (Eq.\,\ref{e1b}), given $\theta = g(\mathrm{X}, \mathrm{V}_1)$ and $\gamma = g(\mathrm{V}_1, \mathrm{V}_2)$, $I'(\mathrm{X} \mathrm{V}_1 \mathrm{V}_2; \mathrm{Z}) \equiv I(\mathrm{X} \mathrm{V}_1 \mathrm{V}_2; \mathrm{Z})/ H(\mathrm{Z})$ was used for binning. Averages are: Low: $\langle I'(\mathrm{X} \mathrm{V}_1 \mathrm{V}_2; \mathrm{Z}) \rangle = 0.12$. Medium: $\langle I'(\mathrm{X} \mathrm{V}_1 \mathrm{V}_2; \mathrm{Z}) \rangle = 0.20$. High: $\langle I'(\mathrm{X} \mathrm{V}_1 \mathrm{V}_2; \mathrm{Z}) \rangle = 0.28$. For systems 3 (Eq.\,\ref{e2a}), $I'(\mathrm{X} \mathrm{V}_1 \mathrm{Y}; \mathrm{Z}) \equiv  I(\mathrm{X} \mathrm{V}_1 \mathrm{Y}; \mathrm{Z})/ H(\mathrm{Z})$ was used, since conditioning on $\theta = g(\mathrm{X}, \mathrm{V}_1)$ and $\mathrm{Y}$ is required to render $\mathrm{X}$ and $\mathrm{Z}$ independent. Averages are: Low: $\langle I'(\mathrm{X} \mathrm{V}_1 \mathrm{Y}; \mathrm{Z}) \rangle = 0.08$. Medium: $\langle I'(\mathrm{X} \mathrm{V}_1 \mathrm{Y}; \mathrm{Z}) \rangle = 0.15$. High: $\langle I'(\mathrm{X} \mathrm{V}_1 \mathrm{Y}; \mathrm{Z}) \rangle = 0.25$. For systems 4 (Eq.\,\ref{e2b}), $I'(\mathrm{X} \mathrm{V}_1 \mathrm{V}_2; \mathrm{Z}) \equiv I(\mathrm{X} \mathrm{V}_1 \mathrm{V}_2; \mathrm{Z})/ H(\mathrm{Z})$ was used, since no sufficient statistics exist. Averages are: Low: $\langle I'(\mathrm{X} \mathrm{V}_1 \mathrm{V}_2; \mathrm{Z}) \rangle = 0.12$. Medium: $\langle I'(\mathrm{X} \mathrm{V}_1 \mathrm{V}_2; \mathrm{Z}) \rangle = 0.20$. High: $\langle I'(\mathrm{X} \mathrm{V}_1 \mathrm{V}_2; \mathrm{Z}) \rangle = 0.28$. The results are shown for two different sets of $\beta'$ values. Note the different scale of the y-axis for the TP and FP rates.}
  \label{f3}
\end{figure*}

To examine the TP and FP rates we stratified the configurations by levels of information. For each type of system, independently of the input $\mathbf{X}$ used, the stratification was based on the normalized mutual information about $\mathrm{Z}$ carried by the minimal input $\mathbf{X}$ resulting in a sufficient set of statistics, or on the largest $\mathbf{X}$ in the case of Figure \ref{f2}D where no statistics exist. See the caption of Figure \ref{f3} for the average information values at each level for each system, and Appendix D.2 for details on the stratification procedure. Note however that the stratification in information levels is not part of the IBSSI method and we only use it to illustrate how information modulates performance. For the systems studied in this section the normalized information values are rather low (averages in the range $[0.03 - 0.28]$), which means that the performance of the IBSSI method is evaluated in demanding cases. See Appendix E.3 for examples covering a wider range of information levels, up to $0.55$.

Figure \ref{f3} shows true positive (TP) rates for the three types of systems containing sufficient statistics, and false positive (FP) rates for all systems. We present the results for two sets of $\beta'$ values, namely $\beta' \in \{ 25, 50, 75, 100\}$ and $\beta' \in \{ 50, 75, 100\}$. The TP rate increases with $N$, and for all types of systems when $N$ is high it is higher for the configurations with higher information. For low $N$, a flip between configurations with high and low information occurs for the systems of Eqs.\,\ref{e1b} and \ref{e2a}. This flip can be understood taking into account how the configurations have been generated, selecting $n \in \{4, 8, 16, 64\}$ for the binomial distribution $p(\mathrm{Z}| \mathbf{Pa}_z)$. The signal-to-noise ratio increases with $\sqrt{n}$, which means that configurations with higher $n$ also tend to have higher information. However, a higher $n$ also implies a poorer sampling of the distribution $p(\mathrm{Z}, \mathbf{X})$, for a given $N$. While a higher information is expected to facilitate the inference of the statistic, a poorer sampling is expected to hinder it (see Appendix E.5 for details). Furthermore, the TP rate increases faster with the set that includes $\beta'= 25$. This is because, with a small sample size, high $\beta'$ values enforce the preservation of differences in the probabilities related to sampling fluctuations and hence lead to selecting $\tilde{\mathbf{X}} = \mathbf{X}$. FP rates are generally low for all types of systems. For the systems of Eqs.\,\ref{e1} and \ref{e2a}, false positives refer to sufficient statistics that are accepted as valid which do not correspond to the underlying functional ones, while for the systems of Eq.\,\ref{e2b} any sufficient statistic accepted is a false positive. As expected, the increased TP rates when including the lower $\beta'= 25$ are accompanied by some (but small) increase in the FP rates. The information level also influences the FP rate. This suggests that imposing a lower bound on information levels to select sufficient statistics can be helpful to reduce false positives. This may be particularly important in cases in which the application of rules R.c-ss and R.nc-ss enables the posterior application of some standard rules of causal orientation, such that the effect of a false positive may propagate.

As explained at the beginning of this section, we have focused on the evaluation of the IBSSI method instead of evaluating a full application of rules R.c-ss and R.nc-ss in order to isolate the performance identifying sufficient statistics from other factors that are not specific to rules R.c-ss and R.nc-ss but also affect their applicability. Since the IBSSI method allows the creation of a new conditional independence with the identified sufficient statistics, our results support its utility to implement the rules as described in Section \ref{ss32}. Overall, these examples show that the IB algorithm is able to identify sufficient statistics with good performance and without \emph{a priori} assumptions of which statistics exist. Besides the concrete performance values obtained, most importantly this analysis allowed us to discuss how the IBSSI procedure combines outputs from the IB method across a range of different parameters, and allowed us to characterize the factors that affect performance. The core requirement for the validity of a sufficient set of statistics is the fulfillment of the selection criteria in lines 16-18 of algorithm 1. Modifying the thresholds, especially $a_I$, can control the FP rate. The condition of line 15 checks the consistency of the statistics found across a range of $\max |\hat{\theta}_{IB}|$, and this range can also be widen to be more restrictive. %The range of $\beta'$ values combined also affects the tradeoff between TP and FP rates.
See Appendix D.3 for a further discussion of possible adjustments of the IBSSI method. In Appendix E we provide further examples, with systems in which the set of statistics contains multiple statistics, in which the statistics have a different form, or the generative mechanism of $\mathrm{Z}$ is different.

\subsection{Sufficient statistics in biologically-plausible Boolean regulatory rules}
\label{ss43}

We here focus on a concrete set of functional equations that have been shown \citep{Li06} to accurately model a biological network, namely the signal transduction network of the hormone abscisic acid in guard cells of plants. \cite{Li06} studied a dynamic model of the signaling process, with the temporal updating of the variables in the system governed by Boolean regulatory rules (Table 1 in \cite{Li06}). Data simulated from these rules has also been included in the workbench of the Causation and Prediction Challenge of \cite{Guyon08}, in which the objective was to reconstruct the causal structure of the underlying system \citep{Jenkins08}. Our aim here is not to address the reconstruction of the whole causal structure \--for which rules R.c-ss and R.nc-ss would need to be embedded in a concrete implementation of structure learning algorithm (see Appendix C)\--, but to examine how sufficient statistics present in the Boolean regulatory rules can be identified. We envisage a scenario in which a system governed by this type of Boolean rules is only partially observed, or there are unknown arguments of the rules, as it is often the case \citep{Geier07, Braga18}. For this purpose, we selected the Boolean rules from \cite{Li06} containing at least three arguments (Figure \ref{fB1}A) and simulated all combinations in which only two of the arguments are observable (see Appendix E.4 for further analysis of cases with three arguments observable). The setting studied corresponds to the causal structure of Figure \ref{fB1}B, where $\mathrm{X}$ and $\mathrm{V}_1$ denote the two observable arguments and $\mathrm{U}_1...\mathrm{U}_n$ the rest of arguments in each Boolean expression. All variables are binary, with values $0$ and $1$. We randomly generated $\mathrm{X}$ and the hidden variables with independent uniform distributions. We modeled the link $\mathrm{X} \rightarrow \mathrm{V}_1$ introducing a dependence $p(\mathrm{V}_1=1|\mathrm{x})= 0.4 + 0.2 \mathrm{x}$. We then generated $\mathrm{Z}$ following the Boolean rules. Using $\mathbf{X} = \{\mathrm{X}, \mathrm{V}_1\}$, we studied the inference of the existing sufficient statistics with the IB algorithm.

\begin{figure*}[t]
  \begin{center}
    \scalebox{0.45}{\includegraphics*{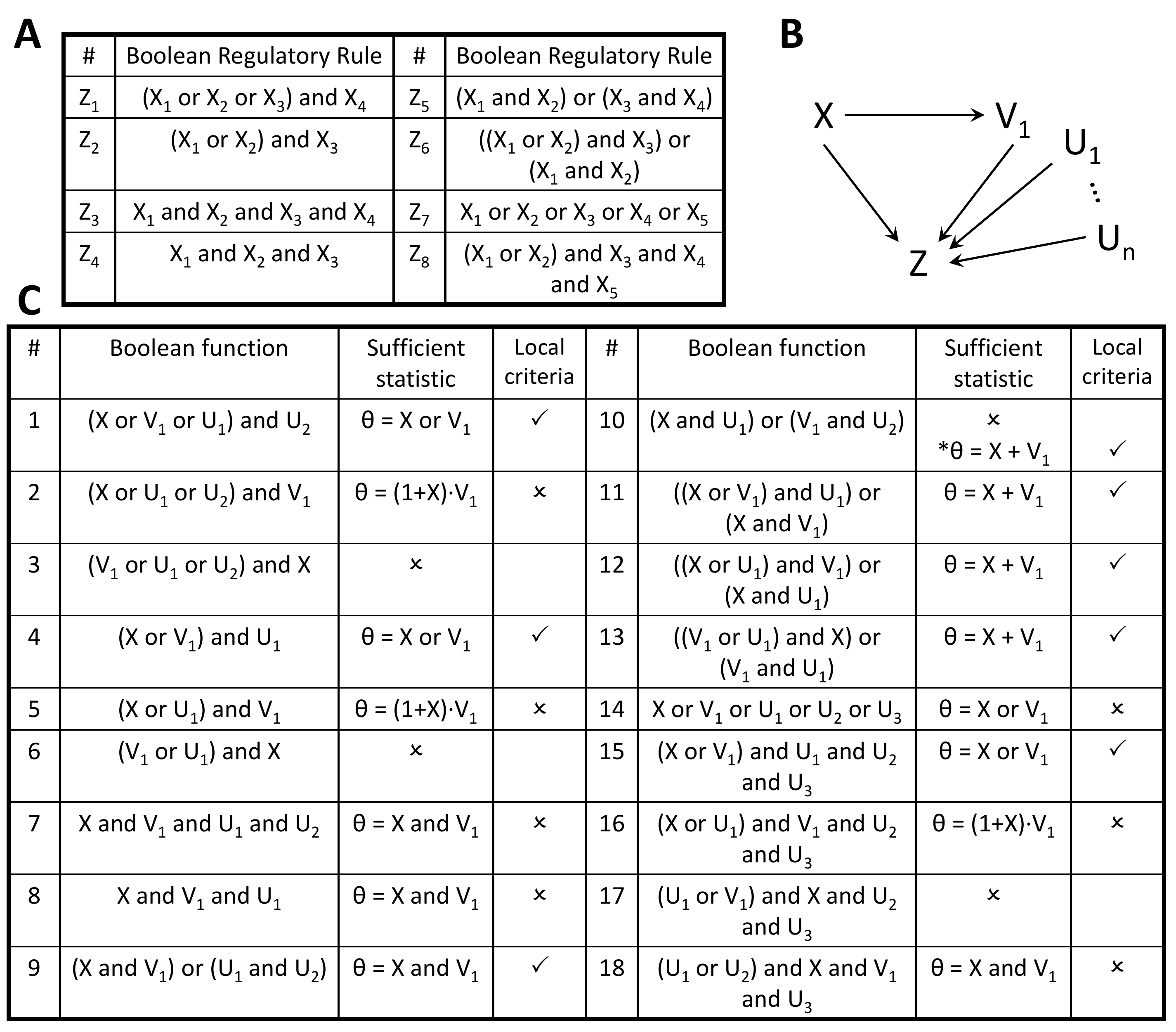}}
  \end{center}
  %\vspace{0.25in}
  \caption{Boolean regulatory rules containing sufficient statistics. \textbf{A}) Rules modeling the signal transduction network studied in \cite{Li06}. The identity of the original variables is not relevant for our analysis and is here omitted \--see caption of Figure S5 for the correspondence with Table 1 in \cite{Li06}. \textbf{B}) Causal structure corresponding to the setting in which we examine the identification of sufficient statistics. Variables $\mathrm{X}$ and $\mathrm{V}_1$ correspond to the only two observable arguments of a Boolean rule, and $\mathrm{U}_1...\mathrm{U}_n$ to the rest of arguments. \textbf{C}) Configurations corresponding to all combinations of pairs of observable variables from the rules in A). In each case we express the Boolean function in terms of the observable and hidden variables, we show the form of the sufficient statistic, when it exists, and we indicate whether the local selection criteria (see text) are fulfilled.}
  \label{fB1}
\end{figure*}

\begin{figure*}[t]
  \begin{center}
    \scalebox{0.6}{\includegraphics*{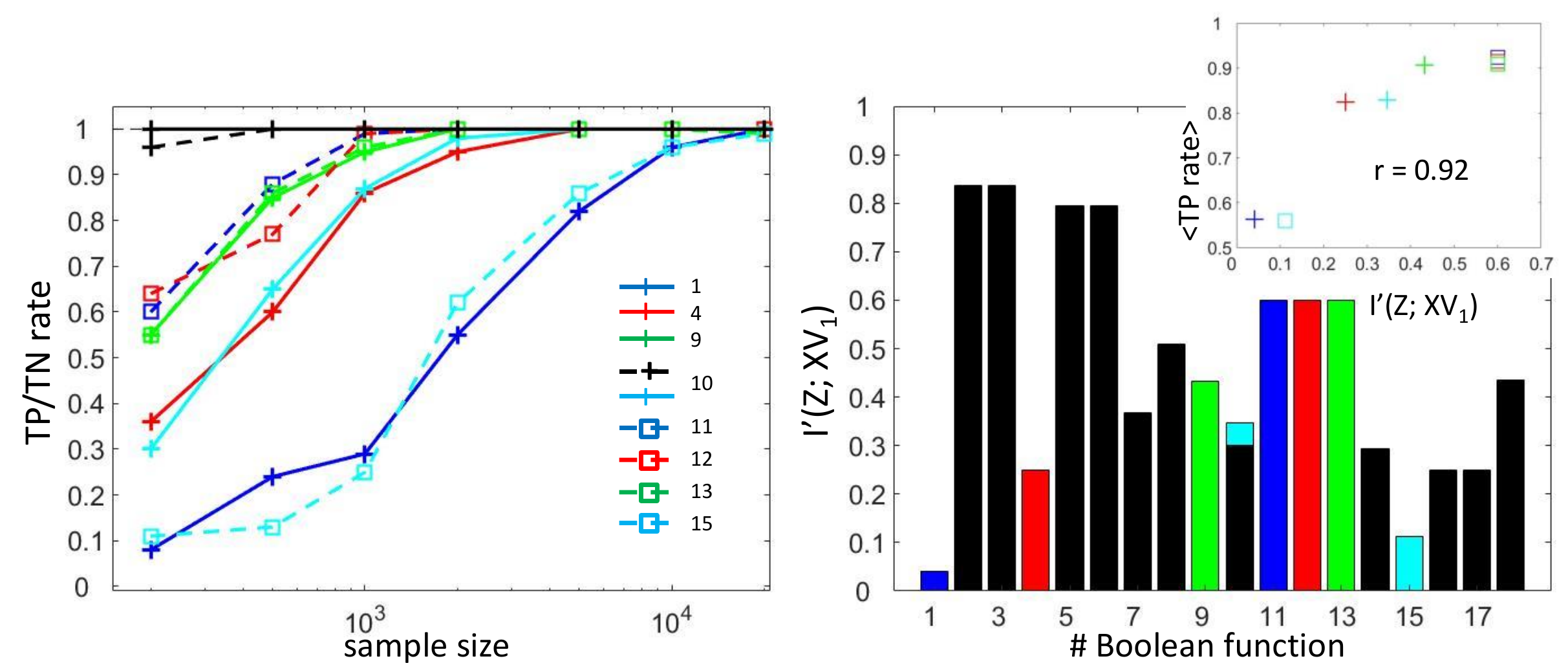}}
  \end{center}
  %\vspace{0.25in}
  \caption{Performance of the IBSSI method identifying sufficient statistics in biologically-plausible Boolean regulatory rules. \textbf{A}) True positive (TP) rates and True negative (TN) rates, for the configurations of Figure \ref{fB1}C containing and not containing sufficient statistics, respectively. For the configurations with no sufficient statistic the TN rates are represented with a solid black line, and all overlap except for configuration 10 (dashed black line) for the lowest sample size. For the configurations with a sufficient statistic the legend numbers the lines in correspondence to Figure \ref{fB1}C. Configuration 10 was both analyzed for a setting producing a statistic or not. \textbf{B}) Normalised information between $\mathrm{Z}$ and the observable variables for each configuration. For the configurations with a sufficient statistic the TP rate is highly correlated with the normalised information.}
  \label{fB2}
\end{figure*}

Figure \ref{fB1}C details all combinations of pairs of observed variables obtained from the rules in Figure \ref{fB1}A. When it exists, we display the form of the sufficient statistic. Note that the sufficient statistics $\theta = \mathrm{X}\,\mathrm{AND}\,\mathrm{V}_1$ and $\theta = \mathrm{X}\,\mathrm{OR}\,\mathrm{V}_1$ could also be expressed as threshold-logic expressions $\mathrm{X} + \mathrm{V}_1 > 1$ and $\mathrm{X} + \mathrm{V}_1 > 0$, respectively, which constitute basic components of artificial neural networks \citep{Mcculloch43}. For configuration 10, a sufficient statistic exists only in the specific case of $p(\mathrm{U}_1)= p(\mathrm{U}_2)$, and we examine both the general and this specific case. Since in the rules $\mathrm{Z}$ is deterministically determined by its parents, we further strengthen the selection criteria of algorithm \ref{alg1} to ensure that, at least for some instantiation $\theta = \theta_0$ of the sufficient statistics embodied in the Boolean functions, neither $\mathrm{Z}$ nor $\mathrm{X}$ is fully determined. The necessity of these additional local criteria can be understood considering the concrete example of configuration $\sharp 2$ in Figure \ref{fB1}C. In this case the form of the sufficient statistic is $\theta = (1+\mathrm{X})\mathrm{V}_1$ and $\theta$ has three values: $\theta =0$ whenever $\mathrm{V}_1=0$ \--given the $\mathrm{AND}$ %`$\mathrm{and}$'
operation\--, and $\theta = 1,2$ when $\mathrm{V}_1=1$ and $\mathrm{X}= 0,1$, respectively. However, $\theta =2$ deterministically determines $\mathrm{X} = \mathrm{Z} = 1$, a value $\theta =0$ does not determine $\mathrm{X}$ but leads deterministically to $\mathrm{Z}=0$, while $\theta = 1$ leaves some uncertainty about $\mathrm{Z}$ given the hidden variables, but univocally determines $\mathrm{X}=0$. That is, the conditions of lines 16-17 in algorithm \ref{alg1} are fulfilled, but there is no $\theta = \theta_0$ for which simultaneously $H(\mathrm{Z}|\theta_0)>0$ and $H(\mathrm{X}|\theta_0)>0$. The additional local criteria demand the existence of such $\theta_0$, since otherwise it is not possible to evaluate the independence $\mathrm{X} \perp \mathrm{Z} |\theta_0 $ or other dependencies and independencies used in rules R.c-ss and R.nc-ss. From all configurations of Figure \ref{fB1}C, a valid sufficient statistic exists for 8 configurations. From the rest, only in 4 cases there is no sufficient statistic, while the validity of the sufficient statistic is rejected based on local criteria for other 7 configurations. While here following \cite{Li06} we study Boolean deterministic rules and limit the source of stochasticity to the hidden arguments of the rules, alternative models have been proposed, e.\,g.\,for gene regulatory networks, which combine Boolean expressions with explicit additional sources of stochasticity \citep{Shmulevich02, Ruczinski04, Dehghannasiri18}. The sufficient statistics discarded based on local criteria could be also exploited in the presence of such additional stochasticity.

Figure \ref{fB2}A shows true positive (TP) rates for the configurations with a sufficient statistic, and true negative (TN) rates for the configurations without statistic, respectively. The rates were calculated over 100 independent simulations for each sample size. As seen from the TN rates, for these examples the IBSSI method does not produce false positives, except in few cases with the lowest sample size for configuration 10, for which indeed a sufficient statistic exists in a concrete setting, namely $p(\mathrm{U}_1)= p(\mathrm{U}_2)$. For the configurations containing a sufficient statistic, the TP rate increases with $N$ and is highly correlated with the normalised information $I(\mathrm{Z}; \mathrm{X} \mathrm{V}_1)/H(\mathrm{Z})$ (Figure \ref{fB2}B).

\section{Selection Bias}
\label{ss5}

\begin{figure*}[t]
  \begin{center}
    \scalebox{0.55}{\includegraphics*{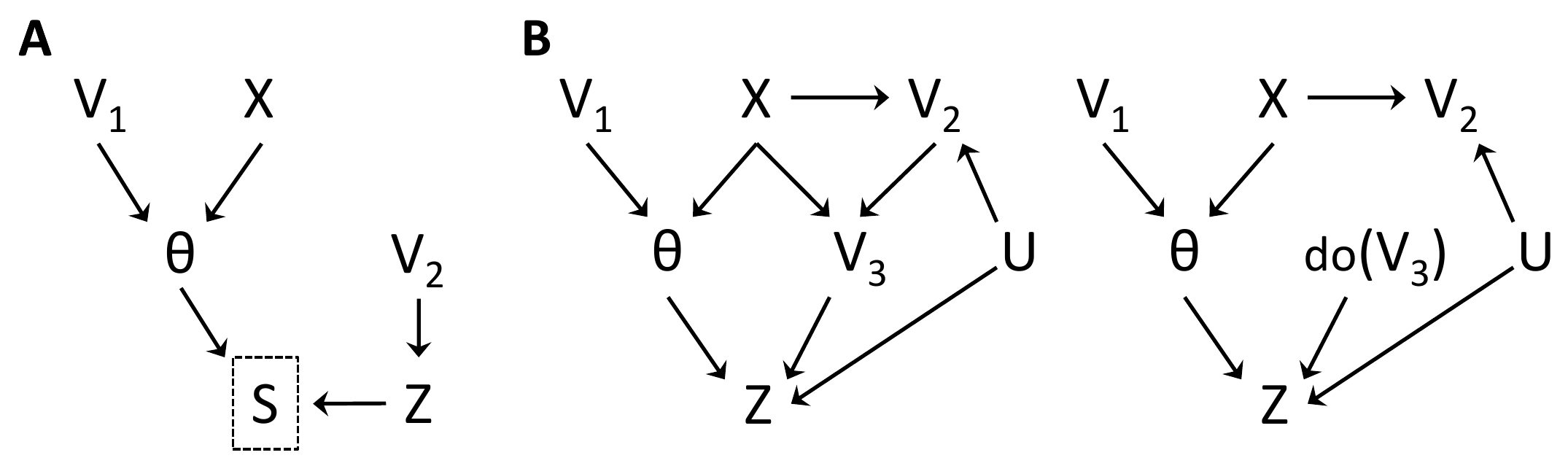}}
  \end{center}
  %\vspace{0.25in}
  \caption{Causal structures studied to assess the performance of the IBSSI method finding sufficient statistics in the presence of selection bias or dormant independencies. \textbf{A}) Causal structure in which the dependence between $\mathrm{X}, \mathrm{V}_1$ and $\mathrm{Z}$ is due to bias selecting $\mathrm{S}$, for which a sufficient statistic $\theta = g(\mathrm{X}, \mathrm{V}_1)$ exists. Variable $\mathrm{S}$ introduces a selection bias and is not necessarily observable. \textbf{B}) Causal structure in which the sufficient statistic $\theta = g(\mathrm{X}, \mathrm{V}_1)$ only results in a conditional independence between $\mathrm{X}$ and $\mathrm{Z}$ when the system is intervened with $do(\mathrm{V}_3)$ (dormant independence). In the unintervened system conditioning on $\mathrm{V}_3$ activates a dependence between $\mathrm{X}$ and $\mathrm{Z}$ through the hidden variable $\mathrm{U}$. The intervention $do(\mathrm{V}_3)$ eliminates all incoming arrows to $\mathrm{V}_3$ and creates the new independence $\mathrm{X} \perp \mathrm{Z} | \theta, do(\mathrm{V}_3)$. The identifiability of the intervention allows applying the IB algorithm to the intervened system. For simplicity, an additional variable $\mathrm{Y}$ is omitted, but in the presence of the causal links $\mathrm{X} \rightarrow \mathrm{Y} \leftarrow \mathrm{Z}$ the sufficient statistic would allow applying R.c-ss, and R.nc-ss could be applied in the presence of $\mathrm{X} \rightarrow \mathrm{Y} \rightarrow \mathrm{Z}$.}
  \label{f4}
\end{figure*}

Standard algorithms of structure learning exploiting conditional independencies have also been extended to systems observed under selection bias \citep{Spirtes96}. When selection bias is present, dependencies between variables otherwise independent may appear because of a constraint introduced in the way a common descendant of those variables is sampled. For example, in Figure \ref{f4}A, $\mathrm{S}$ represents a variable directly related to the selection bias. With no selection bias $\mathrm{X} \perp \mathrm{Z}$, but if the sampling process depends on $\mathrm{S}$, a dependence is introduced according to $\mathrm{X} \notperp \mathrm{Z} | \mathrm{S}$. To introduce a bias it is not necessary that $\mathrm{S}$ takes the same value for all samples, it suffices that the sampled distribution $p^*(\mathrm{S})$ differs from the one determined by its parents, $p(\mathrm{S}| \mathrm{X}, \mathrm{Z}, \mathrm{V}_1)$. Note that here $\mathrm{S}$ is an actual variable of the system, observable or not, as opposed to additional \emph{selection variables} of the type $\mathrm{S}_\mathrm{V}$ defined in \cite{Spirtes96}, which are added to represent sampling properties of associated observable variables $\mathrm{V}$. If $\mathrm{S}$ is observable, a sufficient statistic like the one in Figure \ref{f4}A can be identified using the IB algorithm to determine $\theta$ such that $\mathrm{S} \perp \mathrm{X}| \theta$. Furthermore, $\theta$ can be identified even when $\mathrm{S}$ introduces selection bias and is not observable. This is because apart from $\mathrm{S} \perp \mathrm{X}| \theta$ the sufficient statistic also creates the independence $\mathrm{Z} \perp \mathrm{X}| \theta, \mathrm{S}$. In Figure \ref{f4} we have omitted for simplicity any variable $\mathrm{Y}$ as involved in rules R.c-ss and R.nc-ss, but it is clear that the presence of $\theta$ could be exploited to apply the rules. For a system containing the structure of Figure \ref{f4}A and $\mathrm{X} \rightarrow \mathrm{Y} \leftarrow \mathrm{Z}$, R.c-ss would be applicable, and for a system containing that structure and $\mathrm{X} \rightarrow \mathrm{Y} \rightarrow \mathrm{Z}$, R.nc-ss would be applicable, in that case with $\mathrm{Z} \perp \mathrm{X}| \theta, \mathrm{S}, \mathrm{Y}$.

To study examples of the identification of sufficient statistics in the presence of selection bias we generated a system with the causal structure of Figure \ref{f4}A. The variable $\mathrm{S}$ subjected to selection bias was generated with a binomial GLM, analogously to how we generated $\mathrm{Z}$ for the systems of Eqs.\,\ref{e1}-\ref{e2}. In particular, $\mathrm{S} \sim B(n,p_\mathrm{s})$, with $p_\mathrm{s} = 1/(1+\mathrm{exp}(-h(\mathbf{Pa}_s)))$, where $\mathbf{Pa}_s = \{ \mathrm{X}, \mathrm{V}_1, \mathrm{Z}\}$. The function $h(\mathbf{Pa}_s)$ was modeled as:
\begin{equation}
\begin{split}
h &=  a_0+ a_1 \mathrm{Z} + a_2(\mathrm{X}+\mathrm{V}_1) + a_3 (\mathrm{X}+\mathrm{V}_1)^2  + a_4(\mathrm{X}+\mathrm{V}_1)\mathrm{Z} + a_5(\mathrm{X}+\mathrm{V}_1)^2 \mathrm{Z}.
\label{e3}
\end{split}
\end{equation}

As in previous examples, we used $n \in \{4, 8, 16, 64\}$ to simulate configurations with different signal-to-noise ratio. Variables $\mathrm{X}$, $\mathrm{V}_1$, and $\mathrm{Z}$ are binary independent variables with values $0,1$, but the selection bias introduces a dependence between them. $\mathrm{S}$ is not observable, and a selection bias is introduced sampling only the observable variables when $\mathrm{S}$ is in the range $\max(\mathrm{S})/3 < \mathrm{S} <  2\max(\mathrm{S})/3$. We refer to this range as $\{\mathrm{S}_0\}$. We modeled $\mathrm{V}_2 \rightarrow \mathrm{Z}$ by $p(\mathrm{Z} = 1| \mathrm{V}_2 = \mathrm{v}_2) = 0.4 + 0.2 \mathrm{v}_2$. The selection bias creates a dependence between $\mathrm{Z}$ and $\mathrm{X}$, $\mathrm{V}_1$, which given the form of Eq.\,\ref{e3} results in a sufficient statistic $\theta = \mathrm{X}+\mathrm{V}_1$ for $\mathrm{X}$ with respect to $\mathrm{Z}$ (Fig.\,\ref{f4}A). In contrast to the sufficient statistics studied so far, here $\theta$ is not a functional sufficient statistic in the functional equation of $\mathrm{Z}$ or $\mathrm{X}$, but in the functional equation of $\mathrm{S}$, which is not observable. Both $\mathrm{X} =1,\mathrm{V}_1 =0$ and $\mathrm{X} =0,\mathrm{V}_1 =1$ result in $\theta =1$, and $\theta = \mathrm{X} + \mathrm{V}_1 \in \{0, 1, 2 \}$, with cardinality $|\theta|= 3$. Like for the previous examples, we randomly sampled the coefficients $\mathbf{a}$ to generate $1440$ configurations. Again, we ensured that the generated distributions were faithful to the corresponding graph $G^+_{\theta}$ so that knowledge of the underlying statistics could be used to evaluate the performance of the IBSSI method. First, to ensure the standard faithfulness condition between the distribution and the corresponding graph $G$, we required that the selection bias actually creates a dependence between $\mathrm{Z}$ and $\mathrm{X},\mathrm{V}_1$, imposing a lower bound of $I(\mathrm{Z}; \mathrm{X}, \mathrm{V}_1| \{\mathrm{S}_0\})/ H(\mathrm{Z}| \{\mathrm{S}_0\})\geq 0.05$. Second, because the sufficient statistic is to be inferred based on $\mathrm{Z} \perp \mathrm{X}| \theta, \{\mathrm{S}_0\}$ %instead of on $\mathrm{S} \perp \mathrm{X}| \theta$,
we required that for two events of $\{ \mathrm{X}, \mathrm{V}_1\}$ that following the generative equation of Eq.\,\ref{e3} correspond to a different value of $\theta =  \mathrm{X} + \mathrm{V}_1 $, the difference in $p(\mathrm{Z}| \mathrm{X}, \mathrm{V}_1, \{\mathrm{S}_0\})$ was not smaller than $0.05$.

\begin{figure*}[t]
  \begin{center}
    \scalebox{0.38}{\includegraphics*{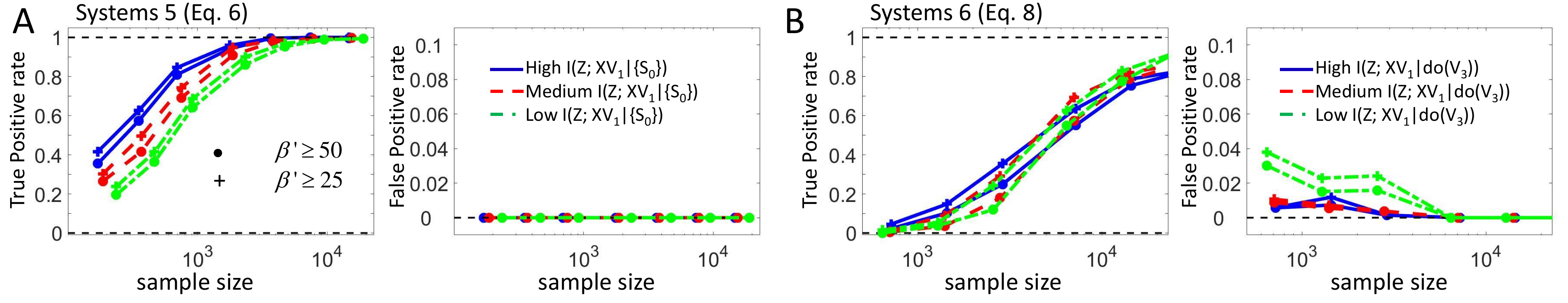}}
  \end{center}
  %\vspace{0.25in}
  \caption{Performance of the Information Bottleneck Sufficient Statistics Inference (IBSSI) method in the presence of selection bias (Eq.\,\ref{e3}) or dormant independencies (Eq.\,\ref{e4}). In both cases we show the true positive rates and false positive rates identifying valid sufficient statistics. Results are shown as a function of the sample size and for two ranges of $\beta'$ values. Configurations are grouped by information levels. \textbf{A}) Systems with selection bias. The selection bias limits observations to samples with $\mathrm{S}$ in the range $\{\mathrm{S}_0\}$ defined by $\max(\mathrm{S})/3 < \mathrm{S} <  2\max(\mathrm{S})/3$. Information levels are determined by $I'(\mathrm{X} \mathrm{V}_1; \mathrm{Z}| \{\mathrm{S}_0\}) \equiv I(\mathrm{X} \mathrm{V}_1; \mathrm{Z}| \{\mathrm{S}_0\})/ H(\mathrm{Z}| \{\mathrm{S}_0\})$, since $\theta = g(\mathrm{X}, \mathrm{V}_1)$. The average information at each level is: Low: $\langle I'(\mathrm{X} \mathrm{V}_1; \mathrm{Z}| \{\mathrm{S}_0\}) \rangle = 0.08$. Medium: $\langle I'(\mathrm{X} \mathrm{V}_1; \mathrm{Z}| \{\mathrm{S}_0\}) \rangle = 0.12$. High: $\langle I'(\mathrm{X} \mathrm{V}_1; \mathrm{Z}| \{\mathrm{S}_0\}) \rangle = 0.20$. \textbf{B}) Systems with dormant independencies. The intervention $do(\mathrm{V}_3)$ creates the new independence $\mathrm{X} \perp \mathrm{Z} | \theta, do(\mathrm{V}_3)$. Information levels are determined by $I'(\mathrm{X} \mathrm{V}_1; \mathrm{Z}| do(\mathrm{V}_3=1)) \equiv I(\mathrm{X} \mathrm{V}_1; \mathrm{Z}| do(\mathrm{V}_3=1 ))/ H(\mathrm{Z}|  do(\mathrm{V}_3=1))$, since $\theta = g(\mathrm{X}, \mathrm{V}_1)$ and the system is intervened with $ do(\mathrm{V}_3= 1)$.  The average information at each level is: Low: $\langle I'(\mathrm{X} \mathrm{V}_1; \mathrm{Z}| do(\mathrm{V}_3= 1)) \rangle = 0.08$. Medium: $\langle I'(\mathrm{X} \mathrm{V}_1; \mathrm{Z}| do(\mathrm{V}_3= 1)) \rangle = 0.13$. High: $\langle I'(\mathrm{X} \mathrm{V}_1; \mathrm{Z}| do(\mathrm{V}_3= 1)) \rangle = 0.18$.}
  \label{f6}
\end{figure*}

Figure \ref{f6}A shows the overall TP and FP rates obtained with the IBSSI method. % A more detailed analysis of the factors affecting the identification of the statistic, analogous to the one of Figure \ref{f2}, is left for Appendix E.4.
Again, a high TP rate is achievable with a low FP rate. Also here performance depends on the information levels, which were stratified by $I(\mathrm{Z}; \mathrm{X}, \mathrm{V}_1 | \{\mathrm{S}_0\})/ H(\mathrm{Z}| \{\mathrm{S}_0\})$. A smaller difference was found between selecting a range $\beta' \in \{ 25, 50, 75, 100\}$ or $\beta' \in \{ 50, 75, 100\}$. These results indicate that the identification of sufficient sets of statistics can be a powerful tool to counteract the impact of selection bias, allowing the recovery of a conditional independence between variables for which the selection bias introduced a dependence.

\section{Sufficient statistics in identifiable intervened systems}
\label{ss6}

So far we have examined examples considering that the whole causal structure of the system was unknown. In a more general setting, the causal structure may already be partially known, either because it can be partially inferred with the standard FCI algorithm, or thanks to side information, which can comprise interventional data \citep{Bareinboim16}. Similarly, when the aim is not to infer the causal structure but to identify a causal effect, it is most commonly assumed that the causal structure is known \citep{Tian02, Shpitser08c}, or concrete hypotheses about the causal structure are confronted. In these scenarios, finding sufficient statistics can also be useful either to further learn the causal structure or to identify causal effects. When parts of the causal structure are known it is possible to exploit also sufficient statistics that only lead to new conditional independencies when the system is intervened. An intervention of variable $\mathrm{V}_i$, denoted by $do(\mathrm{V}_i)$, corresponds to an external modification of the mechanisms of the system such that the functional equation of $\mathrm{V}_i$ is replaced by enforcing externally a certain value $\mathrm{V}_i = \mathrm{v}_i$ ($do(\mathrm{V}_i= \mathrm{v}_i )$). Under certain conditions on the form of the causal structure \citep{Tian02, Shpitser08c}, it is possible to calculate the distribution of the variables in the intervened system from the distribution in the observed system. Graphically, an intervention corresponds to removing the incoming arrows to the intervened variable from its parents. The removal of these arrows eliminates certain paths between variables, and this may lead to new conditional independencies only present in the intervened system \--known as \emph{dormant independencies} \citep{Shpitser08}.

Consider the causal structure of Figure \ref{f4}B and assume that the structure is already known, except for the presence of the sufficient statistic $\theta$. That is, only the associated graph $G$, but not $G^+_{\theta}$, is known. This causal structure may be embedded in a larger system, with other parts of its structure yet to be inferred, such that identifying the sufficient statistic may help to further infer it. In the graph, $\mathrm{U}$ represents a variable whose presence is known, but which is not observable. Despite the presence of the functional statistic $\theta = g(\mathrm{X}, \mathrm{V}_1)$, it is not possible to find a set $\mathbf{S}$ such that $\mathrm{X} \perp \mathrm{Z} | \mathbf{S}, \theta$. This is because $\mathrm{X}$ and $\mathrm{Z}$ are dependent through the path $\mathrm{X} \rightarrow \mathrm{V}_3 \rightarrow \mathrm{Z}$, and conditioning on $\mathrm{V}_3$ to inactivate this path activates the path $\mathrm{X} \rightarrow \mathrm{V}_2 \leftarrow \mathrm{U} \rightarrow \mathrm{Z}$, which cannot be inactivated given that $\mathrm{U}$ is hidden. Alternatively, the path $\mathrm{X} \rightarrow \mathrm{V}_3 \rightarrow \mathrm{Z}$ can also be eliminated by intervening on $\mathrm{V}_3$, instead of conditioning. In the system resulting from the intervention $do(\mathrm{V}_3)$, the sufficient statistic can be identified from the independence $\mathrm{X} \perp \mathrm{Z} | \theta, do(\mathrm{V}_3)$ (Figure \ref{f4}B), and this can be exploited by the IB algorithm because the intervened system is identifiable, that is, the joint distribution of all other variables when intervening $\mathrm{V}_3$ can be calculated from the observational distribution. In more detail, the joint distribution of the observable variables in the intervened system can be identified as \citep{Pearl09}
\small
\begin{equation}
\begin{split}
  %p&(\mathrm{X}, \mathrm{V}_1, \mathrm{V}_2, \mathrm{Z}| do(\mathrm{V}_3 = \mathrm{v}_3)) =  \\&\sum_\mathrm{U} p(\mathrm{Z}| \mathrm{X}, \mathrm{V}_1, \mathrm{V}_3=\mathrm{v}_3, \mathrm{U}) p(\mathrm{V}_2| \mathrm{X}, \mathrm{U}) p(\mathrm{X}) p(\mathrm{U}) p(\mathrm{V}_1) = \\&\sum_\mathrm{U} p(\mathrm{Z}| \mathrm{X}, \mathrm{V}_1, \mathrm{V}_3=\mathrm{v}_3, \mathrm{U}) p(\mathrm{U} | \mathrm{X}, \mathrm{V}_2) p(\mathrm{X}, \mathrm{V}_2) p(\mathrm{V}_1) \ \ =
%  \\&\ p(\mathrm{Z}| \mathrm{X}, \mathrm{V}_1, \mathrm{V}_2, \mathrm{V}_3 = \mathrm{v}_3) p(\mathrm{X}, \mathrm{V}_2) p(\mathrm{V}_1),
&p(\mathrm{X}, \mathrm{V}_1, \mathrm{V}_2, \mathrm{Z}| do(\mathrm{V}_3 = \mathrm{v}_3)) =  \sum_\mathrm{U} p(\mathrm{Z}| \mathrm{X}, \mathrm{V}_1, \mathrm{V}_3=\mathrm{v}_3, \mathrm{U}) p(\mathrm{V}_2| \mathrm{X}, \mathrm{U}) p(\mathrm{X}) p(\mathrm{U}) p(\mathrm{V}_1) = \\&\sum_\mathrm{U} p(\mathrm{Z}| \mathrm{X}, \mathrm{V}_1, \mathrm{V}_3=\mathrm{v}_3, \mathrm{U}) p(\mathrm{U} | \mathrm{X}, \mathrm{V}_2) p(\mathrm{X}, \mathrm{V}_2) p(\mathrm{V}_1) = p(\mathrm{Z}| \mathrm{X}, \mathrm{V}_1, \mathrm{V}_2, \mathrm{V}_3 = \mathrm{v}_3) p(\mathrm{X}, \mathrm{V}_2) p(\mathrm{V}_1),
  \label{e5}
\end{split}
\end{equation}
\normalsize
where the marginalization of the hidden variable $\mathrm{U}$ is possible because $\mathrm{Z} \perp \mathrm{V}_2 | \mathrm{X}, \mathrm{V}_1, \mathrm{V}_3, \mathrm{U}$ and $\mathrm{U} \perp \mathrm{V}_1 \mathrm{V}_3 | \mathrm{X}, \mathrm{V}_2$. There is no causal effect on $\mathrm{X}$, $\mathrm{V}_1$, and $\mathrm{V}_2$, so that $p(\mathrm{X}, \mathrm{V}_1, \mathrm{V}_2| do(\mathrm{V}_3 = \mathrm{v}_3 )$ is equal to $p(\mathrm{X}, \mathrm{V}_1, \mathrm{V}_2)$. The IB method can be applied to infer $\theta$, with the only difference that now the joint distribution of the intervened system in Eq.\,\ref{e5} is used as input.

For a concrete example, $\mathrm{Z}$ was generated with a binomial GLM with $p_\mathrm{z} = 1/(1+\mathrm{exp}(-h(\mathbf{Pa}_z)))$, where $\mathbf{Pa}_z = \{ \mathrm{X}, \mathrm{V}_1, \mathrm{V}_3, \mathrm{U}\}$. The function $h(\mathbf{Pa}_z)$ was defined as:
\begin{equation}
\begin{split}
h &=  a_0+ a_1 \mathrm{U} + a_2(\mathrm{X}+\mathrm{V}_1) + a_3 (\mathrm{X}+\mathrm{V}_1)^2 + a_4(\mathrm{X}+\mathrm{V}_1)\mathrm{U} + a_5(\mathrm{X}+\mathrm{V}_1)^2 \mathrm{U}.
\label{e4}
\end{split}
\end{equation}
Because we model the intervened system with $do(\mathrm{V}_3 = \mathrm{v}_3)$ fixed, the dependence of $\mathrm{Z}$ on $\mathrm{V}_3$ is absorbed in the value of the coefficients. $\mathrm{X}$, $\mathrm{V}_1$, and $\mathrm{U}$ are independent binary variables. The links $\mathrm{X} \rightarrow \mathrm{V}_2 \leftarrow \mathrm{U}$ were modeled generating $\mathrm{V}_2$ from a binomial distribution with $n_{v_2} = 16$ and a mean parameter $p(\mathrm{V}_2 = 1| \mathrm{X}, \mathrm{U}) = 0.1 + 0.3 \mathrm{x} + 0.3 \mathrm{u}$. Variable $\mathrm{V}_3$ was generated from a binomial distribution with $n_{v_3} =1$ and parameter $p(\mathrm{V}_3 =1| \mathrm{X}, \mathrm{V}_2) = 1/(1+\mathrm{exp}(2 -1.5 \mathrm{v}_2/n_{v_2} - 2.5 \mathrm{x})$. We used the intervention $do(\mathrm{V}_3 = 1)$ to identify the sufficient statistic. We followed the same procedure of Section \ref{ss42} to generate $1440$ configurations. The signal-to-noise ratio was controlled with $n_z \in \{4, 8, 16, 64\}$. The probability $p(\mathrm{U} = 1)= 0.5$ was kept constant across configurations. To ensure faithfulness of the distribution to the corresponding graph $G$ when randomly sampling the coefficients $\mathbf{a}$, we required $I(\mathrm{X}, \mathrm{V}_1; \mathrm{Z}| do(\mathrm{V}_3=1))/ H(\mathrm{Z}|do(\mathrm{V}_3=1))\geq 0.05$ and $I(\mathrm{V}_2; \mathrm{Z}| do(\mathrm{V}_3=1))/ H(\mathrm{Z}|do(\mathrm{V}_3=1))\geq 0.05$. As before, we also required a minimum difference of $0.05$ for $p_\mathrm{z}$ according to Eq.\,\ref{e4} for different values of $\theta = \mathrm{X}+\mathrm{V}_1$.

Figure \ref{f6}B shows the overall TP and FP rates. % Like for the case of selection bias, we leave a detailed analysis of the factors affecting the output of the IB algorithm for Appendix E.4.
High performance is achieved with sufficiently high $N$ values, with a high TP and low FP rate. In this system, $\theta$ can only be identified with $\mathbf{X} = \{ \mathrm{X}, \mathrm{V}_1\}$, while the input $\mathbf{X} = \{ \mathrm{X}, \mathrm{V}_1, \mathrm{V}_2\}$ only can produce false positives, since the path $\mathrm{X} \rightarrow \mathrm{V}_2 \leftarrow \mathrm{U} \rightarrow \mathrm{Z}$ leads to $\mathrm{Z} \notperp \mathrm{X}| \theta, \mathrm{V}_2, do(\mathrm{V}_3)$. The low FP rates reflect that the IBSSI method detects this dependence when $\mathbf{X} = \{ \mathrm{X}, \mathrm{V}_1, \mathrm{V}_2\}$ is used, and the selection criteria of algorithm \ref{alg1}, in particular line 18, precludes from selecting an invalid statistic. In this case we found a weak dependence of the TP rates on the amount of information, but the FP rates were also lower for higher information levels.

\section{Conclusions}

We introduced a general framework that uses sufficient statistics to increase the inferential power of previous methods for structure learning that extract causal information from conditional independencies. While the standard formulation of these methods exploits conditional independencies directly verifiable from the probability distribution of the observable variables \citep{Spirtes00,Pearl09}, we argue that the structure in the generative mechanisms of observable variables often contains substructures acting as sufficient statistics, which create additional conditional independencies. We proposed to use the Information Bottleneck (IB) method \citep{Tishby99} to identify the sufficient statistics and introduced the Information Bottleneck Sufficient Statistics Inference (IBSSI) method to select sufficient sets of statistics useful for structure learning. We extended the standard rules of causal discovery from observational data to exploit the independencies associated with the inferred sufficient statistics and applied this approach to data, illustrating that the IBSSI method is able to identify sufficient statistics with a high true positive (TP) rate and low false negative (FN) rate. We validated the IBSSI method in simulated systems specifically designed to contain different types of functional equations and types of sufficient statistics. We equally validated the method with a benchmark model \citep{Jenkins08, Guyon08} composed of Boolean regulatory rules \--in whose structure we identified the presence of sufficient statistics\-- that has previously been shown to accurately model a biological signal transduction network \citep{Li06}. We characterized critical factors that determine the tradeoff between the TP and FN rates and introduced conditions to assess the consistency of the statistics.

To exploit the sufficient statistics, in the main text for simplicity we focused on the two basic rules of causal orientation that allow inferring colliders and noncolliders from conditional independencies \citep{Verma93,Spirtes00}, since these rules underpin the complete set of rules that determines Markov equivalence classes of causal structures \citep{Zhang08c}. More broadly, the sufficient statistics can equally be applied to extend the complete set of rules, as we detail in Appendix C presenting an extension of the Causal Inference (CI) algorithm of \cite{Spirtes00} which integrates those and further new rules based on sufficient statistics together with the standard rules of causal orientation. Importantly, the new rules can interact synergistically with the standard rules, such that some information extracted from them iteratively allows the application of standard orientation rules not applicable otherwise. However, while the new rules can readily be integrated with the standard ones and the IBSSI method provides a procedure to infer sufficient sets of statistics, future work should further examine how to optimally combine the identification of sufficient statistics and their exploitation for structure learning. In particular, when probing the existence if sufficient statistics, especially in large systems, in a first step the partially oriented causal structure retrieved as the output of a standard algorithm such as the FCI \citep{Spirtes00} can help to determine which variables should be selected as potential arguments of the statistics. In subsequent iterative steps, new causal information extracted thanks to already identified statistics can inform the selection of new potential arguments of further statistics.

In the main text, we focused on the identification of sufficient statistics nonparametrically with the IB method. Nonparametric approaches have the advantage of generality, but for specific domains in which the form of the generative mechanisms can properly be modeled, modeling approaches can be less data demanding and computationally intensive. To correctly infer a sufficient set of statistics, a model does not need to properly capture the full functional equation, but only to identify the (possibly much simpler) subcomponents containing the functional sufficient statistics. We discuss in Appendix F how modeling approaches can equally be applied to identify statistics. Indeed, the form of a fitted model may already indicate the presence of sufficient statistics, which can then be tested subsequently. Furthermore, since the identification of sufficient sets of statistics relies on characterizing substructures embodied in the functional equations, we expect this method to be particularly adaptable to an hybrid approach \citep{Ogarrio16, Jabbari17}, such that instead of recovering a single partially oriented graph, multiple causal structures are scored, quantifying the confidence in different inferred sufficient statistics. Similarly, sufficient statistics embodied in the functional equations are expected to constitute simpler and possibly more robust submodules preserved by a causal mechanism across domains, as opposed to the whole set of parameters of the functional equations. Accordingly, the sufficient statistics could also be used by complementary techniques for structure learning which exploit the invariance of generative models across domains \citep{Peters16,Ghassami17,Besserve18,Heinze18}. Furthermore, as we illustrated in Section \ref{ss6}, the identification of sufficient statistics can easily be combined with methods to calculate the effect of external interventions in the system \citep{Shpitser08c, Shpitser08}. Using information from new independencies associated with sufficient statistics can equally be useful to select optimal interventions designed to discriminate between Markov-equivalent causal structures \citep{Hauser14,Triantallou15,Kocaoglu17,Ghassami18, Agrawal2019}, or to determine interventional Markov equivalence classes \citep{Hauser12}.

In this work we implemented the IB method with the original iterative procedure introduced by \cite{Tishby99}. However, more refined implementations of the IB method using deep networks \citep{Alemi17,Belghazi2018,Wieczorek18} promise to provide more efficient procedures to estimate sufficient sets of statistics also from high-dimensional data and for continuous variables. Deep learning approaches to infer generative models \citep{Kocaoglu18,Goudet18} can also be useful to characterize the subfunctions corresponding to sufficient statistics for high-dimensional data. A major challenge for the algorithms of structure learning based on conditional independencies is that of scalability to large systems \citep{Kalisch07,Raghu18,Ramsey19}. Large systems also represent a challenge for the application of the IBSSI method, in particular regarding the selection of which variables should be used as potential arguments of unknown underlying sufficient statistics. However, because the identification of sufficient statistics is especially conceived to exploit substructures within the functional equations that may depend on a substantially lower number of variables, we can expect that the method is particularly useful in large systems, for which conditional independencies between the observable variables involve a large number of conditioning variables. Such large complex causal structures comprise gene regulatory networks \citep{Maathuis10,Neto10,Banf17,Glymour19} or brain connectivity networks \citep{Chicharro14,Sanchez19,Reid19}. The identification of causally relevant low-dimensional sufficient statistics is also a main objective in systems neuroscience \citep{Cunningham14}, since understanding the neural code requires characterizing how information in the representation of sensory stimuli in neural population responses is transmitted across brain areas and transformed into a representation of behavioral decisions \citep{Chicharro14b,Panzeri17, Piasini17}. The IBSSI method promises to be particularly useful for this type of highly interconnected large systems with ubiquitous hidden variables, for which conditional independencies between the observable variables may be scarce.

% Acknowledgements should go at the end, before appendices and references
\acks{This work was supported by the BRAIN Initiative (Grants No. R01 NS108410, R01 NS109961 and No. U19 NS107464 to S.P.) and by the Fondation Bertarelli.}

\newpage

\appendix
\section{Faithfulness and stability for systems with structural sufficient statistics}
\label{a1}

In this section we extend to causal structures with sufficient statistics the notions of faithfulness \citep{Spirtes00} and stability \citep{Pearl09} which underpin the use of rules based on conditional independencies for structure learning. We first review their standard formulation. As discussed in the main article, faithfulness \citep[Ch.\,2 in][]{Spirtes00} ensures the mapping between conditional independencies present in a distribution and d-separations present in a graph:

\vspace*{2mm}
\noindent \textbf{Definition S1}\ \textbf{Faithfulness between a causal graph and a probability distribution}: \emph{A probability distribution $p(\mathbf{V})$ on $\mathbf{V}$ variables and a directed acyclic graph $G$ on those variables are faithful to one another if and only if $(\mathbf{X} \perp \mathbf{Y} |\mathbf{Z})_P \Leftrightarrow (\mathbf{X} \perp \mathbf{Y} |\mathbf{Z})_G$ for any disjoint three sets of variables $\mathbf{X}$, $\mathbf{Y}$, and $\mathbf{Z}$.}

\vspace*{2mm}
Here $(\mathbf{X} \perp \mathbf{Y} |\mathbf{Z})_P$ refers to a conditional independence between the variables, while $(\mathbf{X} \perp \mathbf{Y} |\mathbf{Z})_G$ refers to the separability of their corresponding nodes. The concept of stability is tightly related to faithfulness, but relies on introducing a space of parameters characterizing the functional equations. In particular, a causal model \citep{Pearl09} is defined as

\vspace*{2mm}
\noindent \textbf{Definition S2}\ \textbf{Causal model}: \emph{A causal model is a pair $M = \langle G, \Phi_G \rangle$ consisting of a causal structure $G$ and a set of parameters or functions $\Phi_G$ compatible with $G$. $\Phi_G$ defines a function $\mathrm{V}_i := f_i(\mathbf{Pa}_{\mathrm{V}_i}, \mathrm{U}_i)$ for each $\mathrm{V}_i \in \mathbf{V}$, where $\mathbf{V}$ is the set of variables associated with the nodes in the causal structure, $\mathbf{Pa}_{\mathrm{V}_i}$ are the parents of $\mathrm{V}_i$ in $G$, and $\mathrm{U}_i$ are exogenous random noises independent for each $\mathrm{V}_i$. $\Phi_G$ also assigns a probability measure $p(\mathrm{U}_i)$ to each $\mathrm{U}_i$.}

\vspace*{2mm}
We indicate that $\Phi_G$ comprises both parameters and functions because in general the functional equations $f_i$ can only be characterized by parameters if previously defining a basis of functions. Because a causal model determines both the distribution of the noise variables and the functional equations, it specifies completely the joint distribution $p(\mathrm{V})$ of the variables in the graph $G$. The stability of a causal model is defined as \citep{Pearl09}

\vspace*{2mm}
\noindent \textbf{Definition S3}\ \textbf{Stability of a causal model}: \emph{Let $I(p(\mathbf{V}))$ denote the set of all conditional independencies embodied in $p(\mathbf{V})$. A causal model $M = \langle G, \Phi_G \rangle$ generates a stable distribution $p(\mathbf{V};\langle G, \Phi_G \rangle)$ if and only if $I(p(\mathbf{V};\langle G, \Phi_G \rangle)) \subseteq I(p(\mathbf{V};\langle G, \Phi_G' \rangle))$ for any set $\Phi_G'$.}

\vspace*{2mm}
That is, a causal model is stable if no conditional independencies depend on the specific selection of the parameters or functions defining the functional equations. In that case, because the conditional independencies can only be determined by the causal structure, the assumption of stability for the causal model ensures faithfulness between the distribution resulting from the model and its causal graph. In particular,
$(\mathbf{X} \perp \mathbf{Y} |\mathbf{Z})_G \Rightarrow (\mathbf{X} \perp \mathbf{Y} |\mathbf{Z})_P$ is guaranteed by construction of the causal model, since definition S2 requires that $\Phi_G$ is compatible with $G$. On the other hand, $(\mathbf{X} \perp \mathbf{Y} |\mathbf{Z})_P \Rightarrow (\mathbf{X} \perp \mathbf{Y} |\mathbf{Z})_G$ is ensured by $I(p(\mathbf{V};\langle G, \Phi_G \rangle)) \subseteq I(p(\mathbf{V};\langle G, \Phi_G' \rangle))$ for any set $\Phi_G'$. For example, in a linear model stability refrains from a variable $\mathrm{V}_i$ being independent of one of its parents because the corresponding linear coefficient is zero, given that changing the coefficient would eliminate the independence. Conversely, the fact that a certain variable is not a parent of $\mathrm{V}_i$, sets its coefficient always to zero for any $\Phi_G'$ compatible with $G$.

We now extend these definitions considering instead of the graphs associated only with the variables in the system, the augmented graphs that also represent sufficient statistics present in the functional equations.

\vspace*{1.8mm}
\noindent \textbf{Definition S4}\ \textbf{Causal model with sufficient statistics}: \emph{A causal model with sufficient statistics is a triplet $M = \langle G , G^+_{\theta}, \Phi_{G, G^+_{\theta}} \rangle$ consisting of a causal structure $G$ for a set of variables $\mathbf{V}$, an augmented causal structure $G^+_{\theta}$ that also represents the set $\Theta$ of existing sufficient statistics in the functional equations of $\mathbf{V}$, and a set of parameters or functions $\Phi_{G, G^+_{\theta}}$ compatible with $G$ and $G^+_{\theta}$. For each $\theta_i \in \Theta$, $\Phi_{G, G^+_{\theta}}$ defines a deterministic function $\theta_i = g_i(\mathbf{Pa}^{G^+_{\theta}}_{\theta_i})$, where $\mathbf{Pa}^{G^+_{\theta}}_{\theta_i} \subset \mathbf{V}$ are the parents of $\theta_i$ in $G^+_{\theta}$ and correspond to the arguments of the statistic. For the variables $\mathrm{V}_i \in \mathbf{V}$, $\Phi_{G, G^+_{\theta}}$ defines a function $\mathrm{V}_i := f_i(\mathbf{Pa}^{G^+_{\theta}}_{\mathrm{V}_i}, \mathrm{U}_i)$, where $\mathbf{Pa}^{G^+_{\theta}}_{\mathrm{V}_i} \subset \{ \mathbf{V}, \Theta \}$ are the parents of $\mathrm{V}_i$ in $G^+_{\theta}$, and $\mathrm{U}_i$ are exogenous random noises independent for each $\mathrm{V}_i$. $\Phi_{G, G^+_{\theta}}$ also assigns a probability measure $p(\mathrm{U}_i)$ to each $\mathrm{U}_i$. The parenthood structure of $G$ and $G^+_{\theta}$ is consistent such that, for $\Theta_{\mathrm{V}_i} = \mathbf{Pa}^{G^+_{\theta}}_{\mathrm{V}_i} \cap \Theta$, and $\mathbf{Pa}_{\Theta_{\mathrm{V}_i}} = \bigcup_{\theta \in \Theta_{\mathrm{V}_i}} \mathbf{Pa}^{G^+_{\theta}}_{\theta}$, the parents of $\mathrm{V}_i$ in $G$ correspond to $\mathbf{Pa}^G_{\mathrm{V}_i} = \left ( \mathbf{Pa}^{G^+_{\theta}}_{\mathrm{V}_i} \setminus \Theta_{\mathrm{V}_i} \right ) \cup \mathbf{Pa}_{\Theta_{\mathrm{V}_i}} $.}

\vspace*{1.8mm}
Note that $\mathbf{Pa}^G_{\mathrm{V}_i} \subset \mathbf{V}$, consistently with the fact that $G$ does not contain sufficient statistics. The relation between $G$ and $G^+_{\theta}$ as characterized in Definition S4 formalizes the procedure to construct the augmented graph $G^+_{\theta}$, as discussed in Section \ref{ss2}. In particular, any parent of $\mathrm{V}_i$ that only appears in the functional equation $f_i$ of $\mathrm{V}_i$ as an argument of sufficient statistics, loses its parenthood status for $\mathrm{V}_i$ in $G^+_{\theta}$, so that the parents of $\mathrm{V}_i$ in $G$ correspond to its parents in $G^+_{\theta}$ that are not sufficient statistics together with the parents of the sufficient statistics embedded in its functional equation. This correspondence could be extended straightforwardly if further considering that some arguments of the statistics could themselves be other statistics. The set of parameters or functions $\Phi_{G, G^+_{\theta}}$ has to be compatible not only with $G$ but with $G^+_{\theta}$. Causal models with trivial sufficient statistics, such as $\theta = \mathrm{X}$, are excluded. For example, if there is a sufficient statistic $\theta = g(\mathrm{X})$ in the functional equation of $\mathrm{Z}$, with $\mathrm{X}$ as a single argument, then $g(\mathrm{X})$ has to be some noninvertible subfunction within the functional equation $f_\mathrm{z}$. This means that it may correspond to $\mathrm{X}^2$, $\cos(\mathrm{X})$, or $\max{(0, \mathrm{X})}$, but not to $\mathrm{X}^3$. Stability is formulated equivalently for a causal model with sufficient statistics, taking into account that now the changes between $\Phi_{G, G^+_{\theta}}$ and any $\Phi_{G, G^+_{\theta}}'$ are constrained by the compatibility with $G^+_{\theta}$:

\vspace*{2mm}
\noindent \textbf{Definition S5}\ \textbf{Stability of a causal model with sufficient statistics}: \emph{A causal model with sufficient statistics $M = \langle G, G^+_{\theta}, \Phi_{G, G^+_{\theta}} \rangle$ generates a stable probability distribution $p(\mathbf{V};\langle G, G^+_{\theta}, \Phi_{G, G^+_{\theta}} \rangle)$ if and only if $I(p(\mathbf{V};\langle G, G^+_{\theta}, \Phi_{G, G^+_{\theta}} \rangle )) \subseteq I(p(\mathbf{V};\langle G, G^+_{\theta}, \Phi'_{G, G^+_{\theta}} \rangle))$ for any set $\Phi'_{G, G^+_{\theta}}$.}

\vspace*{2mm}
In the same way that the stability of a causal model is based on the comparison of $\Phi_G$ only with other $\Phi_G'$ compatible with the corresponding graph $G$, the stability of a causal model with sufficient statistics relies on the comparison of $\Phi_{G, G^+_{\theta}}$ only with other sets $\Phi'_{G, G^+_{\theta}}$ compatible with $G^+_{\theta}$. This means that, in the same way that for a linear system a null linear coefficient for all non-parent variables is kept fixed for all $\Phi_G'$, similarly if for example the functional equation of $\mathrm{Z}$ depends only on $\mathrm{X}$ linearly through $a_{zx}\mathrm{X}^n$ with $n=2$, a non-null value of $a_{zx}$ can vary across $\Phi'_{G, G^+_{\theta}}$ unconstrained, but $n$ can only be an even exponent, to preserve the sufficient statistic. The restriction of comparisons to other configurations also compatible with the same graph $G^+_{\theta}$ determines the changes that can be introduced in the parameters or subfunctions determining the functional equations. The presence of sufficient statistics is considered as a constitutive constraint imposed by the laws governing a system, which cannot be modified. Faithfulness can equally be extended to relate probability distributions and the augmented graphs $G^+_{\theta}$:

\vspace*{2mm}
\noindent \textbf{Definition S6}\ \textbf{Faithfulness between a causal graph with sufficient statistics and a probability distribution}: \emph{A probability distribution $p(\mathbf{V})$ on $\mathbf{V}$ variables and a directed acyclic graph $G^+_{\theta}$ on those variables and on a set of statistics $\Theta$ are faithful to one another if and only if $\Theta$ can be deterministically determined from $\mathbf{V}$ and, given $p(\mathbf{V}, \Theta)$, $(\mathbf{X} \perp \mathbf{Y} |\mathbf{Z})_P \Leftrightarrow (\mathbf{X} \perp \mathbf{Y} |\mathbf{Z})_{G^+_{\theta}}$ for any disjoint three sets of variables $\mathbf{X}$, $\mathbf{Y}$, and $\mathbf{Z}$ such that no variables in
$\mathbf{X}$ or $\mathbf{Y}$ are deterministically determined by $\mathbf{Z}$.}

\vspace*{2mm}
Although $\Theta$ is not observed, $p(\Theta| \mathbf{V})$ is a deterministic mapping that allows constructing $p(\mathbf{V}, \Theta)$ from $p(\mathbf{V})$. Like for the standard definitions of stability of a causal model and faithfulness between a distribution and a causal graph, also in this case a stable causal model with sufficient statistics ensures that the resulting distribution and the graph $G^+_{\theta}$ are faithful to one another. Faithfulness between the distribution and the graph $G^+_{\theta}$ ensures that, for any candidate $p(\hat{\Theta}| \mathbf{V})$, a conditional independence is created in the distribution only if a consistent sufficient set of statistics exists in the graph. For example, consider the case of a system where $\mathrm{X}$ and $\mathrm{V}$ are parents of $\mathrm{Z}$, but there is no sufficient statistic that channels their influence on $\mathrm{Z}$ (Figure S1A). %(Figure \ref{f7}A).
If in this case a candidate sufficient statistic is created by a deterministic mapping $p(\hat{\theta} | \mathrm{X}, \mathrm{V})$, faithfulness ensures that $\mathrm{X},\mathrm{V}\notperp \mathrm{Z}| \hat{\theta}$ for any $\hat{\theta}$. Conversely, Figure S1B %\ref{f7}B
shows a case in which $\mathrm{X}$ and $\mathrm{V}$ are parents of $\mathrm{Z}$ in $G$, but only through a sufficient statistic $\theta$. In this case faithfulness ensures that when properly estimating the statistic ($\hat{\theta} = \theta$) a conditional independence $(\mathrm{X},\mathrm{V}\perp \mathrm{Z}| \hat{\theta})_{P}$ is created, consistent with the corresponding d-separation $(\mathrm{X},\mathrm{V}\perp \mathrm{Z}| \theta)_{G^+_{\theta}}$. The restriction to sets such that no variables in $\mathbf{X}$ and $\mathbf{Y}$ are deterministically determined by $\mathbf{Z}$ is discussed below.

Importantly, as in the standard case with no sufficient statistics, the assumption that a distribution and its corresponding causal structure are faithful to one another does not regard the practical issue of the estimation of the distribution $p(\mathbf{V})$ from data. Faithfulness does not preclude from obtaining false positives when the form of $p(\Theta| \mathbf{V})$ is determined from a finite sample size. It only ensures that, if a false positive is obtained, it is due to a poor estimation of the probability distribution, or due to a poor evaluation of the dependencies. Similarly, the fulfillment of faithfulness does not guarantee that an algorithm such as the IB algorithm will successfully infer $\Theta$, even when it exists. %Furthermore, the faithfulness assumption does not regard the exact functional form of the statistics embodied in $p(\mathbf{V})$.  when we quantified the performance of the IBSSI procedure inferring sufficient sets of statistics, we implemented a more demanding criterion to consider an inferred set of statistic as a true positive, namely that the form of the inferred statistics was consistent with the form of the known underlying statistics (Section \ref{ss4}). That is, we ensured that the IBSSI method correctly infers not only the presence of the statistics but their concrete form.

\begin{figure*}
  \begin{center}
    \scalebox{0.6}{\includegraphics*{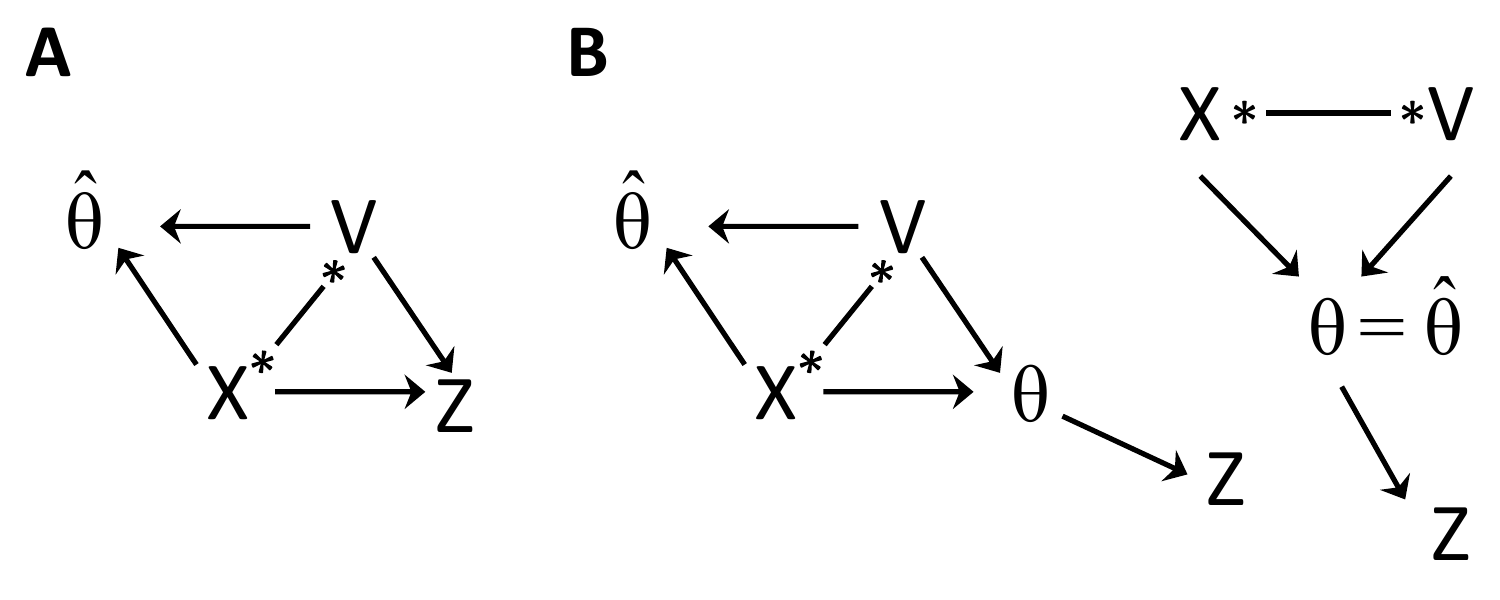}}
  \end{center}
  \captionsetup{labelformat=empty}
  \caption{Figure S1: Faithfulness of sufficient statistics. \textbf{A}) Augmented causal structure $G^+_{\theta}$ of a system in which $\mathrm{X}$ and $\mathrm{V}$ are the parents of $\mathrm{Z}$ without any sufficient statistic. The link $\mathrm{X} * \-- * \mathrm{V}$ indicates that any causal link is possible between $\mathrm{X}$ and $\mathrm{V}$. The graph is extended incorporating the parenthood structure of an estimated sufficient statistic $\hat{\theta}$, for which $\theta = g(\mathrm{X}, \mathrm{V})$ in the functional equation of $\mathrm{Z}$ is hypothesized. \textbf{B}) Augmented causal structure $G^+_{\theta}$ of a system in which $\mathrm{X}$ and $\mathrm{V}$ are the parents of $\mathrm{Z}$ only through the sufficient statistic $\theta$. Again the graph is extended incorporating the parenthood structure of the estimated sufficient statistic $\hat{\theta}$. The same causal structure is represented when the estimate correctly characterizes the sufficient statistic ($\hat{\theta} = \theta$). In A) $(\mathrm{X},\mathrm{V}\notperp \mathrm{Z}| \hat{\theta})_{G^+_{\theta}}$ and under the faithfulness assumption $(\mathrm{X},\mathrm{V}\notperp \mathrm{Z}| \hat{\theta})_P$, that is, no conditional independence is created by any estimated $\hat{\theta}$. In B) $(\mathrm{X},\mathrm{V}\perp \mathrm{Z}| \theta)_{G^+_{\theta}}$ and under the faithfulness assumption $(\mathrm{X},\mathrm{V}\perp \mathrm{Z}| \hat{\theta})_{P}$ when $\hat{\theta} = \theta$.}
  \label{f7}
\end{figure*}

Finally, despite the fact that the representation of a system containing sufficient statistics could be subsumed in definition S2 as a causal model in which some subset $\Theta$ of the variables are deterministically determined, we conceive these variables as qualitatively different, in the sense that they correspond to latent variables vicariously defined by the form of the functional equation of other variables. Deterministic relations limit the applicability of standard structure learning algorithms based on conditional independencies \citep{Spirtes00}, since they can create conditional independencies incompatible with the causal structure. In the presence of deterministic relations, the isomorphic mapping between conditional independencies and d-separability does not hold in general, which hinders the reconstruction of Markov equivalence classes of causal structures based on conditional independencies. In those cases, an extension of d-separability is required to relate graphical separability and independencies \citep{Geiger90,Spirtes00} and different refined algorithms of structure learning have been proposed to reconstruct equivalence classes \citep[e.g.\,][]{Lemeire2012,Mabrouk2014}. Following the assumptions of the standard algorithms \citep{Spirtes00}, the applicability of an extended algorithm incorporating the rules based on sufficient statistics is conceived for systems that do not contain deterministic relations between the variables associated with the causal structure $G$, so that the structural sufficient statistics are the only variables generated by deterministic relations. For these systems, the mapping between independence and d-separability holds for $\mathbf{X} \perp \mathbf{Y} |\mathbf{Z}$, when applied to sets of variables with positive conditional entropies $H(\mathrm{X}_i|\mathbf{Z})$, $H(\mathrm{Y}_i|\mathbf{Z})$, for all $\mathrm{X}_i \in \mathbf{X}$ and $\mathrm{Y}_i \in \mathbf{Y}$, as specified in definition S6. For example, for a system associated with Figure S1B %\ref{f7}B
the extended faithfulness assumption does not regard the relation between $(\theta \perp \mathrm{Z}| \mathrm{X}, \mathrm{V})_P$ and $(\theta \perp \mathrm{Z}| \mathrm{X}, \mathrm{V})_{G^+_{\theta}}$, since $H(\theta | \mathrm{X}, \mathrm{V})=0$, but this does not affect the new rules of causal orientation, since the conditions of independence examined only involve sufficient statistics appearing in the conditioning set.

\section{Sufficient sets of statistics with auxiliary statistics}
\label{a1a}

We here formalize the notion of a sufficient set of statistics that contains also auxiliary statistics, that is, statistics that do not have as an argument none of the two variables for which a new conditional independence is created:

\vspace*{1mm}
\noindent \textbf{Definition S7}\ \textbf{Sufficient set of statistics with auxiliary statistics for a pair of variables}: \emph{A sufficient set of statistics composed by a set of statistics $\Psi_{x,z}(\alpha)$ and an auxiliary set $\Lambda_{\mathbf{S}'}(\beta)$ of $M$ statistics for the variables in the set $\mathbf{S}'$ nonoverlapping to $\mathrm{X}, \mathrm{Z}$ exists if there is a set $\bar{\mathbf{S}}$ nonoverlapping to $\{ \mathbf{S}', \mathrm{X}, \mathrm{Z}\}$ and there is a sufficient set of statistics $\bar{\Psi}_{x,z}(\bar{\alpha}, \mathbf{S}')= \{ \Psi_{x,z}(\alpha), \Lambda_{\mathbf{S}'}(\beta)\}$, with $\bar{\alpha} = \{\alpha, \beta\}$, $\mathrm{X}, \mathrm{Z} \notin \bigcup \alpha$, $\{ \mathbf{S}', \mathrm{X}, \mathrm{Z} \} \cap \bigcup \beta = \emptyset$, such that, for $\bar{\mathbf{S}}_{\bar{\Psi}_{x,z}} = \{ \bar{\Psi}_{x,z}(\bar{\alpha}, \mathbf{S}') , \bar{\mathbf{S}}\}$, $\mathrm{X} \perp \mathrm{Z} | \bar{\mathbf{S}}_{\bar{\Psi}_{x,z}}$. The set of $K$ statistics $\Psi_{x,z}(\alpha)$ has to be obtainable as $\Psi_{x,z}(\alpha) = \{\psi_1(\mathrm{X},\mathrm{Z}; \tilde{\mathbf{V}}_1) ,...,\psi_K(\mathrm{X},\mathrm{Z}; \tilde{\mathbf{V}}_K)\}$, with $\alpha = \{ \tilde{\mathbf{V}}_1,...,\tilde{\mathbf{V}}_K\}$ and a set of functions $\psi_i = g_i(\mathrm{W}_i, \tilde{\mathbf{V}}_i)$  $\forall \psi_i \in \Psi_{x,z}(\alpha)$, with $\forall i\ \mathrm{W}_i= \mathrm{Z} $ or $\forall i\ \mathrm{W}_i= \mathrm{X}$. The set $\Lambda_{\mathbf{S}'}(\beta)$ has to be obtainable as $\Lambda_{\mathbf{S}'}(\beta) = \{\lambda_{1}(\tilde{\mathbf{S}}'_1; \tilde{\mathbf{S}}^*_1 ),...,\lambda_{M}(\tilde{\mathbf{S}}'_M; \tilde{\mathbf{S}}^*_M)  \}$, with $\beta = \{\tilde{\mathbf{S}}^*_1,..., \tilde{\mathbf{S}}^*_M \}$ and a set of functions $\lambda_{i} = g_i(\tilde{\mathbf{S}}'_i, \tilde{\mathbf{S}}^*_i)\ \forall i$, where $\tilde{\mathbf{S}}'_i \subseteq \mathbf{S}'$.}

Note that $\mathbf{S}'$ can overlap with $\alpha$, that is, the variables for which the auxiliary statistics exist can be arguments of the statistics $\Psi_{x,z}(\alpha)$. The set $\beta$ contains all other arguments of the auxiliary statistics apart from $\mathbf{S}'$. A statistic $\lambda_{i}(\tilde{\mathbf{S}}'_i; \tilde{\mathbf{S}}^*_i )$ is called auxiliary because $\mathrm{X}$ and $\mathrm{Z}$ are not in its arguments. Under the faithfulness assumption, auxiliary statistics also correspond to underlying functional sufficient statistics, but for the variables in $\mathbf{S}'$, or for intermediate variables being both colliders and noncolliders in paths creating a dependence between $\mathrm{X}$ and $\mathrm{Z}$, as described below Definition 3. If the sufficient set of statistics contains auxiliary statistics, rules R.c-ss and R.nc-ss can equally be applied with $\bar{\Psi}_{x,z}(\bar{\alpha}, \mathbf{S}')$ and $\bar{\mathbf{S}}_{\bar{\Psi}_{x,z}}$ instead of $\Psi_{x,z}(\alpha)$ and $\mathbf{S}_{\Psi_{x,z}}$.

The same logic explained in Section \ref{ss32} holds for the selection of $\mathbf{X}$ in the general case of a sufficient set of statistics comprising auxiliary statistics. The input must at least be $\mathbf{X} = \{ \mathrm{X}, \mathbf{S}', \bar{\alpha}, \bar{\mathbf{S}} \}$. That is, the input must include the variable $\mathrm{X}$ for which the sufficient set of statistics has to be identified, the variables $\mathbf{S}'$ for which auxiliary statistics are also to be identified, $\bar{\alpha}$ which comprises all the other arguments of all statistics, and the required conditioning set $\bar{\mathbf{S}}$. If the algorithm performs correctly, the output for this $\mathbf{X}$ will be $\tilde{\mathbf{X}} = \bar{\mathbf{S}}_{\bar{\Psi}_{x,z}}$ or $\tilde{\mathbf{X}} = \bar{\mathbf{S}}_{\bar{\Psi}_{x,z}} \backslash \mathrm{Y}$, for $\mathrm{X} \perp \mathrm{Z}| \bar{\mathbf{S}}_{\bar{\Psi}_{x,z}}$ and $\mathrm{X} \perp \mathrm{Z}| \bar{\mathbf{S}}_{\bar{\Psi}_{x,z}} \backslash \mathrm{Y}$, respectively. Examples of systems with auxiliary statistics are analyzed in Section \ref{ss42} and Appendix E below.

\section{An augmented algorithm for causal orientation}
\label{a1c}

We here show how rules using sufficient statistics can be inserted within a standard algorithm such as the FCI \citep{Spirtes00} or IC$^*$ \citep{Pearl09} algorithms. For simplicity, we incorporate the new rules within a simplified version of the Causal Inference (CI) algorithm of \cite{Spirtes00}, not including rules that exploit long distance independencies, which would require introducing some additional concepts such as \emph{definite discriminating paths} \citep{Spirtes00}. Furthermore, as mentioned in the Discussion, an important point for future research is to determine the best strategy to iteratively select which input variables $\mathbf{X}$ to use for the IBSSI method in order to find a new sufficient set of statistics, given the causal knowledge available. This available knowledge may have been learned either from the output of a standard algorithm, or as a result of a previous iteration exploiting other sufficient sets of statistics already inferred. In this presentation of an augmented algorithm we do not address the question of how to iteratively infer new sufficient sets of statistics, and we also leave aside any estimation issues. We assume that the statistics can be inferred with the IBSSI method, in the same way that in the standard algorithms it is assumed that conditional independencies can be correctly estimated. The algorithm below hence focuses on how to exploit the sufficient sets of statistics, not on how to infer them. If the rules that involve sufficient statistics are ignored, this algorithm is equivalent to a simplified version of the CI algorithm, as mentioned above. For the standard rules we indicate explicitly how they rely on the rules R.c and R.nc, following the nomenclature introduced in Section \ref{ss1}. We use $\bullet \--$ to indicate that the presence of an arrow at the end of an edge is undetermined, and we use $* \--$ as a placeholder for either $\bullet \--$, $\--$, or $\leftarrow$.

\vspace*{2mm}
\noindent \textbf{Causal Inference Algorithm with Sufficient Statistics (CI-ss Algorithm)}

\vspace*{1.5mm}
\noindent \textbf{Input:} Sampled distribution $p(\mathbf{V})$

\vspace*{1.5mm}
\noindent \textbf{Output:} Partially oriented graph

\vspace*{1.5mm}
\noindent \textbf{A)} \emph{Determine adjacencies}:

\begin{enumerate}

\item For each pair of variables $\mathrm{X}$ and $\mathrm{Z}$, search for a conditioning set $\mathbf{S}_{\mathrm{xz}} \subset \mathbf{V}$ nonoverlapping to $\{\mathrm{X}, \mathrm{Z}\}$ such that $\mathrm{X}$ and $\mathrm{Z}$ are independent conditioned on $\mathbf{S}_{\mathrm{xz}}$.

If there is no such $\mathbf{S}_{\mathrm{xz}}$, mark that $\mathrm{X}$ and $\mathrm{Z}$ are adjacent due to an unknown causal relation ($\mathrm{X} \bullet \-- \bullet \mathrm{Z}$); if there is such $\mathbf{S}_{\mathrm{xz}}$, record $\mathbf{S}_{\mathrm{xz}}$.

\end{enumerate}

\vspace*{1mm}
\noindent \textbf{B)} \emph{Orient arrows}:

\begin{enumerate}
 \item \emph{Determine colliders and noncolliders}:
    \begin{enumerate}
       \item From rules R.c and R.nc: For each pair of nonadjacent variables $\mathrm{X}$ and $\mathrm{Z}$ both adjacent to $\mathrm{Y}$ check if $ \mathrm{Y} \in \mathbf{S}_{\mathrm{xz}}$. If $ \mathrm{Y} \in \mathbf{S}_{\mathrm{xz}}$, following rule R.nc mark $\mathrm{X} *\-- * \mathrm{Y} * \-- * \mathrm{Z}$ as $\mathrm{X} *-\underline{*\mathrm{Y}*}-* \mathrm{Z}$, indicating that $\mathrm{Y}$ is a noncollider. If $ \mathrm{Y} \notin \mathbf{S}_{\mathrm{xz}}$, following rule R.c orient $\mathrm{X} *\-- * \mathrm{Y} * \-- * \mathrm{Z}$ as $\mathrm{X} *\rightarrow \mathrm{Y} \leftarrow* \mathrm{Z}$, indicating that $\mathrm{Y}$ is a collider.
       \item From rules R.c-ss and R.nc-ss: For each pair of adjacent variables $\mathrm{X}$ and $\mathrm{Z}$ both adjacent to $\mathrm{Y}$ search for a sufficient set of statistics $\Psi_{x,z}(\alpha)$ and a conditioning set $\mathbf{S}_{\mathrm{xz}}$ nonoverlapping with $\{ \mathrm{X}, \mathrm{Z}\}$ such that $\mathrm{X}$ and $\mathrm{Z}$ are independent conditioned on $\mathbf{S}_{\Psi_{x,z}} = \{\mathbf{S}_{\mathrm{xz}}, \Psi_{x,z}(\alpha)\}$. If $ \mathrm{Y} \in \mathbf{S}_{\mathrm{xz}}$ or $\mathrm{Y} \in \bigcup \alpha$, following R.nc-ss, given $\mathbf{\Psi}^{(y)}_{x,z}(\alpha)$ the subset of statistics in $\Psi_{x,z}(\alpha)$ which have $\mathrm{Y}$ as an argument, check if $\mathrm{X} \notperp \mathrm{Z} | \mathbf{S}_{\Psi_{x,z}} \backslash \{\mathrm{Y}, \mathbf{\Psi}^{(y)}_{x,z}(\alpha)\}$ and $\mathrm{X} \notperp \mathrm{Y} |\mathbf{S}'$,  $\mathrm{Y} \notperp \mathrm{Z} |\mathbf{S}'$, $\forall \mathbf{S}' \subseteq \mathbf{S}_{\Psi_{x,z}}  \backslash \{\mathrm{Y}, \mathbf{\Psi}^{(y)}_{x,z}(\alpha) \}$. If this is fulfilled, mark $\mathrm{X} *\-- * \mathrm{Y} * \-- * \mathrm{Z}$ as $\mathrm{X} *-\underline{*\mathrm{Y}*}-* \mathrm{Z}$, indicating that $\mathrm{Y}$ is a noncollider and record $\mathbf{S}_{\Psi_{x,z}}$. If $ \mathrm{Y} \notin \mathbf{S}_{\mathrm{xz}}$ and $\mathrm{Y} \notin \bigcup \alpha$, following R.c-ss, check if $\mathrm{X} \notperp \mathrm{Y} |\mathbf{S}_{\Psi_{x,z}}$ and $\mathrm{Y} \notperp \mathrm{Z} |\mathbf{S}_{\Psi_{x,z}}$. If this is fulfilled, orient $\mathrm{X} *\-- * \mathrm{Y} * \-- * \mathrm{Z}$ as $\mathrm{X} *\rightarrow \mathrm{Y} \leftarrow* \mathrm{Z}$, indicating that $\mathrm{Y}$ is a collider and record $\mathbf{S}_{\Psi_{x,z}}$.

    \end{enumerate}
  \item Repeat until no more edges can be oriented:
    \begin{enumerate}
        \item If $\mathrm{X}$ and $\mathrm{Z}$ are nonadjacent, $\mathrm{X} *\rightarrow \mathrm{Y} \leftarrow * \mathrm{Z}$, $\mathrm{W}$ is adjacent to $\mathrm{Y}$, and $\mathrm{W} \in \mathbf{S}_{\mathrm{xz}}$, orient $\mathrm{Y} *\-- * \mathrm{W}$ as $\mathrm{Y} \leftarrow * \mathrm{W}$.
        \item If $\mathrm{X}$ and $\mathrm{Z}$ are adjacent, $\mathrm{X} *\rightarrow \mathrm{Y} \leftarrow * \mathrm{Z}$, $\mathrm{W}$ is adjacent to $\mathrm{Y}$, and $\mathrm{W} \in \mathbf{S}_{\Psi_{x,z}}$ or $\Psi^{(w)}_{x,z}(\alpha)\neq \emptyset$, orient $\mathrm{Y} *\-- * \mathrm{W}$ as $\mathrm{Y} \leftarrow * \mathrm{W}$.
        \item If there is a directed path from $\mathrm{X}$ to $\mathrm{Y}$ and an edge $\mathrm{X} *\-- * \mathrm{Y}$, orient $\mathrm{X} *\-- * \mathrm{Y}$ as $\mathrm{X} * \rightarrow \mathrm{Y}$
        \item if $\mathrm{X} * \rightarrow \underline{\mathrm{Y}*}\-- * \mathrm{Z}$, then orient as $\mathrm{X} * \rightarrow \mathrm{Y} \rightarrow \mathrm{Z}$
    \end{enumerate}
\end{enumerate}

\vspace*{2mm}
Only steps \textbf{B}.1.(b) and \textbf{B}.2.(b) rely on sufficient statistics. Without these steps, the algorithm is equivalent to a simplified version of the Causal Inference (CI) Algorithm of section 6.7 in \cite{Spirtes00}. Step \textbf{A} corresponds to steps A-B) of that algorithm, step \textbf{B}.1.(a) is equivalent to their step C), and steps \textbf{B}.2.(a,c-d) correspond to a simplified version of their step D) which does not exploit long distance independencies.

Step \textbf{B}.1.(a) follows from the standard rules R.c and R.nc. Step \textbf{B}.1.(b) follows from R.c-ss and R.nc-ss. Step \textbf{B}.2.(a) exploits that conditioning on a child of a collider activates the paths through the collider in the same way as conditioning on the collider itself. Step \textbf{B}.2.(b) is the counterpart of step \textbf{B}.2.(a) using sufficient statistics. The steps \textbf{B}.2.(c-d) rely on combining pieces of causal information already previously inferred, and therefore do not have a separate counterpart based on sufficient statistics, they directly can be applied independently of whether sufficient statistics were used to infer that causal information. As mentioned in the Discussion, the addition of steps \textbf{B}.1.(b) and \textbf{B}.2.(b) can increase the specification of the causal structure not only because of their direct application, but also enabling the iterative application of steps \textbf{B}.2.(c-d), synergistically with the standard rules.

The newly introduced steps \textbf{B}.1.(b) and \textbf{B}.2.(b) do not exploit all the causal information that can be learned from inferring sufficient statistics. A straightforward way to further exploit them would be to extend the use of definite discriminating paths \citep{Spirtes00}, which are used in the part of step D) of the CI algorithm here simplified. Furthermore, steps \textbf{B}.1.(b) and \textbf{B}.2.(b) only exploit the new independencies created by the sufficient sets of statistics, but do not use the information about which variables appear as arguments of the statistics to orient edges in the graph. This is for the following reasons. First, given a pair $\mathrm{X}, \mathrm{Z}$, the IBSSI method, taking $\mathrm{Z}$ as the target variable and $\mathrm{X} \in \mathbf{X}$, retrieves a compressed representation $\hat{\theta}_{IB}$ from the input variables $\mathbf{X}$ which corresponds to $\mathbf{S}_{\Psi_{x,z}} = \{\mathbf{S}_{\mathrm{xz}}, \Psi_{x,z}(\alpha)\}$. That is, the representation $\hat{\theta}_{IB}$ does not distinguish between the statistics $\Psi_{x,z}(\alpha)$, and the conditioning set $\mathbf{S}_{\mathrm{xz}}$, it only provides us the necessary set to create a new independence. This means that the information about which are the arguments of the statistics is not immediately available. Second, not only the arguments are not explicitly identified, but it is not guaranteed that the sufficient statistics are embodied in the functional equation of $\mathrm{Z}$. This is because, as mentioned in Section \ref{ss2}, a new conditional independence can also result from sufficient statistics in the functional equation of intermediate variables adjacent to $\mathrm{X}$ and $\mathrm{Z}$ that are both colliders and noncolliders between them. The statistics can also be embodied in the functional equation of some variables along causal paths from $\mathrm{X}$ to $\mathrm{Z}$ that have not been included in $\mathbf{X}$. For these reasons, even assuming under the extended faithfulness assumption that the statistics found correspond to underlying functional statistics in the system, some extra procedure posterior to the identification of $\theta_{IB}$ would be required in order to exploit knowledge about the composition of the statistics. The development of this analysis is left for a subsequent contribution.

\section{Supplementary description of the characterization of the IBSSI method and its performance}
\label{a1b}

We here describe in more detail the faithfulness constraints imposed to select the sets of configurations for our simulated systems and the criteria used to bin the configurations into information levels. Subsequently, we will discuss additional criteria and adjustments of the IBSSI method to select sufficient statistics.

\subsection{Implementation of faithfulness constraints}

We expand the discussion of faithfulness constraints of Sections \ref{ss4}-\ref{ss6}. We have used two different types of constraints to generate simulated systems. The first type ensures that the distributions are consistent with the underlying sufficient statistics. For all systems we required that, for those conditional probabilities involved in the inference of the sufficient set of statistics, the selected set of parameters resulted in a minimum difference of $0.05$ between probability values that should be different according to the form of the functional equations. That is, we ensured that, for different instantiations of the conditioning set of variables, the resulting probability values were only equal because those different instantiations mapped to the same value of an existing structural sufficient set of statistics and not because the specific coefficients selected created the equality. For the systems following Eqs.\,\ref{e1} and \ref{e2}, since this type of constraint was imposed directly to $p(\mathrm{Z}| \mathbf{Pa}_z)$, we did not impose any further constraint to require a minimum magnitude of $I(\mathrm{Z}; \mathbf{Pa}_z)$, which would further guarantee the faithfulness between the distribution and the parenthood structure of $\mathrm{Z}$ in the graph. Indeed, given the average information levels reported in the caption of Figure \ref{f3}, we have tested the performance of the IBSSI method for cases with a remarkably low level of information between the arguments of a sufficient statistic and the variable in whose functional equation it is embedded ($\langle I'(\mathrm{X} \mathrm{V}_1; \mathrm{Z}) \rangle = 0.03$ for the low information level of Figure \ref{f3}A).

A second type of constraint was further imposed to the systems with selection bias (Section \ref{ss5}) and with intervened distributions (Section \ref{ss6}) to more generally guarantee the faithfulness between the distributions and the corresponding causal structures. For the case of the systems with selection bias, we imposed a lower bound of $I(\mathrm{Z}; \mathrm{X}, \mathrm{V}_1| \{\mathrm{S}_0\})/ H(\mathrm{Z}| \{\mathrm{S}_0\})\geq 0.05$ to ensure that indeed conditioning on $\{\mathrm{S}_0\}$ was introducing a dependence between $\mathrm{Z}$ and $\mathrm{X}, \mathrm{V}_1$. Similarly, in Section \ref{ss6}, we imposed a lower bound $I(\mathrm{Z};\mathrm{X}, \mathrm{V}_1| do(\mathrm{V}_3=1))/ H(\mathrm{Z}|do(\mathrm{V}_3=1))\geq 0.05$ and $I(\mathrm{Z}; \mathrm{V}_2| do(\mathrm{V}_3=1))/ H(\mathrm{Z}|do(\mathrm{V}_3=1))\geq 0.05$ to verify that the simulated configurations had a probability distribution faithful to the corresponding causal structure in the intervened system.

These constraints to discard unfaithful configurations can also be considered from a pragmatic perspective as a way to limit the difficulty of the examples studied. Indeed, although in theory any nonzero dependence associated with an edge in the graph is consistent with the faithfulness assumption, for any practical analysis that has to test that dependence its magnitude matters. Moreover, while these constraints were used \emph{a priori} when generating the sets of configurations for each system, we also implemented an \emph{a posteriori} procedure to check that the configurations selected did not create unfaithful sufficient sets of statistics, not corresponding to the structural statistics defined in the functional equations. For this purpose, for any $\hat{\theta}_{IB}$ that was accepted following the selection criteria of algorithm \ref{alg1} (lines 16-18), for any sample size $N$ and input values $\mathbf{X}$, $\beta'$, and $\max |\hat{\theta}_{IB}|$, we checked if it was consistent with a true underlying set of statistics. For any case in which it was not consistent, we reevaluated the selection criteria with a substantially larger sample size, to determine whether a false positive was caused by estimation issues or because of the presence of an unfaithful set of statistics. In particular, we simulated data from that configuration with a larger sample size of $N = 2 \cdot 10^6$ and used the previously determined projection $p(\hat{\theta}_{IB} | \mathbf{X})$ to reevaluate the selection criteria. If the selection criteria were still fulfilled, the configuration was considered as producing an unfaithful sufficient set of statistics. Following the same criterion used to determine false positives (Section \ref{ss42}), a configuration was discarded if for any $\beta'$ it led to the acceptance of an unfaithful sufficient set of statistics. This \emph{a posteriori} analysis supported the validity of the two types of constraints used to \emph{a priori} discard configurations, since we found that only a $1.6 \%$ of the configurations accepted by the \emph{a priori} constraints were \emph{a posteriori} assessed as unfaithful. These configurations were excluded when quantifying the true and false positive rates.

\subsection{Stratification of information levels}

As discussed in Section \ref{ss42}, information levels were constructed based on the normalized information about $\mathrm{Z}$ contained in the minimal input $\mathbf{X}$ resulting in a sufficient set of statistics, or on the largest $\mathbf{X}$, when no statistics exist. The bins were selected with two criteria. First, each bin had to include at least  $150$ configurations. Second, average information values associated with the levels should span well the range of values covered by all configurations. Because this range is specific for each type of system, the bins are also system-specific. We determined the levels by the two bounds $[a, b]$, of the medium level. The low level contains all configurations with information lower than $a$, and the high level contains those with information higher than $b$. For the systems from Eq.\,\ref{e1a}, the normalised information $I'(\mathrm{X} \mathrm{V}_1; \mathrm{Z})$ was used with the medium bin determined by $[0.05, 0.1]$. For the systems from Eqs.\,\ref{e1b} and \ref{e2b}, $I'(\mathrm{X} \mathrm{V}_1 \mathrm{V}_2; \mathrm{Z})$ was used, with $[0.15, 0.25]$. For the systems from Eq.\,\ref{e2a}, $I'(\mathrm{X} \mathrm{V}_1 \mathrm{Y}; \mathrm{Z})$ was used, with $[0.1, 0.2]$. For the systems of Eqs.\,\ref{e3} and \ref{e4}, $[0.1, 0.15]$ was used, with $I'(\mathrm{X} \mathrm{V}_1; \mathrm{Z}| \{\mathrm{S}_0\})$ and $I'(\mathrm{X} \mathrm{V}_1; \mathrm{Z}| do(\mathrm{V}_3=1))$, respectively. For the examples discussed in Appendix E below, we will provide the binning details when introducing the systems.

\subsection{Adjustments to the IBSSI method}

We here expand the discussion at the end of Section \ref{ss42} about factors that can be adjusted to set the tradeoff between sensitivity and specificity. As seen from comparing performance across information levels, using a lower bound on information to select inferred sets of statistics can help to decrease the false positives rate. %Furthermore, a finer analysis of which is the contribution to $I(\mathrm{Z}; \mathbf{X})$ of different events $\mathbf{X} = \mathbf{x}$ compressed into a single value of $\hat{\theta}_{IB}$ can help to assess whether $\hat{\theta}_{IB}$ is likely reflecting a true underlying set of statistics. In particular, the mutual information can be decomposed as
%\begin{equation}
%\label{sm0}
%\tag{S1}
%I(\mathrm{Z}; \mathbf{X}) = \sum_\mathbf{x} p(\mathbf{x}) KL(p(\mathrm{Z}|\mathbf{x}); p(\mathrm{Z})),
%\end{equation}
%that is, as the mean of the divergencies $KL(p(\mathrm{Z}|\mathbf{x}); p(\mathrm{Z}))$. The existence of a sufficient set of statistics should be reflected not only in $I(\mathrm{Z}; \mathbf{X})= I(\mathrm{Z}; \hat{\theta}_{IB})$, but in the preservation of each term $KL(p(\mathrm{Z}|\mathbf{x}); p(\mathrm{Z}))= KL(p(\mathrm{Z}|\hat{\theta}_{IB}(\mathbf{x})); p(\mathrm{Z}))$, where $\hat{\theta}_{IB}(\mathbf{x})$ is the value taken by $\hat{\theta}_{IB}$ given $\mathbf{X} = \mathbf{x}$. However, when enforcing a low value of $\max |\hat{\theta}_{IB}|$ as input to the IB algorithm, a compressed representation may be selected which merges the summands corresponding to events $\mathbf{X} = \mathbf{x}$ that have a low contribution to the information because their probability $p(\mathbf{x})$ is low. Accordingly, a way to further gain confidence in the validity of an accepted $\hat{\theta}_{IB}$ is to check that the projection $p(\hat{\theta}_{IB}|\mathbf{X})$ is compressing events of $\mathbf{X}$ with a substantial contribution to the information, and not events with simply a low probability $p(\mathbf{x})$.
%
%Beyond the examination of the information $I(\mathrm{Z}; \mathbf{X})$, we also discussed in Sections \ref{ss33} and \ref{ss42} how different components of algorithm \ref{alg1} can be adjusted to control the tradeoff between the TP and FP rates.
In algorithm 1, the thresholds used in the selection criteria (lines 16-18) as well as the conditions of lines 4 and 11 can be adjusted. In Appendix E.5 we show that the requirement in line 4 that already for $\max |\hat{\theta}_{IB}| = |\mathbf{X}|$ the output $\hat{\theta}_{IB}$ fulfills $|\hat{\theta}_{IB}|< |\mathbf{X}|$ may be quite demanding for low $N$. On the other hand, the requirement of consistency of line 11 can be extended to a larger range of $\max |\hat{\theta}_{IB}|$, or equally used to check the consistency of candidate statistics inferred with different $\beta'$ values. In particular, we adopted a conservative criterion when quantifying the FP rates, considering a false positive any configurations for which a false $\hat{\theta}_{IB}$ had been accepted for any $\beta'$. A more refined implementation could verify the consistency across those $\beta'$ for which $\hat{\theta}_{IB}$ had been accepted, and an error would only occur in the case that the false sufficient set of statistics was consistent across $\beta'$ values.

\section{Supplementary examples}
\label{a2}

\begin{figure*}
  \begin{center}
    \scalebox{0.5}{\includegraphics*{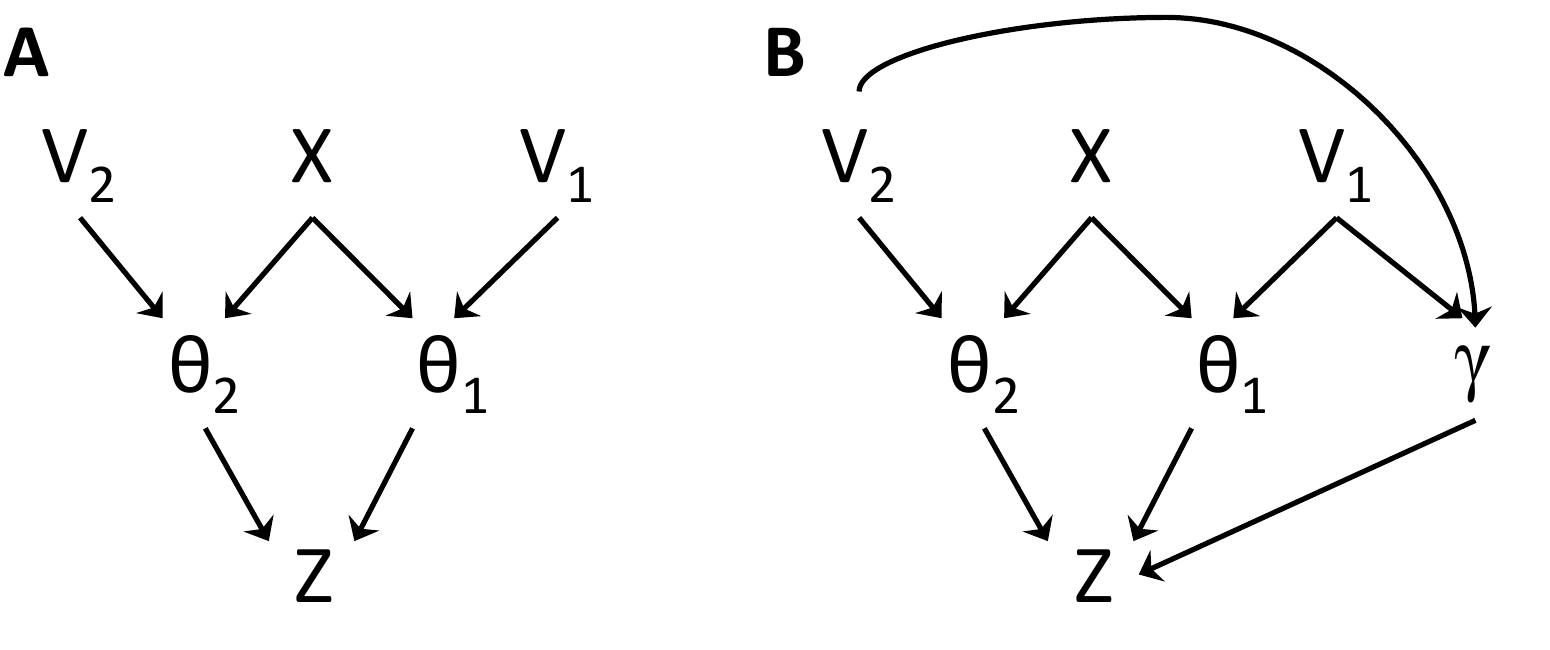}}
  \end{center}
  \captionsetup{labelformat=empty}
  \caption{Figure S2: Additional causal structures studied with the IBSSI method. \textbf{A}) Causal structure with a sufficient set composed by two statistics $\theta_1$ and $\theta_2$. \textbf{B}) Causal structure with two statistics $\theta_1$ and $\theta_2$ and an auxiliary statistic $\gamma$. Systems with the causal structures in A) and B) are described in Eqs.\,\ref{sm1} and \ref{sm2}, respectively.}
  \label{f4b}
\end{figure*}

We here study further examples of systems containing sufficient statistics. We examine examples with multiple sufficient statistics and examples of statistics with an alternative subfunctional form. We also consider systems with an alternative generative mechanism for $\mathrm{Z}$, instead of a binomial GLM. As in previous examples, we studied performance across $1440$ system configurations. In each case we simulated $K = 40$ systems, generating random values of the coefficients, and generated all combinations of $p(\mathrm{X}=1)$ and $p(\mathrm{V}_1=1)$ with values $\{ 0.3, 0.5, 0.7\}$. We also generated data for four different values of a parameter controlling the signal-to-noise ratio, which is specifically described below for each generative mechanisms.

\subsection{Multiple Sufficient Statistics}

We here consider a system with the structure of Figure S2A, %\ref{f4b}A,
containing two statistics. $\mathrm{Z}$ is again generated with a binomial GLM, with the conditional mean determined by $p_\mathrm{z} = 1/(1+\mathrm{exp}(-h(\mathbf{Pa}_z)))$. We selected $h(\mathbf{Pa}_z)$ to be
\begin{equation}
\label{sm1}
\tag{S1}
\begin{split}
h &=  a_0+ a_1(\mathrm{X}+\mathrm{V}_2) + a_2(\mathrm{X}+\mathrm{V}_2)^2 + a_3(\mathrm{X}+\mathrm{V}_1) + a_4(\mathrm{X}+\mathrm{V}_1)(\mathrm{X}+\mathrm{V}_2) \\
& + a_5(\mathrm{X}+\mathrm{V}_1)^2 + a_6(\mathrm{X}+\mathrm{V}_1)^2(\mathrm{X}+\mathrm{V}_2)^2,
\end{split}
\end{equation}
where $\theta_1 = \mathrm{X}+\mathrm{V}_1$ and $\theta_2 = \mathrm{X}+\mathrm{V}_2$. To ensure faithfulness to $G^+_{\theta}$ we discarded random instantiations of coefficients $\mathbf{a}$ if for two events of $\mathbf{Pa}_z$ that following Eq.\,\ref{sm1} should correspond to a different value of $p_z$, the difference in $p_z$ was smaller than $0.05$.

\begin{figure*}
  \begin{center}
    \scalebox{0.38}{\includegraphics*{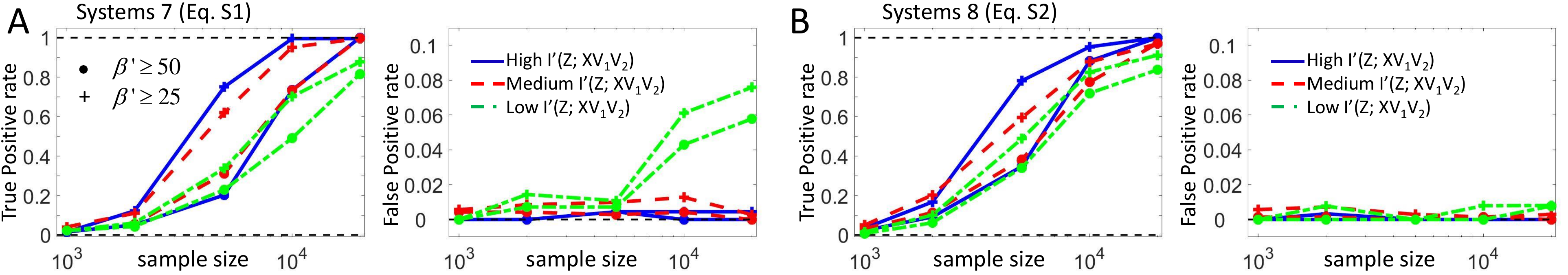}}
  \end{center}
  %\vspace{0.25in}
  \captionsetup{labelformat=empty}
  \caption{Figure S3: Performance of the IBSSI method for the systems of Figure S2. \textbf{A}) Results for systems generated following Eq.\,\ref{sm1}. $I'(\mathrm{X} \mathrm{V}_1 \mathrm{V}_2; \mathrm{Z}) \equiv I(\mathrm{X} \mathrm{V}_1 \mathrm{V}_2; \mathrm{Z})/ H(\mathrm{Z})$ is used for binning the information levels. The average information at each level is: Low: $\langle I'(\mathrm{X} \mathrm{V}_1 \mathrm{V}_2; \mathrm{Z}) \rangle = 0.12$. Medium: $\langle I'(\mathrm{X} \mathrm{V}_1 \mathrm{V}_2; \mathrm{Z}) \rangle = 0.20$. High: $\langle I'(\mathrm{X} \mathrm{V}_1 \mathrm{V}_2; \mathrm{Z}) \rangle = 0.28$. \textbf{B}) Results for systems generated following Eq.\,\ref{sm2}. Also $I'(\mathrm{X} \mathrm{V}_1 \mathrm{V}_2; \mathrm{Z})$ is used for binning. The average information at each level is: Low: $\langle I'(\mathrm{X} \mathrm{V}_1 \mathrm{V}_2; \mathrm{Z}) \rangle = 0.08$. Medium: $\langle I'(\mathrm{X} \mathrm{V}_1 \mathrm{V}_2; \mathrm{Z}) \rangle = 0.15$. High: $\langle I'(\mathrm{X} \mathrm{V}_1 \mathrm{V}_2; \mathrm{Z}) \rangle = 0.24$.}
  \label{f8}
\end{figure*}

Here $\theta_i \in \{0, 1, 2\}$, with $\theta_i =1$ for the events $\mathrm{X} =0,\mathrm{V}_i =1$ or $\mathrm{X} =1,\mathrm{V}_i =0$, both for $i= 1,2$. In both cases $I(\mathrm{X};\mathrm{Z}|\theta_i =1)>0$, because the two events differ in the value of $\mathrm{X}$, which affects $\mathrm{Z}$ through the other statistic. Accordingly, only $I(\mathrm{X};\mathrm{Z}|\theta_1, \theta_2)=0$. As discussed in the main text, the algorithm infers $\hat{\theta}_{IB}$ in the space of $\theta_1 \otimes \theta_2$. The underlying cardinality is $|\theta_{IB}|=7$, with only $\mathrm{X}=0, \mathrm{V}_1 =1, \mathrm{V}_2 =1$ and $\mathrm{X}=1, \mathrm{V}_1 =0, \mathrm{V}_2 =0$ leading to the same value of $p_\mathrm{z}$. The form of the functional equation is analogous to the one of Eq.\,\ref{e1b}, now with the second sufficient statistic $\theta_2$ instead of the auxiliary statistic $\gamma = \mathrm{V}_1 + \mathrm{V}_2$ present in Eq.\,\ref{e1b}. Also in Eq.\,\ref{e1b} the cardinality was $|\theta_{IB}|=7$, in that case with $\mathrm{X}=0, \mathrm{V}_1 =1, \mathrm{V}_2 =0$ and $\mathrm{X}=1, \mathrm{V}_1 =0, \mathrm{V}_2 =1$ leading to the same value of $p_\mathrm{z}$. The IB algorithm is able to detect the proper statistics without \emph{a priori} knowledge of how many statistics exist and whether they are statistics for $\mathrm{X}$ or auxiliary statistics. As seen in Figure S3A, the dependence of the TP and FP rates on the sample size, range of $\beta'$, and information levels, is analogous to the one from Eq.\,\ref{e1b} (second column in Figure \ref{f3}). Information levels were stratified based on $I'(\mathrm{X} \mathrm{V}_1 \mathrm{V}_2; \mathrm{Z})$ with $[0.15, 0.25]$ determining the bounds as described in Appendix D above.

\subsection{Sufficient Statistics with Other Forms}

We now examine systems with another form of the sufficient statistics. Again we consider $\mathrm{Z}$ generated from a binomial GLM, with the following form of $h(\mathbf{Pa}_z)$
\begin{equation}
\label{sm2}
\tag{S2}
h =  a_0+ a_1 \mathrm{X}\mathrm{V}_1 + a_2 \mathrm{X}\mathrm{V}_2 + a_3 \mathrm{V}_1 \mathrm{V}_2,
\end{equation}
where $\theta_1 = \mathrm{X} \mathrm{V}_1$, $\theta_2 = \mathrm{X} \mathrm{V}_2$, and $\gamma = \mathrm{V}_1 \mathrm{V}_2$ (Figure S2B). %(Figure \ref{f4b}B).
Again $\mathrm{X}$, $\mathrm{V}_1$, and $\mathrm{V}_2$ are binary with values $0$ or $1$. A set of $1440$ configurations was generated as described in Section \ref{ss42}, with the same faithfulness criterion. Conditioning only in $\theta_1$ and $\theta_2$ does not separate $\mathrm{X}$ and $\mathrm{Z}$, since conditioning on them creates a dependence between $\mathrm{X}$ and $\mathrm{V}_1\mathrm{V}_2$. The independence is only obtained with $\mathrm{X} \perp \mathrm{Z} | \theta_1, \theta_2, \gamma$. Each statistic $\theta_1$, $\theta_2$, and $\gamma$ takes values $0$ or $1$, but the space formed by $\theta_1 \otimes \theta_2 \otimes \gamma$ has cardinality $5$, with the same value for the four events of $\{ \mathrm{X}, \mathrm{V}_1, \mathrm{V}_2\}$ in which less than two of the variables have value $1$, and with a different value for the other four events in which at least two of the variables have value $1$. As seen in Figure S3B, also for this type of systems high TP rates and low FP rates are achieved. Information levels were stratified based on $I'(\mathrm{X} \mathrm{V}_1 \mathrm{V}_2; \mathrm{Z})$ with $[0.1, 0.2]$ determining the bounds as described in Appendix D above.

\begin{figure*}[t]
  \begin{center}
    \scalebox{0.38}{\includegraphics*{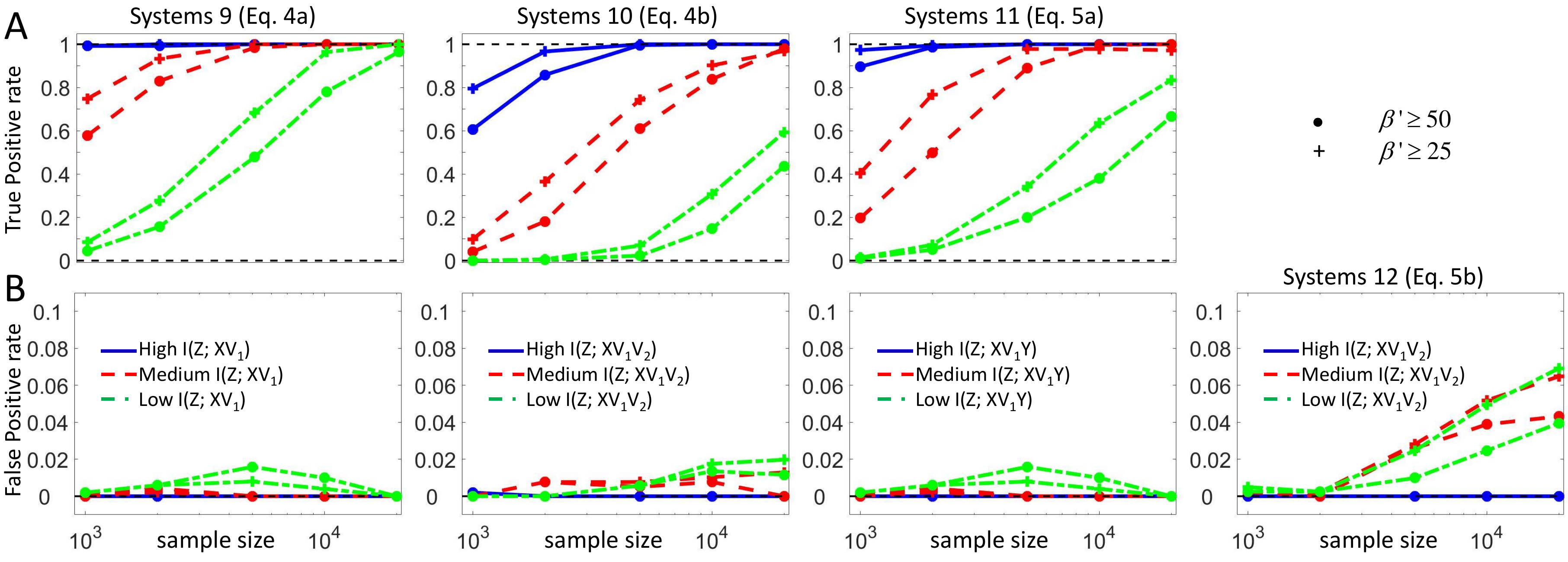}}
  \end{center}
  %\vspace{0.25in}
  \captionsetup{labelformat=empty}
  \caption{Figure S4: Performance of the IBSSI method for systems with the same causal structure of Figure \ref{f02}, the same form of the sufficient statistics as in Eqs.\,\ref{e1} and \ref{e2}, but with $\mathrm{Z}$ generated with the functional equation of Eq.\,\ref{sm3}, instead of with binomial GLMs as in previous examples. \textbf{A}-\textbf{D}) Results for systems generated following Eq.\,\ref{sm3}, with the causal structure of panels A-D in Figure \ref{f02}. The structure of the figure is analogous to Figure \ref{f3}. The average information at each level is: \textbf{A}) Low: $\langle I'(\mathrm{X} \mathrm{V}_1; \mathrm{Z}) \rangle = 0.1$. Medium: $\langle I'(\mathrm{X} \mathrm{V}_1; \mathrm{Z}) \rangle = 0.22$. High: $\langle I'(\mathrm{X} \mathrm{V}_1; \mathrm{Z}) \rangle = 0.36$. \textbf{B}) Low: $\langle I'(\mathrm{X} \mathrm{V}_1 \mathrm{V}_2; \mathrm{Z}) \rangle = 0.13$. Medium: $\langle I'(\mathrm{X} \mathrm{V}_1 \mathrm{V}_2; \mathrm{Z}) \rangle = 0.30$. High: $\langle I'(\mathrm{X} \mathrm{V}_1 \mathrm{V}_2; \mathrm{Z}) \rangle = 0.54$. \textbf{C}) Low: $\langle I'(\mathrm{X} \mathrm{V}_1 \mathrm{V}_2; \mathrm{Z}) \rangle = 0.11$. Medium: $\langle I'(\mathrm{X} \mathrm{V}_1 \mathrm{V}_2; \mathrm{Z}) \rangle = 0.30$. High: $\langle I'(\mathrm{X} \mathrm{V}_1 \mathrm{V}_2; \mathrm{Z}) \rangle = 0.53$. \textbf{D}) Low: $\langle I'(\mathrm{X} \mathrm{V}_1 \mathrm{V}_2; \mathrm{Z}) \rangle = 0.19$. Medium: $\langle I'(\mathrm{X} \mathrm{V}_1 \mathrm{V}_2; \mathrm{Z}) \rangle = 0.30$. High: $\langle I'(\mathrm{X} \mathrm{V}_1 \mathrm{V}_2; \mathrm{Z}) \rangle = 0.55$.}
  \label{f9}
\end{figure*}

\subsection{Other Generative Mechanisms}

We now examine systems consistent with the same causal structures of Figure \ref{f02} and containing the same subfunctional forms corresponding to the sufficient statistics of Eqs.\,\ref{e1} and \ref{e2}, but embedded in a different functional form for $\mathrm{Z}$. In Section \ref{ss42}, the generative mechanism of $\mathrm{Z}$ was a binomial GLM with $p_\mathrm{z} = 1/(1+\mathrm{exp}(-h(\mathbf{Pa}_z)))$. Here we consider systems with $f_z(\mathbf{Pa}_z)$ defined as
\begin{equation}
\label{sm3}
\tag{S3}
\mathrm{Z} = \lfloor \exp(h(\mathbf{Pa}_z)) + \sigma_\xi \xi_z \rceil,
\end{equation}
where $h(\mathbf{Pa}_z)$ takes the same form of the subfunctions in Eqs.\,\ref{e1} and \ref{e2} containing the sufficient statistics, and $\xi_z$ corresponds to a source of noise with a standard Gaussian distribution, with zero mean and unit variance. The signal-to-noise ratio is controlled by $\sigma_\xi$, analogously to how for the binomial GLMs it was controlled modifying the number of trials $n$. In this case we generated systems with $\sigma_\xi \in \{ 0.5, 1, 2, 3\}$. The symbol $\lfloor \cdot \rceil$ indicates a rounding operation. This operation may be part of the underlying generative mechanism of $\mathrm{Z}$, or may be a convenient discretization to apply the IB algorithm. It does not affect the existence of the sufficient statistics, which only depends on the form of the subfunction $h(\mathbf{Pa}_z)$. Eq.\,\ref{sm3} models the functional equation of $\mathrm{Z}$ and not its probabilistic mechanism, as it is the case when GLMs are used, and hence, in order to ensure that the generated configurations faithfully represented the embodied sufficient statistics, we required that $\exp(h(\mathbf{Pa}_z))$ \--which determines the conditional mean of $\mathrm{Z}$\-- had at least a difference of $0.5$ for events of $\mathbf{Pa}_z$ for which, according to Eqs.\,\ref{e1} and \ref{e2}, the conditional mean should differ.

Figure S4 shows the TP and FP rates as a function of the sample size for these systems. In comparison to Figure \ref{f3}, here the selection of the signal-to-noise ratio levels results in a wider range of information levels, which explains why the TP rate depends more on the information level than on the selection of the $\beta'$ set. Also in this case high TP rates are achieved while preserving low FP rates. The information levels were based on the same normalised information measures as in Section \ref{ss42}, but with the bounds adapted to these examples. For the systems from Eq.\,\ref{e1a}, the normalised information $I'(\mathrm{X} \mathrm{V}_1; \mathrm{Z})$ was used with the medium bin determined by $[0.15, 0.3]$. For the other three types of systems the bins were determined by $[0.2, 0.4]$, with $I'(\mathrm{X} \mathrm{V}_1 \mathrm{V}_2; \mathrm{Z})$ for the systems from Eqs.\,\ref{e1b} and \ref{e2b}, and with $I'(\mathrm{X} \mathrm{V}_1 \mathrm{Y}; \mathrm{Z})$ for the systems from Eq.\,\ref{e2a}.

\begin{figure*}[t]
  \begin{center}
    \scalebox{0.4}{\includegraphics*{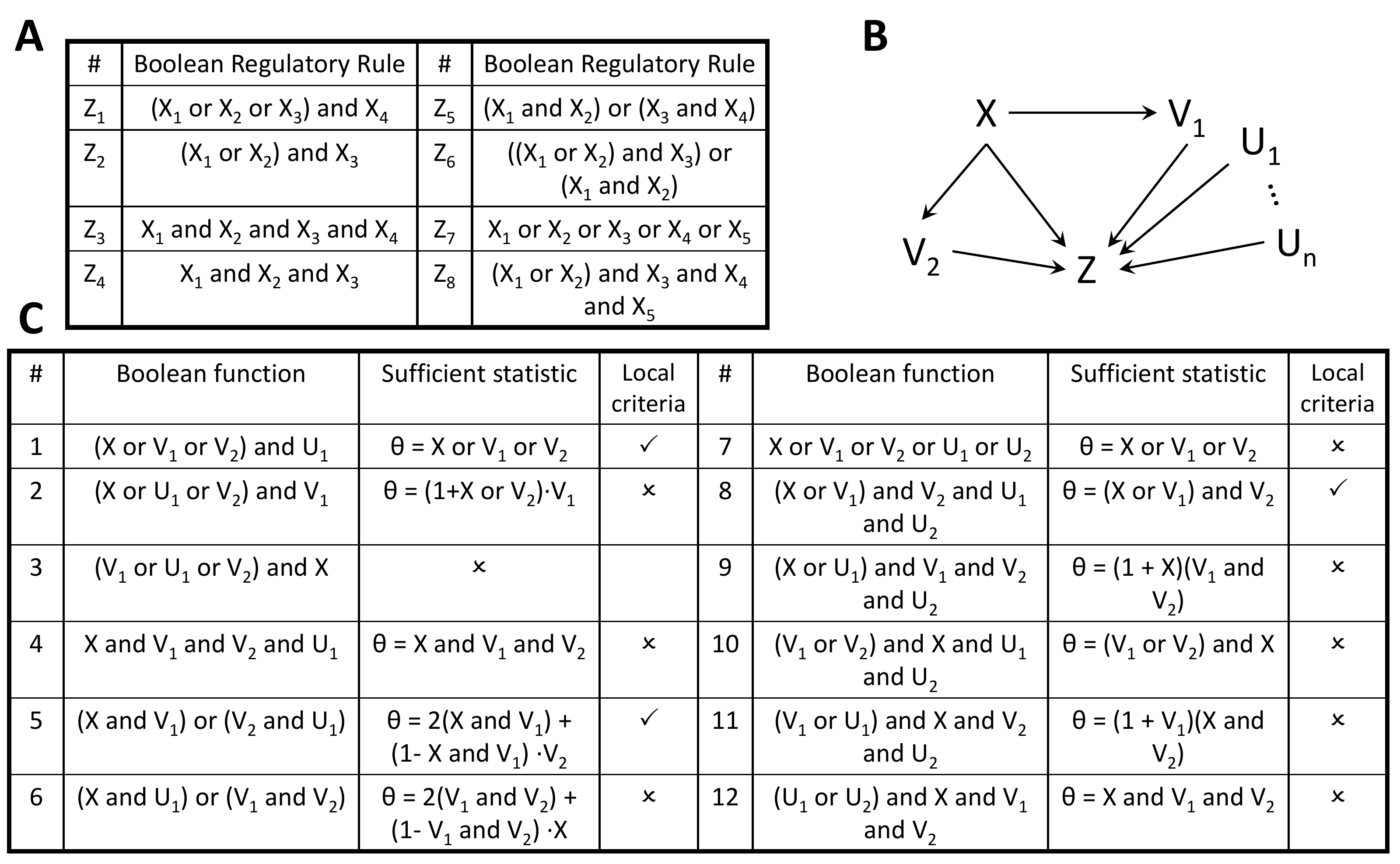}}
  \end{center}
  %\vspace{0.25in}
  \captionsetup{labelformat=empty}
  \caption{Figure S5: Boolean regulatory rules containing sufficient statistics. \textbf{A}) Boolean rules from \cite{Li06}. Several original rules have the same form. The correspondence to Table 1 in \cite{Li06} is: $\mathrm{Z}_1 \sim \mathrm{CaIM}, \mathrm{KOUT}$; $\mathrm{Z}_2 \sim \mathrm{GPA1}, \mathrm{KAP}, \mathrm{Ca}^{2+}_c$; $\mathrm{Z}_3 \sim \mathrm{Atrboh}$; $\mathrm{Z}_4 \sim \mathrm{H}^+ \mathrm{ATPase}, \mathrm{Malate}, \mathrm{ROS}, \mathrm{ABI1}$; $\mathrm{Z}_5 \sim \mathrm{CIS}$; $\mathrm{Z}_6 \sim \mathrm{AnionEM}$; $\mathrm{Z}_7 \sim \mathrm{Depolar}$; $\mathrm{Z}_8 \sim \mathrm{Closure}$. \textbf{B}) Causal structure corresponding to the case studied here, in which three arguments of the Boolean functions are observable. \textbf{C}) Configurations corresponding to all combinations of the Boolean rules in A) with three arguments observable.}
  \label{fBS1}
\end{figure*}

\subsection{Boolean regulatory rules}

In Section \ref{ss43} we studied the identification of sufficient statistics present in the Boolean regulatory rules introduced by \cite{Li06} for the case in which two arguments of the rules are observable. We here further examine configurations with three observable arguments (Figure S5B). We analyzed only rules with 4 or more arguments, so that $\mathrm{Z}$ is not fully determined by the observable arguments. From 12 configurations (Figure S5C), one does not contain a sufficient statistic, while 8 statistics are rejected based on local criteria. Only in 3 cases a valid sufficient statistic exists. Figure S6 shows the performance of the IBSSI method identifying the statistics. The true negative rate is always 1 for all configurations without a sufficient statistic. For the configurations with a statistic, the true positive rate increases with the sample size and is correlated with the normalised information about $\mathrm{Z}$ contained in the arguments of the statistic.

\begin{figure*}[t]
  \begin{center}
    \scalebox{0.55}{\includegraphics*{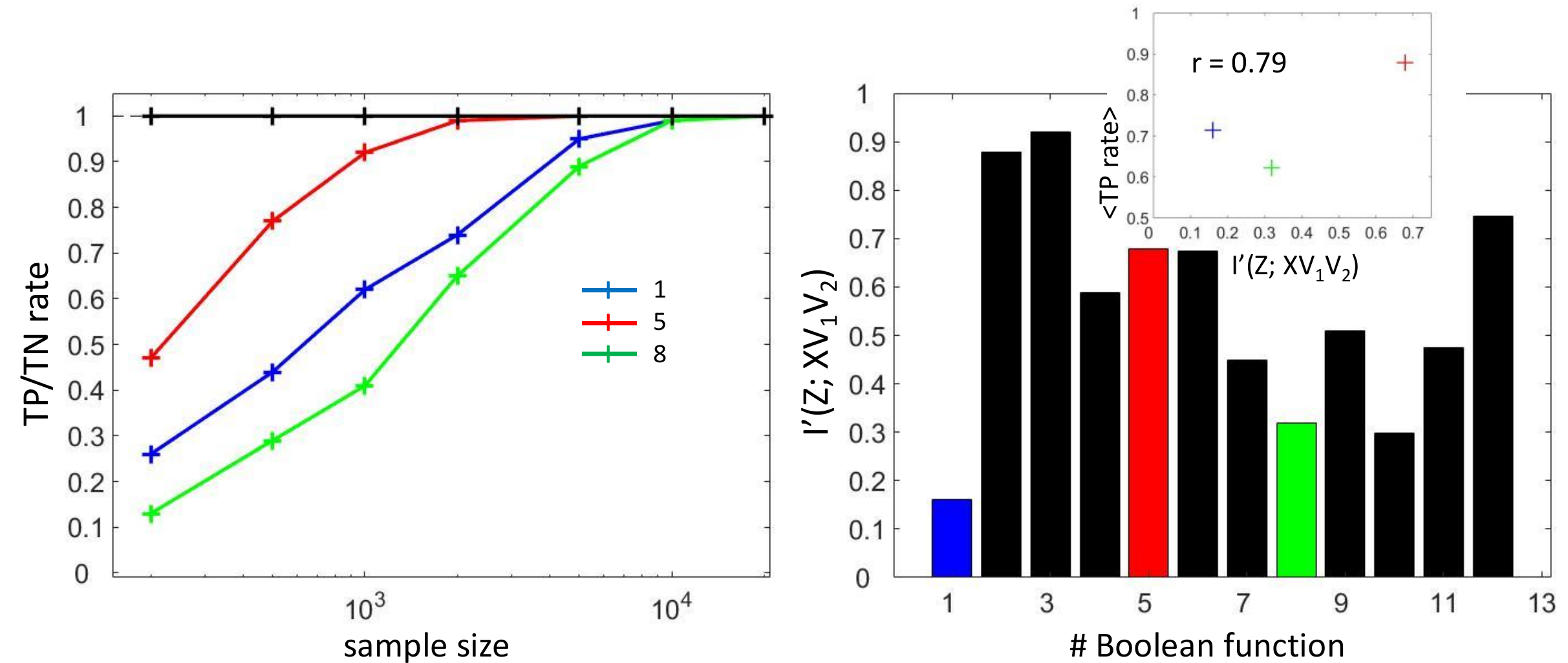}}
  \end{center}
  %\vspace{0.25in}
  \captionsetup{labelformat=empty}
  \caption{Figure S6: Performance of the IBSSI method identifying sufficient statistics in Boolean functions. The structure is analogous to Figure \ref{fB2}, but for the configurations of Figure S5.%\ref{fBS1}C.
}
  \label{fBS2}
\end{figure*}

\subsection{Supplementary characterization of the IB algorithm in dependence on sample size and information levels}

We here extend the analysis of Section \ref{ss42} to study how the output of the IB algorithm depends on its inputs, as well as on the sample size and information level. While in Figure \ref{f2} we examined the average cardinality of the output and the selection ratio as a function of $\max |\hat{\theta}_{IB}|$, $\beta$, and $\mathbf{X}$ for a fixed sample size, we here fix $\beta' = 25$ and study the dependence on the sample size. Furthermore, we define the Identification Ratio as the ratio of configurations for which the IB algorithm retrieves a sufficient set of statistics consistent with the underlying sufficient statistics structurally present. Like for the overall performance of the IBSSI method, the identification ratio of the IB algorithm for fixed inputs is calculated using knowledge of which are the underlying statistics. While the identification ratio and the selection ratio are related, especially when the sample size $N$ is low they may differ. This is because, even if the form of the sufficient statistics is correctly identified, a poor estimation of the distribution $p(\mathrm{Z}, \mathbf{X})$ may result in a rejection of the selection criteria (lines 16-18 of algorithm \ref{alg1}). On the other hand, for an output $\hat{\theta}_{IB}$ inconsistent with the structural statistics, the selection criteria may result in a false positive. For those systems and inputs $\mathbf{X}$ for which a sufficient set of statistics does not exist, we quantify the identification ratio as the ratio of configurations for which $\tilde{\mathbf{X}} = \mathbf{X}$ is retrieved.

For this detailed analysis, we focus on the systems from Eq.\,\ref{e1} (Figure S7). We present the results for at most two values of $\max |\hat{\theta}_{IB}|$. In all cases we use $\max |\hat{\theta}_{IB}| = |\mathbf{X}|$. Furthermore, when a sufficient statistic $\theta_{IB}$ exists for a given type of system and a given $\mathbf{X}$, we also use $\max |\hat{\theta}_{IB}| = |\theta_{IB}|$. The first two columns of Figure S7 show the results for the configurations from Eq.\,\ref{e1a} and the last two from Eq.\,\ref{e1b}. As expected, performance increases with $N$. It is also sensitive to the selection of $\max |\hat{\theta}_{IB}|$. When $\max |\hat{\theta}_{IB}|$ matches the cardinality of the sufficient statistic \--given a certain $\mathbf{X}$\--, the identification ratio is high even for low $N$, and the selection ratio increases sooner with $N$. When $\max |\hat{\theta}_{IB}| = |\mathbf{X}|$, more data are necessary to identify the statistics and the average cardinality of $\hat{\theta}_{IB}$ decreases with $N$ towards the one of the underlying $\theta_{IB}$. This suggests that the requirement in algorithm \ref{alg1} that already for $\max |\hat{\theta}_{IB}| = |\mathbf{X}|$ the returned $\hat{\theta}_{IB}$ fulfills $|\hat{\theta}_{IB}|< |\mathbf{X}|$ (line $4$) may be quite demanding for low $N$ and may be relaxed, especially for systems in which the cardinality of the sufficient statistics is expected to be substantially lower than the one of its arguments.

\begin{figure*}[t]
  \begin{center}
    \scalebox{0.4}{\includegraphics*{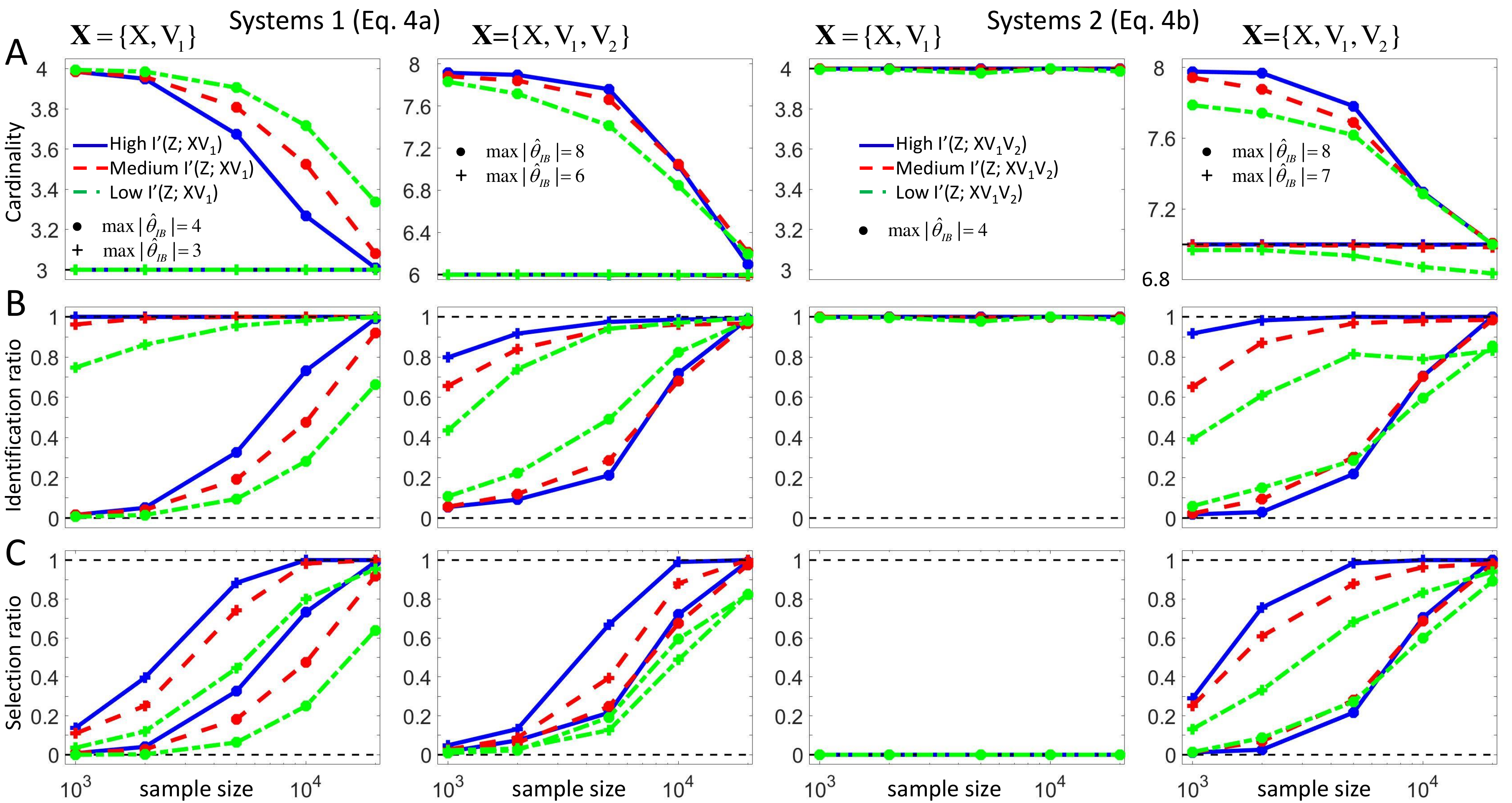}}
  \end{center}
  %\vspace{0.25in}
  \captionsetup{labelformat=empty}
  \caption{Figure S7: Identification of sufficient statistics with the IB algorithm as a function of sample size for the systems from Eqs.\,\ref{e1}. \textbf{A}) Average cardinality across configurations of the inferred $\hat{\theta}_{IB}$. Each panel shows $|\hat{\theta}_{IB}|$ for a specific type of system and a given input $\mathbf{X}$. Information levels are defined like in Figure \ref{f3}. In all panels the results are shown for $\max |\hat{\theta}_{IB}| = |\mathbf{X}|$ ($\bullet$ markers). Furthermore, when a sufficient statistic $\theta_{IB}$ exists, also results using $\max |\hat{\theta}_{IB}| = |\theta_{IB}|$ ($+$ markers) are shown. \textbf{B}) Identification ratio of existing sufficient statistics. \textbf{C}) Selection ratio of the sufficient statistics found with the IB algorithm, following the selection criteria of algorithm \ref{alg1} (lines $16-18$). For all this analysis $\beta' = 25$.}
  \label{f2a}
\end{figure*}

With $\max |\hat{\theta}_{IB}|= |\mathbf{X}|$ and $\mathbf{X} = \{ \mathrm{X}, \mathrm{V}_1, \mathrm{V}_2 \}$, performance is not always higher for a higher information level. For the system from Eq.\,\ref{e1a}, the average $|\hat{\theta}_{IB}|$ decays faster for configurations with low information and their identification ratio is higher. This can be understood taking into account how the configurations have been generated, selecting $n \in \{4, 8, 16, 64\}$ for the binomial distribution $p(\mathrm{Z}| \mathbf{Pa}_z)$. The signal-to-noise ratio increases with $\sqrt{n}$, which means that configurations with higher $n$ also tend to have higher information. However, a higher $n$ also implies a poorer sampling of the distribution $p(\mathrm{Z}, \mathbf{X})$, for a given $N$. While higher information is expected to facilitate the inference of the statistic, a poorer sampling is expected to hinder it. The balance between these two effects depends on the dimensionality of $\mathbf{X}$ and on $N$. When  $\max |\hat{\theta}_{IB}|= |\mathbf{X}|$ is selected as opposed to $\max |\hat{\theta}_{IB}|= |\theta_{IB}|$, the sampling effect is exacerbated because the dimensionality of $\tilde{\mathbf{X}} \otimes \mathbf{X}$, where $p(\tilde{\mathbf{X}}| \mathbf{X})$ has to be determined, is higher. For the system from Eq.\,\ref{e1b}, changes in the balance between the influence of information and sampling accuracy are reflected in a flip for increasing $N$ across information levels for the selection ratio and identification ratio. Both the amount of information and the sampling accuracy can be estimated from the data, and can serve as criteria to decide when to apply the IB method.

\section{Model-based approaches for the identification of sufficient statistics}
\label{ss34}

We here briefly discuss how models could equally be used to infer sufficient statistics. To illustrate how sufficient statistics can appear in parametric models we take as example the case of the widely used generalized linear models (GLMs) \citep{Nelder72}. If the probability distribution of $\mathrm{Z}$ given its parents conforms to a GLM, the influence of the parents on $\mathrm{Z}$ occurs through a linear predictor $\eta =  \mathbf{\beta}^\top \mathbf{Pa}_z$, which determines the conditional mean of $\mathrm{Z}$ through the link function $g$, namely $\mathrm{E}[\mathrm{Z}|\mathbf{Pa}_z] = g^{-1}(\eta)$. Therefore, $\eta$, or a subcomponent of $\mathbf{\beta}^\top \mathbf{Pa}_z$, works as a sufficient statistic for some parents of $\mathrm{Z}$, when the model accurately captures the generative mechanism of $\mathrm{Z}$. In more detail, consider a GLM for $\mathrm{Z}$ with $\eta =  \mathbf{\beta}^\top \mathbf{V}$ and $\mathrm{X} \in \mathbf{V}$. Given a set $\mathbf{S}$ that allows inactivating other paths between $\mathrm{X}$ and $\mathrm{Z}$ other than the direct link through $\eta$, and which does not include all the other predictors $\mathbf{V} \backslash \mathrm{X}$, then there is a sufficient statistic $\theta_z(\mathrm{X}; \tilde{\mathbf{V}})$ with $\tilde{\mathbf{V}} = \mathbf{V} \backslash \{\mathrm{X} , \mathbf{S} \}$, which has the form $\theta_z(\mathrm{X}; \tilde{\mathbf{V}}) = \beta_x \mathrm{X} +  \mathbf{\tilde{\beta}}^\top \tilde{\mathbf{V}}$, where $\tilde{\beta}$ are the coefficients of variables $\tilde{\mathbf{V}}$.

Note that to correctly infer a sufficient set of statistics, a model does not need to properly capture the full functional equation, but only to identify the (possibly much simpler) subcomponents containing the functional sufficient statistics. This will be particularly important when the complexity of the underlying functional equation cannot be well captured within a specific parametric family used to model it. Moreover, if some parents are hidden variables, no model will be able to completely fit the underlying functional equation, while it may still be possible to correctly model subfunctions that act as sufficient statistics for specific parents. Therefore, a modeling approach to infer sufficient statistics will differ from the common use of modeling in score-based methods of structure learning \citep{Chickering2002}, which rank causal structures based on the goodness of fit of models of the full functional equations. The implementation of a model-based approach for the identification of sufficient statistics will be pursued in a future contribution.

\section{Code implementation of the IB algorithm}

We briefly describe some additional details of the code. Additional comments will be found in the code which is to be publicly available. To implement the IB algorithm we adapted the implementation provided by Shabab Bazrafkan, from the Cognitive, Connected $\&$ Computational Imaging Research Group (C3imaging.org), National University of Ireland Galway. That code is publicly available at https://www.mathworks.com as matlabcentral/fileexchange/65937-information-bottleneck-iterative-algorithm. Apart from the most relevant parameters described in Section 4, the algorithm also has some other input parameters. As a stopping criterion, a precision of $10^{-7}$ was selected for changes in the divergence between $p(\tilde{\mathbf{X}}|\mathbf{X})$ for subsequent iterations. A maximum number of $10^{4}$ iterations was selected. We also selected to run the minimization procedure starting from $200$ different starting points, to reduce the selection of local minima. With respect to the original code, our modifications introduce some refinements to deal with probability distributions without a full support and to avoid numerical errors with very high $\beta$ values in the exponent of Eq.\,\ref{IBeq} (see the commented code for details).

\small

%\bibliography{spikedistancesyPRE3}

\end{document}